%% file: preprint.tex
\documentclass{article}
\usepackage{iclr2027_conference,times}
\iclrfinalcopy % Named preprint only; the submission entry point stays anonymous.
\input{preamble}
\newcommand{\mapocodelink}{\url{https://github.com/BorrisonXiao/MAPO-release}}
\author{%
  \textbf{Cihan Xiao}\textsuperscript{1, 2}\thanks{Work done during internship at Tencent.} \quad
  \textbf{Yiwen Shao}\textsuperscript{2} \quad
  \textbf{Chenxing Li}\textsuperscript{2} \quad
  \textbf{Xiang He}\textsuperscript{2} \\[1.0ex]
  \textbf{Zhenwen Liang}\textsuperscript{2} \quad
  \textbf{Steve Yves}\textsuperscript{2} \quad
  \textbf{Sanjeev Khudanpur}\textsuperscript{1} \quad
  \textbf{Liefeng Bo}\textsuperscript{2} \\[1.0ex]
  \textsuperscript{1}Johns Hopkins University \quad 
  \textsuperscript{2}Tencent Hunyuan \\
  \texttt{cxiao7@jhu.edu} \\
}

\begin{document}
\maketitle
\lhead{Preprint} % Do not claim an accepted ICLR publication.
\input{manuscript}

\end{document}

%% file: preamble.tex
\usepackage[utf8]{inputenc}
\usepackage[T1]{fontenc}
\usepackage{amsmath,amssymb,amsfonts}
\usepackage{graphicx,subcaption,multirow}
\usepackage{tcolorbox}
\usepackage{hyperref,url,booktabs,nicefrac,microtype,xcolor}
\usepackage{algorithm,algorithmic,wrapfig}
\usepackage{placeins} % Keep the Appendix D benchmark tables before qualitative examples.
\title{Escape the Language Prior: Mitigating Late-Stage Modality Collapse in Audio Reasoning via Modality-Aware Policy Optimization}

%% file: manuscript.tex
\begin{abstract}
Audio and omni-modal large language models exhibit impressive cross-modal reasoning capabilities. However, standard reinforcement learning methods such as GRPO do not explicitly weight individual tokens by their dependence on the non-text source modality. This exacerbates \textit{late-stage modality collapse} during extended chain-of-thought generation, where models progressively abandon the primary source signal in favor of compressed textual priors, leading to confident but ungrounded hallucinations.
To address this, we introduce Modality-Aware Policy Optimization (MAPO), a novel dual-branch reinforcement learning framework. First, MAPO dynamically concentrates the policy gradient on modality-critical tokens using a modality relevance mask, which is derived from the cross-modal differential entropy between an audio-ablated reference and the multimodal policy. Second, it integrates an auxiliary attention loss branch that applies a targeted, temporally scaled penalty to the model's internal attention distributions. This encourages sustained attention to the source modality throughout the reasoning trace.
Evaluations on complex audio reasoning benchmarks demonstrate that MAPO substantially improves long-horizon reasoning fidelity and multimodal instruction following, achieving highly competitive performance on several key benchmarks among open-weights models. By relying strictly on native statistical signals rather than domain-specific inductive biases, MAPO offers a promising foundation for mitigating epistemic collapse across diverse multimodal systems.
\end{abstract}

\section{Introduction}

\begin{figure}[!htb]
    \centering
    \includegraphics[width=1.0\textwidth]{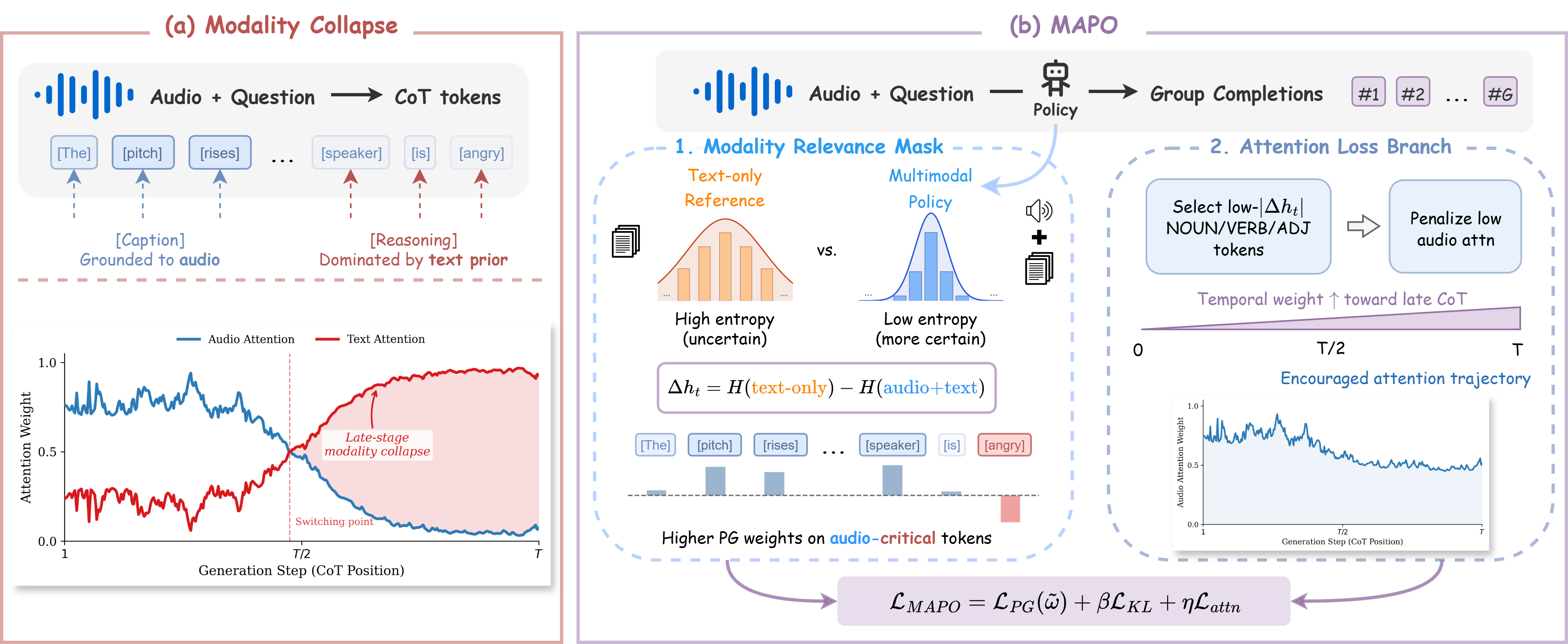}
    \caption{Overview of the MAPO framework. \textbf{(a)} Late-stage modality collapse, where attention shifts from the source audio to the text prior during CoT reasoning. \textbf{(b)} MAPO mitigates this via a dual-branch architecture. A \textit{modality relevance mask} uses cross-modal differential entropy ($\Delta h_t$) to focus the policy gradient on audio-critical tokens. Simultaneously, an \textit{attention loss branch} applies a temporally scaled penalty to substantive tokens with low audio attention, encouraging sustained access to source audio.}
    \label{fig:method}
\end{figure}

Audio and omni-modal large language models have demonstrated remarkable capabilities by integrating non-text modalities into complex reasoning systems. However, post-training these models using standard reinforcement learning algorithms (e.g., PPO~\citep{schulman2017proximalpolicyoptimizationalgorithms}, DPO~\citep{rafailov2024directpreferenceoptimizationlanguage}, GRPO~\citep{shao2024deepseekmathpushinglimitsmathematical}) exposes a critical structural inefficiency: their objectives do not explicitly weight individual tokens by their dependence on the source modality. In cross-modal generation, tokens are fundamentally not equal. Many serve purely as linguistic scaffolding that the model's internal language prior can accurately predict. Conversely, a critical subset of tokens relies entirely on the non-text source modality to capture essential acoustic attributes or events. Uniform token coefficients provide no explicit mechanism to prioritize positions requiring active cross-modal grounding.

Maintaining access to the source is especially important during extended cross-modal chain-of-thought (CoT)~\citep{wei2023chainofthoughtpromptingelicitsreasoning} reasoning. We find that, in this process, models can increasingly abandon the uncompressed primary source signal (e.g., raw audio) in favor of their own heavily compressed textual summaries, a pattern we refer to as \textit{late-stage modality collapse}. Relying on these intermediate captions causes the model's reasoning to drift into ungrounded hallucinations. Crucially, as the model succumbs to its language prior and ignores the source modality, it becomes confidently disconnected. This epistemic collapse masks the model's true uncertainty, which makes standard statistical interventions such as entropy maximization ineffective at recovering lost grounding.

To address these intertwined challenges, we introduce \textbf{Modality-Aware Policy Optimization (MAPO)}, a novel RL framework designed to break the language prior and encourage sustained cross-modal grounding. Building upon GRPO, MAPO modifies the learning objective through two complementary mechanisms. First, it introduces a \textit{Modality Relevance Mask} that reweights the policy gradient. By computing the cross-modal differential entropy between a reference model with text-only inputs and the multimodal policy, MAPO dynamically upweights the gradient contribution of tokens where the source modality significantly alters the predictive distribution. Second, an auxiliary \textit{Attention Loss Branch} directly combats late-stage modality collapse. Guided by an inverse relevance mask and temporally increasing weights, it explicitly penalizes the neglect of the source modality at linguistically substantive tokens (identified via part-of-speech gating). Combined with a soft task-failure gate, this penalty encourages the model to maintain attention to the source signal deep into the reasoning chain. We focus our empirical evaluation on a wide variety of audio reasoning tasks as a rigorous testbed.\footnote{Our code is available at \mapocodelink.}

In summary, our main contributions are threefold:
\begin{enumerate}
    \item We identify and formalize the limitations of uniform token weighting and the phenomenon of late-stage modality collapse in cross-modal reinforcement post-training, highlighting how the language prior masks true model uncertainty during long-horizon reasoning.
    \item We propose MAPO, a dual-branch optimization framework that leverages cross-modal differential entropy to prioritize modality-critical tokens and introduces a targeted attention penalty to enforce sustained cross-modal grounding. 
    \item Through extensive empirical evaluation on complex audio reasoning tasks (encompassing speech, music, and sound events), we demonstrate that MAPO improves both the fidelity of long-form reasoning and the model's robustness against language-prior hallucinations, establishing a robust algorithmic foundation for mitigating late-stage modality collapse.
\end{enumerate}

\section{Related work}

\textbf{Omni-modal large language models.}
Recent advancements in omni-modal large language models enable unified, native processing of text, vision, and audio, bypassing the information loss inherent to cascaded systems. Our empirical investigations utilize Qwen3-Omni-Thinking~\citep{xu2025qwen3omnitechnicalreport}, a model specifically optimized for extended cross-modal chain-of-thought generation, distinguishing it from its instruction-following counterpart, Qwen3-Omni-Instruct~\citep{xu2025qwen3omnitechnicalreport}.
Beyond Qwen3, the cross-modal reasoning landscape includes proprietary systems like the Gemini~\citep{comanici2025gemini25pushingfrontier} series. It also includes open-weight models such as Audio Flamingo 3~\citep{goel2025audioflamingo3advancing}, Step-Audio-2~\citep{wu2025stepaudio2technicalreport}, MiMo-Audio~\citep{coreteam2025mimoaudioaudiolanguagemodels}, and Covo-Audio~\citep{wang2026covoaudiotechnicalreport}, which pioneer specialized native audio reasoning and full-duplex conversational capabilities. However, despite these architectural leaps, those systems are trained via standard text-based post-training objectives. By uniformly weighting all generated tokens, these models remain structurally susceptible to language-prior dominance and late-stage modality collapse.

\textbf{Group Relative Policy Optimization (GRPO).}
GRPO~\citep{shao2024deepseekmathpushinglimitsmathematical} aligns models efficiently by estimating advantages over a sampled group rather than utilizing a separate value model. For a prompt $x$ and a group of $G$ completions $\{y^{(g)}\}_{g=1}^{G}$, the relative advantage $\hat{A}^{(g)}$ normalizes the reward $r(x, y^{(g)})$ within the group:
$$\hat{A}^{(g)} = \frac{r(x, y^{(g)}) - \text{mean}(\{r(x, y^{(g')})\})}{\text{std}(\{r(x, y^{(g')})\}) + \varepsilon}$$

The policy $\pi_\theta$ is optimized via a clipped surrogate objective regularized by a per-token sample-based estimator of the Kullback-Leibler (KL) divergence against a frozen reference policy $\pi_{\text{ref}}$. Specifically, we utilize the $k_3$ unbiased estimator to ensure computational efficiency over the full vocabulary sum. For the $g$-th completion of length $T_g$ with importance ratio $\rho_t^{(g)} = \frac{\pi_\theta(y_t^{(g)} \mid y_{<t}^{(g)}, x)}{\pi_{\text{old}}(y_t^{(g)} \mid y_{<t}^{(g)}, x)}$:
$$\mathcal{L}_{\text{PG}}^{\text{GRPO}} = -\frac{1}{T_g}\sum_{t=1}^{T_g} \min\!\Big(\rho_t^{(g)} \hat{A}^{(g)}, \;\text{clip}(\rho_t^{(g)}, 1{-}\epsilon, 1{+}\epsilon)\,\hat{A}^{(g)}\Big)$$
$$\mathcal{L}_{\text{KL}} = \frac{1}{T_g}\sum_{t=1}^{T_g}\Big[\exp\!\big(\log\pi_{\text{ref}}(y_t^{(g)}) - \log\pi_\theta(y_t^{(g)})\big) - \big(\log\pi_{\text{ref}}(y_t^{(g)}) - \log\pi_\theta(y_t^{(g)})\big) - 1\Big]$$

The final objective is $\mathcal{L}_{\text{GRPO}} = \mathcal{L}_{\text{PG}}^{\text{GRPO}} + \beta \mathcal{L}_{\text{KL}}$. GRPO assigns the same weight, $1/T_g$, to each token when computing the policy loss for a completion. The resulting gradient contributions nevertheless differ across tokens because they depend on token-specific likelihood ratios and score-function gradients. To make the weighting explicitly reflect dependence on the source modality, MAPO uses cross-modal differential entropy to determine each token's relative weight.

\textbf{Cross-Modal Alignment and Grounding.} Recent approaches inject modality-aware feedback through external rewards, often relying on large teacher models to provide auxiliary sequence-level signals (e.g., \citep{audiothinker_250808039}, \citep{audiodeepthinker}). While effective, these extrinsic methods apply feedback uniformly, algorithmically ignoring the crucial token-by-token variance in modality dependence. Because MAPO intrinsically resolves this granular inefficiency without requiring heavy external reward models, it serves as a distinct, orthogonal framework that remains fully compatible with these sequence-level reward strategies.

At the granular level, prior methods such as PAPO \citep{papo} and concurrent works in the vision-language domain have recognized the limitations of uniform token weighting and attention decay. For instance, \citep{huang2025spotlight} isolates perceptually pivotal tokens to focus policy updates, while \citep{wang2026vgpo} combats temporal visual forgetting by progressively elevating visual expectations through attention compensation. While these methods share our high-level motivation of targeted policy optimization, they rely heavily on domain-specific visual similarity and dependency metrics.

In contrast, MAPO intrinsically addresses the epistemic failure mode without requiring external reward models or domain-specific heuristics. Previous granular approaches \citep{entropy_tokens_260313366} have utilized standard predictive entropy to isolate dependent tokens. Yet, standard policy entropy fails during late-stage modality collapse. As the language prior takes over, the model becomes pathologically confident in its hallucinations, causing the entropy signal to vanish exactly when external grounding is needed most. MAPO overcomes this by anchoring its metric against a text-only reference, exposing hidden modality dependence and directly steering the internal attention distribution.

% ----------Method----------

\section{Modality-Aware Policy Optimization (MAPO)}

To formalize the MAPO framework, we first establish the cross-modal reinforcement learning problem setting and define the notation used throughout our methodology. Following this, we introduce two critical quantitative metrics that serve as the foundation of our method: a measure of the model's internal cross-modal grounding (\textit{audio attention mass}), and a statistical signal to isolate key modality-dependent tokens (\textit{cross-modal differential entropy}). We utilize these metrics to dissect the phenomenon of late-stage modality collapse before introducing our dual-branch optimization objective.

\subsection{Problem formulation}

We consider a cross-modal generation setting where an omni-modal language model, featuring a transformer decoder with $L$ layers and $H$ attention heads per layer, acts as the policy $\pi_\theta$ parameterized by $\theta$. The model receives a multimodal prompt $x$, which inherently comprises both a text instruction sequence $x_{\text{text}}$ and a non-text source modality $x_{\text{audio}}$ (e.g., continuous acoustic features mapped to the model's input space).
Given the prompt $x$, the policy autoregressively generates a text completion $y = (y_1, \dots, y_T)$, where each token $y_t \in \mathcal{V}$ belongs to the textual vocabulary $\mathcal{V}$, and $T$ denotes the total number of generated tokens for that specific completion. The likelihood of generating the full sequence is formulated as $\prod_{t=1}^T \pi_\theta(y_t \mid y_{<t}, x)$.

During post-training via GRPO, for a given prompt $x$, the policy samples a group of $G$ independent completions, denoted as $\{y^{(g)}\}_{g=1}^{G}$. Standard reinforcement learning objectives construct the sequence-level loss by uniformly averaging the token-level losses over the sequence length $T$. MAPO challenges this uniform $1/T$ dependency. Our objective is to design a dynamic, token-level weighting mechanism that structurally differentiates tokens relying purely on the language prior (dependent only on $y_{<t}$ and $x_{\text{text}}$) from those requiring cross-modal perception (critically dependent on $x_{\text{audio}}$).

\subsection{Audio attention mass and late-stage modality collapse}

To dynamically quantify the extent to which the model grounds its reasoning in the non-text source signal at any given generation step, we track the \textit{audio attention mass}. Conceptually, this metric captures the \textit{aggregate probability mass within the model's self-attention distribution that is routed specifically to the cross-modal input sequence}, serving as a direct proxy for the active perceptual bandwidth allocated to the audio signal.

Let $\alpha_{t,j}^{(l,h)}$ denote the attention weight that the query at generation position $t$ assigns to the key at sequence position $j$ within layer $l \in \{1, \dots, L\}$ and attention head $h \in \{1, \dots, H\}$. If $\mathcal{S}_{\text{audio}}$ represents the set of token indices corresponding to the primary audio input $x_{\text{audio}}$, the raw audio attention mass for a specific head is defined as $m_t^{(l,h)} = \sum_{j \in \mathcal{S}_{\text{audio}}} \alpha_{t,j}^{(l,h)}$. Empirically, cross-modal perception is rarely distributed uniformly across all attention heads; rather, specific heads specialize in modality routing. To characterize this behavior, we define both the \textit{max-head} and \textit{mean-head} attention mass for a given layer $l$:
$$m_t^{(l, \text{max})} = \max_{h} m_t^{(l,h)}, \quad m_t^{(l, \text{mean})} = \frac{1}{H} \sum_{h=1}^H m_t^{(l,h)}$$

To capture the strongest grounding signal for our optimization objective, we utilize the max-head reduction. The final audio attention mass $a_t$ for token $t$ is computed by averaging this max-head signal across a targeted set of deep transformer layers $L_{\text{tgt}}$ (e.g., the final layers before the output projection):
$$ a_t = \frac{1}{|L_{\text{tgt}}|}\sum_{l \in L_{\text{tgt}}} m_t^{(l, \text{max})} $$

\begin{figure}[!htb]
    \centering
    \includegraphics[width=0.88\textwidth]{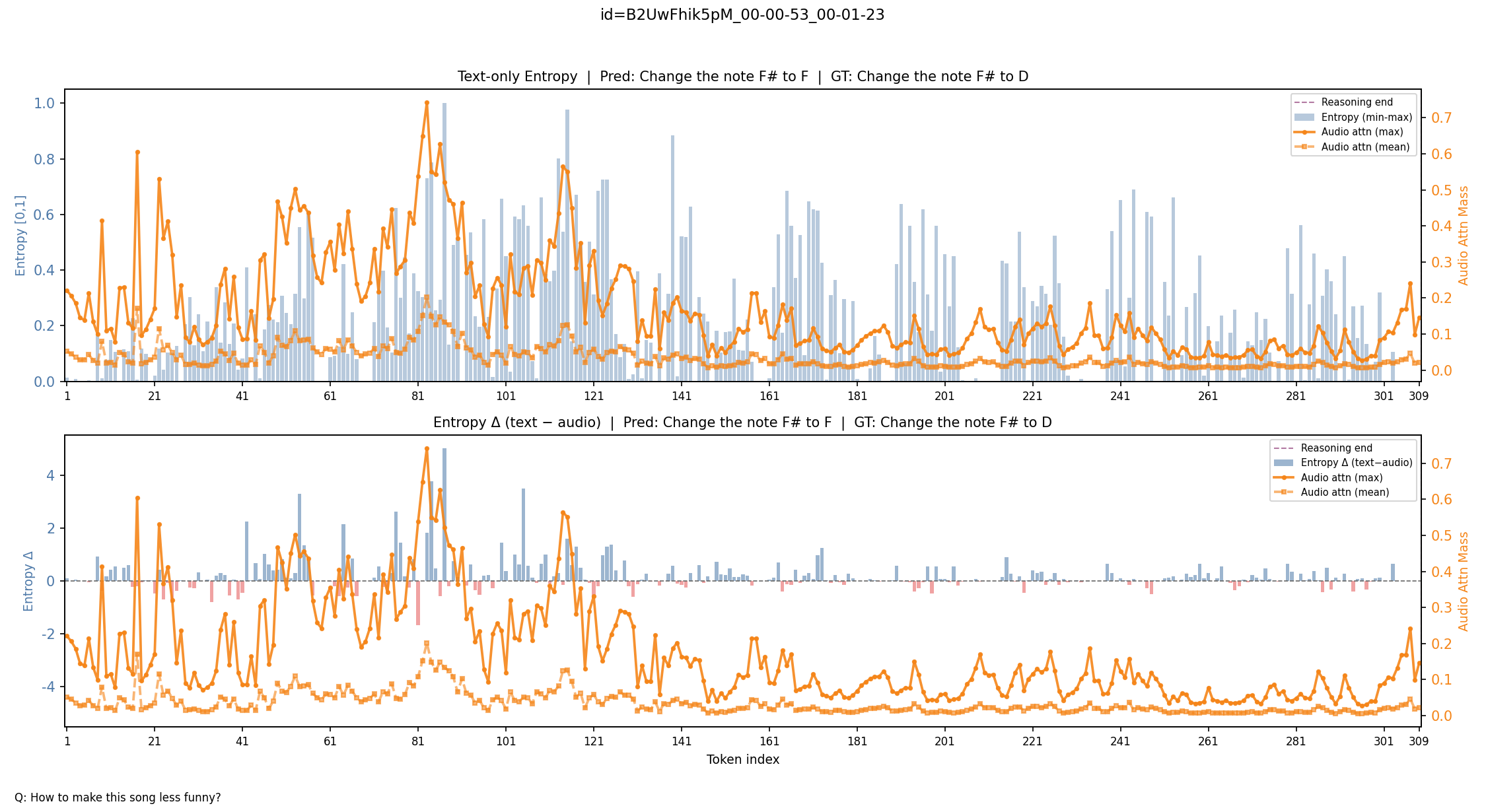}
    \caption{Late-stage modality collapse and key token extraction. Plots display the temporal decay of audio attention mass (\textbf{solid:} max-head reduction, \textit{dashed:} mean-head reduction) alongside uncertainty metrics. \textbf{(a)} Text-only entropy fails to track this loss of source grounding. \textbf{(b)} cross-modal differential entropy effectively isolates true ``key tokens''; this signal correlates with the attention decay.}
    \label{fig:modality_collapse}
\end{figure}

This metric exposes the structural vulnerability of cross-modal CoT generation. As illustrated by the line charts in Figure \ref{fig:modality_collapse}, while the initial reasoning steps maintain a high audio attention mass, this mass exhibits a pronounced decay as the generation lengthens. In the late stages of generation, $a_t$ collapses near zero, indicating that the model has effectively stopped ``listening'' to the raw audio, instead auto-regressively feeding on its own textual context.

\subsection{Cross-modal differential entropy}

To counteract modality collapse, we must reliably identify the ``key tokens'', which represent the specific reasoning steps that require audio perception. To expose this hidden modality dependence, we introduce \textit{cross-modal differential entropy} ($\Delta h_t$), the token-level predictive entropy difference between a text-only reference policy and the multimodal policy:
$$ \Delta h_t = H(\pi_{\text{text-ref}}(\cdot \mid y_{<t}, x_{\text{text}})) - H(\pi_\theta(\cdot \mid y_{<t}, x)) $$

Both terms use the same text prompt and generated prefix $y_{<t}$, including any caption already generated; only $x$ contains source audio. In our diagnostic analyses, both terms are evaluated using the same fixed model checkpoint. This comparison isolates the effect of audio conditioning on predictive uncertainty at each generation position, conditional on the information already expressed in the prefix. As demonstrated in Figure \ref{fig:modality_collapse}(b), cross-modal differential entropy correlates with the trend of audio mass decay.

During training, we compute cross-modal differential entropy between the live multimodal policy and a text-only reference policy. The reference policy is configurable: its parameters may follow the current policy or remain fixed. Our experiments use a fixed reference for GPU memory efficiency, reusing the model already maintained for the GRPO KL penalty (Appendix~\ref{sec:training_reference}).

\subsection{Modality relevance mask and reweighted policy gradient}
\label{sec:mrm}

MAPO incorporates cross-modal differential entropy into the GRPO objective through a token-level \textit{Modality Relevance Mask} ($\tilde{\omega}$). This mask replaces uniform averaging coefficients with weights determined by $|\Delta h_t|$, assigning greater emphasis to positions with larger absolute cross-modal differential entropy, subject to clipping.

To construct this mask, we apply a softmax over the absolute differential entropy $|\Delta h_t|$. Because cross-modal CoT reasoning lengths vary drastically, we stabilize the weight assignment using a length-scaled temperature $\tau_T = \tau_{\text{base}} \log(\max(T, 2))$. To keep token weights comparable across completions of different lengths, we multiply the softmax weights by $T$, giving a mean weight of one before clipping. We then cap each weight at $C_{\mathrm{mask}}$ to limit large individual weights. The final detached mask is:
\begin{equation}
    \tilde{\omega}_t = \min\!\left( T \cdot \frac{\exp\!\big(|\Delta h_t| / \tau_T\big)}{\sum_{t'=1}^{T} \exp\!\big(|\Delta h_{t'}| / \tau_T\big)}, \; C_{\mathrm{mask}} \right)
\end{equation}

Clipping can reduce the total weight assigned to a completion. We therefore normalize by the sum of the weights, so the policy loss remains a weighted average over the generated tokens. Let $\ell_t^{\text{PG}} = -\min\!\big(\rho_t \hat{A}, \;\text{clip}(\rho_t, 1{-}\epsilon, 1{+}\epsilon)\,\hat{A}\big)$ denote the GRPO policy loss at token $t$. For a sampled group of $G$ completions $\{y^{(g)}\}_{g=1}^G$ with lengths $T_g$, the MAPO reweighted policy gradient replaces the uniform $1/T$ averaging with:
\begin{equation}
    \mathcal{L}_{\text{PG}}(\tilde{\omega}) = \frac{1}{G} \sum_{g=1}^{G} \frac{\sum_{t=1}^{T_g} \tilde{\omega}_t^{(g)} \cdot \ell_t^{\text{PG}, (g)}}{\sum_{t=1}^{T_g} \tilde{\omega}_t^{(g)}} \label{eq:pg_reweight}
\end{equation}
This normalization makes the policy loss invariant to a common positive rescaling of the final detached weights, while retaining the relative token weights assigned by the modality relevance mask.

\subsection{Attention loss branch}

\begin{wrapfigure}{r}{0.34\textwidth}
    \centering
    \includegraphics[width=\linewidth]{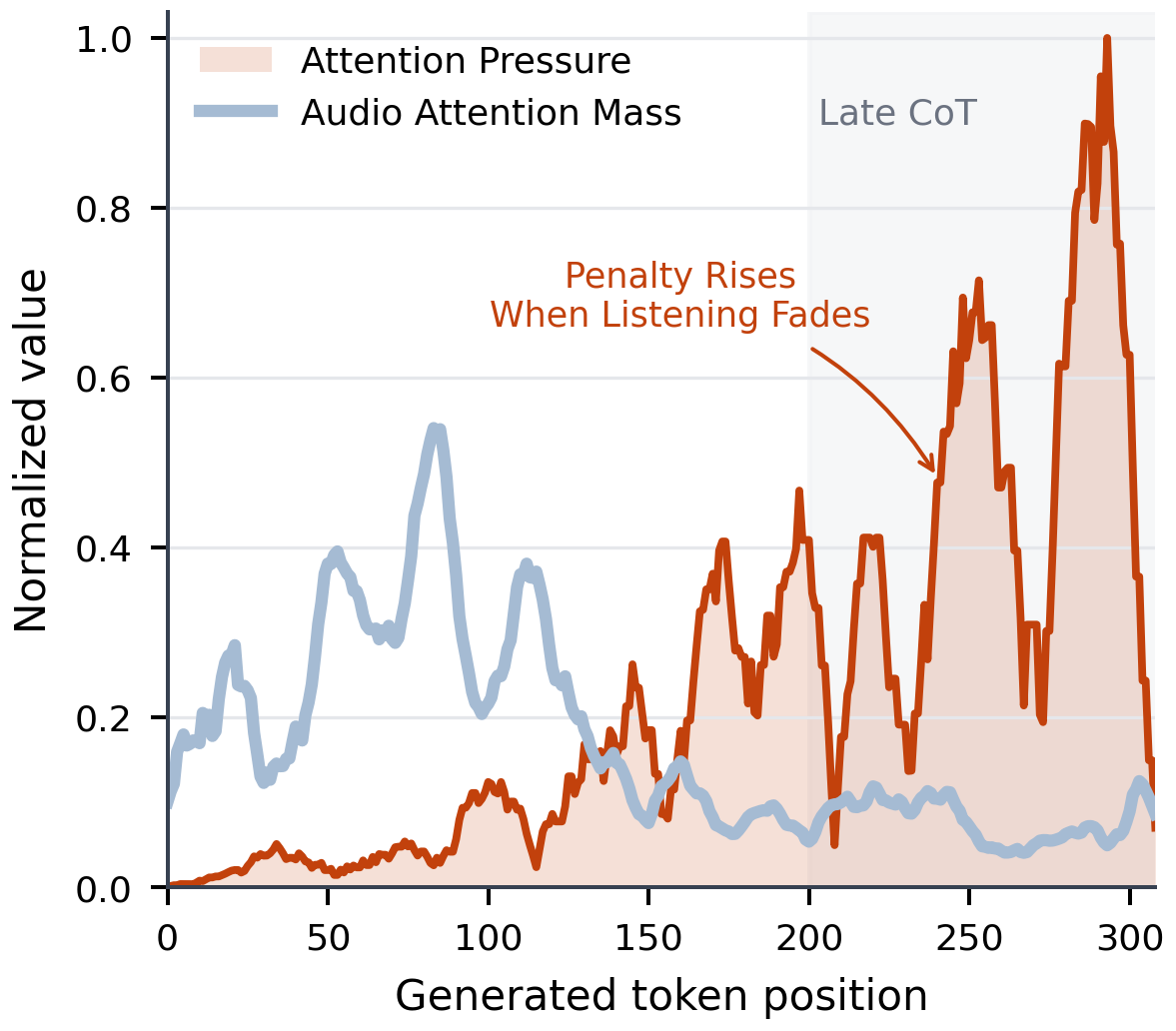}
    \caption{Attention loss trajectory.}
    \label{fig:inline_entropy}
\end{wrapfigure}
While the modality relevance mask corrects the policy gradient budget, it fundamentally relies on the model maintaining some latent awareness of the cross-modal signal. To directly combat late-stage modality collapse, MAPO incorporates an auxiliary \textit{attention loss branch} ($\mathcal{L}_{\text{attn}}$) that penalizes the neglect of the source signal deep within the reasoning chain. We define a token-level log-penalty based on the audio attention mass: $\ell_{t}^{\text{attn}} = -\log(a_t + \varepsilon)$. To avoid forcing unnatural cross-modal attention on functional vocabulary, we restrict this penalty to linguistically substantive tokens using a binary part-of-speech (POS) gate $g_t^{\text{pos}} \in \{0, 1\}$.

Optimization pressure is focused on positions exhibiting severe grounding deficits via an \textit{Inverse Relevance Weight} $\tilde{\nu}_t$. This mask targets substantive tokens with suspiciously low differential entropy, normalized over the valid active subset $T_{\text{pos}} = \sum_t g_t^{\text{pos}}$ and capped at $C_{\mathrm{mask}}$:
\begin{equation}
    \tilde{\nu}_t = \min\!\left( T_{\text{pos}} \cdot \frac{g_t^{\text{pos}} \cdot \exp\!\big(-|\Delta h_t| / \tau_T\big)}{\sum_{t'=1}^{T} g_{t'}^{\text{pos}} \cdot \exp\!\big(-|\Delta h_{t'}| / \tau_T\big)}, \; C_{\mathrm{mask}} \right)
\end{equation}

To structurally counteract attention decay, we apply a non-decreasing temporal weight $(t/T)^\kappa$ ($\kappa \ge 0$), concentrating the penalty at the tail end of the sequence. To prevent distorting successful reasoning trajectories, the sequence-level penalty is scaled by the advantage magnitude $|\hat{A}|$ and a soft task-failure gate $\hat{f} = \max(\mathbb{1}[r_{\text{acc}} \leq \theta_{\text{fail}}], \alpha)$. Capping this prefactor at $C_{\mathrm{pref}}$ to prevent explosive gradients, the final POS-gated mean reduction for a group of $G$ completions is:
\begin{equation}
    \mathcal{L}_{\text{attn}} = \frac{1}{G}\sum_{g=1}^{G} \min\!\Big(\hat{f}^{(g)} \cdot |\hat{A}^{(g)}|, \, C_{\mathrm{pref}}\Big) \frac{1}{N_{\mathrm{pos}}^{(g)}} \sum_{t=1}^{T_g} (t/T_g)^\kappa \cdot \tilde{\nu}_t^{(g)} \cdot \ell_{t}^{\text{attn},(g)} \label{eq:attn_loss}
\end{equation}
Here, $N_{\mathrm{pos}}=\max(1,T_{\mathrm{pos}})$; when no token passes the POS gate, we set $\tilde{\nu}_t=0$ throughout the sequence.

The attention loss is computed directly from audio attention mass. It therefore provides an additional training signal that encourages the model to maintain attention to the source audio at the selected tokens.

\begin{algorithm}[!htb]
\caption{Modality-Aware Policy Optimization (MAPO)}
\label{alg:mapo}
\begin{algorithmic}[1]
\REQUIRE $\pi_\theta$, $\pi_{\text{text-ref}}$, $\pi_{\text{ref}}$, $r$, learning rate $\alpha$, hyperparameters $\beta, \eta, \kappa, \tau_{\text{base}}$
\FOR{iteration $= 1, 2, \dots$}
    \STATE Sample $x = (x_{\text{text}}, x_{\text{audio}})$
    \STATE Sample $G$ completions $\{y^{(g)}\}_{g=1}^G \sim \pi_\theta(\cdot \mid x)$
    \STATE Compute advantages $\hat{A}^{(g)}$ using $r(x, y^{(g)})$
    \FOR{$g=1 \dots G$}
        \STATE $\Delta h_t^{(g)} \leftarrow H(\pi_{\text{text-ref}}) - H(\pi_\theta), \quad \forall t$
        \STATE Extract $a_t^{(g)}$; compute masks $\tilde{\omega}_t^{(g)}$ and $\tilde{\nu}_t^{(g)}$ using $\Delta h_t^{(g)}$
        \STATE Compute $\mathcal{L}_{\text{PG}}^{(g)}$ (Eq.~\ref{eq:pg_reweight}) and $\mathcal{L}_{\text{attn}}^{(g)}$ (Eq.~\ref{eq:attn_loss})
        \STATE $\mathcal{L}_{\text{KL}}^{(g)} \leftarrow D_{\text{KL}}(\pi_{\text{ref}} \parallel \pi_\theta)$
    \ENDFOR
    \STATE $\mathcal{L}_{\text{MAPO}} \leftarrow \frac{1}{G}\sum_{g=1}^G \big(\mathcal{L}_{\text{PG}}^{(g)} + \beta \mathcal{L}_{\text{KL}}^{(g)} + \eta \mathcal{L}_{\text{attn}}^{(g)}\big)$
    \STATE $\theta \leftarrow \theta - \alpha \nabla_\theta \mathcal{L}_{\text{MAPO}}$
\ENDFOR
\end{algorithmic}
\end{algorithm}

\subsection{Total MAPO objective}

The complete training procedure for MAPO is summarized in Algorithm \ref{alg:mapo}. Combining the reweighted policy gradient, the attention-steering branch, and the sequence-level KL divergence penalty, the total optimization objective for MAPO is defined as:
\begin{equation}
    \mathcal{L}_{\text{MAPO}} = \mathcal{L}_{\text{PG}}(\tilde{\omega}) + \beta \cdot \mathcal{L}_{\text{KL}} + \eta \cdot \mathcal{L}_{\text{attn}}
\end{equation}
where $\beta$ regulates the deviation from the reference policy and $\eta$ dictates the strength of the cross-modal attention penalty. Intuitively, these two core mechanisms operate synergistically: the policy gradient reweighting ($\mathcal{L}_{\text{PG}}(\tilde{\omega})$) dynamically identifies existing key tokens to disproportionately allocate the optimization budget, while the attention loss branch ($\mathcal{L}_{\text{attn}}$) actively encourages the model to create and sustain these modality-dependent key tokens deeper into the extended reasoning trace.

% ----------Experiments----------

\section{Experiments}

\textbf{Experimental Setup.}
We initialize our multimodal policy $\pi_\theta$ with the Qwen3-Omni-30B-A3B-Thinking model. This same base model serves as both the standard frozen reference ($\pi_{\text{ref}}$) for the KL penalty and the text-only reference ($\pi_{\text{text-ref}}$) used to compute the cross-modal differential entropy. Training is conducted on 32 NVIDIA H20 GPUs distributed across 4 compute nodes. To efficiently decouple autoregressive generation from gradient computation, we employ an asynchronous training setup: 3 nodes are dedicated to policy optimization and hosting the frozen reference models, while the remaining node operates as a dedicated vLLM server for high-throughput trajectory rollouts.

\textbf{Reward Formulation.}
The final reward $r(x, y)$ prioritizes correctness while enforcing structural and logical constraints:
\begin{equation}
    r(x, y) = \lambda_{\text{acc}} r_{\text{acc}}(x, y) + \lambda_{\text{format}} r_{\text{format}}(x, y) + \lambda_{\text{cons}} r_{\text{cons}}(x, y)
\end{equation}
Here, $r_{\text{acc}}$ checks the final answer against the ground truth, and $r_{\text{format}}$ checks that the response follows the chain-of-thought template. The consistency reward $r_{\text{cons}}$ \citep{audiothinker_250808039} uses a frozen Qwen3-30B-A3B-Instruct checker hosted on the dedicated vLLM rollout server to penalize contradictory or prematurely truncated reasoning (see Appendix \ref{sec:appendix_prompt} for the prompt). In our experiments, we use default weights of $\lambda_{\text{acc}}=2$, $\lambda_{\text{format}}=1$, and $\lambda_{\text{cons}}=1$.

\textbf{Training Data and Experimental Settings.}
We first examine MAPO's components in preliminary experiments on AVQA~\citep{yang2022avqa}, using both LoRA and full fine-tuning. Its curated question--answer pairs and emphasis on basic sound-event understanding make it a natural starting point for examining how the model responds to different optimization components.

Following the Audio-DeepThinker~\citep{audiodeepthinker}, we continue full fine-tuning of MAPO and GRPO from their respective AVQA-trained checkpoints on its 29k-example audio QA dataset. This mixture consists of diverse acoustic perception capabilities across four domains: mixed general-audio reasoning (48.6\%, e.g., AudioSet~\citep{Gemmekeetal2017}, TACOS~\citep{primus2025tacostemporallyalignedaudiocaptions}), speech and paralinguistic understanding (27.1\%, e.g., LibriTTS-R~\citep{koizumi2023librittsrrestoredmultispeakertexttospeech}, IEMOCAP~\citep{Busso2008}), structural music understanding (17.1\%, e.g., AudioSkills~\citep{Ghoshetal2025}, MusicCaps~\citep{Agostinellietal2023}), and acoustic scene classification (7.1\%, e.g., CochlScene~\citep{jeong2022cochlsceneacquisitionacousticscene}).

\subsection{Ablation results}

We begin by examining how MAPO's two optimization branches contribute to downstream performance, starting with the modality relevance mask and then adding the auxiliary attention loss. Table \ref{tab:ablation_main} examines the modality relevance mask ($\tilde{\omega}$) and the attention loss ($\mathcal{L}_{\text{attn}}$) in the preliminary LoRA experiments, and compares full MAPO with GRPO under full fine-tuning.

\input{tables/ablation}

With LoRA, the modality relevance mask raises the four-benchmark average from 69.48 for GRPO to 70.46. Incorporating the auxiliary attention loss ($\mathcal{L}_{\text{attn}}$) further improves the average to 71.39, with gains on every benchmark. This advantage over GRPO is also present under full fine-tuning: MAPO achieves 73.25 compared with 70.95 for GRPO, a gain of 2.30 points. The preliminary full-fine-tuning results appear in Appendix \ref{sec:phase1_results}, followed by analyses of training and attention dynamics under varying penalty weights ($\eta$).

\FloatBarrier
\subsection{Main results}

We next evaluate MAPO against representative multimodal and audio-reasoning systems. Figures \ref{fig:modality_benchmarks} and \ref{fig:capability_benchmarks} summarize performance on modality-centric (MMAU~\citep{sakshi2024mmaumassivemultitaskaudio}, MMAR~\citep{ma2025mmarchallengingbenchmarkdeep}) and capability-centric (MMSU~\citep{wang2026mmsumassivemultitaskspoken}, MMAU-Pro~\citep{kumar2025mmauprochallengingcomprehensivebenchmark}) benchmarks, respectively.

\input{figures/modality_results}

\textbf{Modality-Centric Performance.} MAPO achieves competitive performance on modality-centric benchmarks (Figure \ref{fig:modality_benchmarks}), with scores of 77.83 on MMAU and 70.83 on MMAR. These improve on Qwen3-Omni-Thinking by 2.83 and 3.93 points, respectively. Its MMAU score is comparable to Nova 2 Omni (77.80), while Audio-DeepThinker~\citep{audiodeepthinker} reports higher scores on MMAU and MMAR (78.50 and 74.00).

\textbf{Capability and Instruction Following.} The improvements over the base model extend to capability-centric evaluations (Figure \ref{fig:capability_benchmarks}). On MMSU, MAPO reaches 79.35, compared with 76.30 for Qwen3-Omni-Thinking, 77.26 for Audio-DeepThinker, and 75.90 for Step-Audio-R1~\citep{stepaudior1}. On MMAU-Pro, the overall score increases from 62.63 to 64.98. The category-level comparisons in Appendix \ref{sec:detailed_benchmarks} show improvements in both MMSU perception and reasoning, as well as MMAU-Pro audio instruction following. Qualitative examples in Appendix \ref{sec:qual_examples} illustrate the model's reasoning behavior.

\input{figures/capability_results}

\FloatBarrier
\section{Conclusion}

In this work, we introduced Modality-Aware Policy Optimization, a novel reinforcement learning framework designed to mitigate the critical issue of late-stage modality collapse in long-form cross-modal reasoning. By synergizing a dynamic modality relevance mask driven by a cross-modal differential entropy with a targeted attention loss branch, MAPO structurally breaks the dominance of the language prior and enforces sustained grounding in the uncompressed source signal. Our empirical evaluations on complex audio reasoning tasks demonstrate that this dual-mechanism approach significantly improves both reasoning fidelity and instruction adherence. Crucially, while our contributions and empirical validations are focused entirely within the acoustic domain, MAPO's foundational token-reweighting and attention-steering mechanisms rely solely on native attention weights and predictive entropy rather than modality-specific inductive biases. As such, exploring the application of this framework to other rich, continuous non-text modalities, such as vision and video, represents an exciting avenue for future work in robust omni-modal post-training.

\label{sec:main_text_end}
\clearpage
\input{sections/statements}
\bibliographystyle{iclr2027_conference}
\bibliography{refs}
\clearpage
\input{sections/neurips_appendix}

%% file: tables/ablation.tex
% Generated by scripts/make_table_ablation.py; do not edit by hand.
\begin{table}[!htb]
\centering
\caption{Component analysis of MAPO. MAPO uses both the modality relevance mask and attention loss unless marked mask-only. Scores are mean $\pm$ sample standard deviation across three seeds. Avg. is the unweighted mean over the four benchmarks.}
\label{tab:ablation_main}
\setlength{\tabcolsep}{4pt}
\begin{tabular}{@{}p{0.22\linewidth}rrrrr@{}}
\toprule
\textbf{Method} & \textbf{MMAU} & \textbf{MMAR} & \textbf{MMSU} & \textbf{MMAU-Pro} & \textbf{Avg.} \\
\midrule
\multicolumn{6}{l}{\textit{LoRA}} \\
GRPO & 73.70 $\pm$ 0.46 & 66.57 $\pm$ 0.35 & 76.61 $\pm$ 0.17 & 61.05 $\pm$ 0.27 & 69.48 $\pm$ 0.14 \\
MAPO mask-only & 75.37 $\pm$ 0.32 & 67.03 $\pm$ 0.15 & 76.55 $\pm$ 0.34 & 62.88 $\pm$ 0.38 & 70.46 $\pm$ 0.10 \\
MAPO & 76.40 $\pm$ 0.20 & 68.40 $\pm$ 0.26 & 77.27 $\pm$ 0.10 & 63.50 $\pm$ 0.04 & 71.39 $\pm$ 0.13 \\
\midrule
\multicolumn{6}{l}{\textit{Full fine-tuning}} \\
GRPO & 75.53 $\pm$ 0.55 & 68.43 $\pm$ 0.25 & 76.48 $\pm$ 0.26 & 63.35 $\pm$ 0.10 & 70.95 $\pm$ 0.11 \\
MAPO & \textbf{77.83} $\pm$ 0.25 & \textbf{70.83} $\pm$ 0.21 & \textbf{79.35} $\pm$ 0.15 & \textbf{64.98} $\pm$ 0.28 & \textbf{73.25} $\pm$ 0.09 \\
\bottomrule
\end{tabular}
\end{table}

%% file: figures/modality_results.tex
% Generated by scripts/make_figure_benchmark_pairs.py; do not edit by hand.
\begin{figure}[!htb]
\centering
\includegraphics[width=\linewidth]{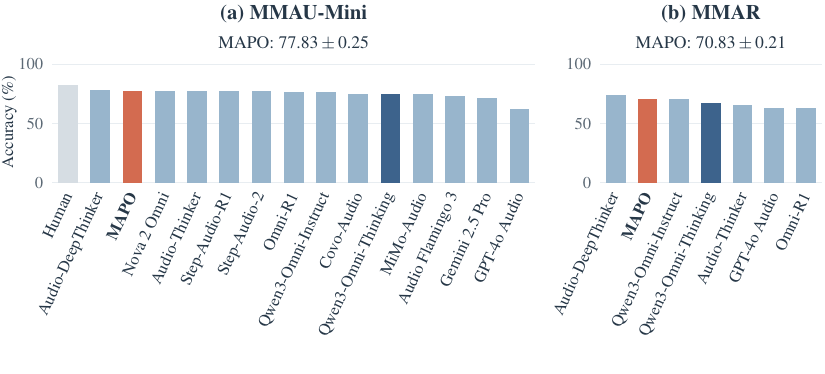}
\caption{Accuracy (\%) on MMAU and MMAR. MAPO (coral) is annotated with mean $\pm$ sample SD across three seeds; dark blue marks its base model and gray denotes human performance. Each panel shows reported results, ordered by accuracy; other scores are point estimates. Exact scores and category-level results are in Appendix~\ref{sec:detailed_benchmarks}.}
\label{fig:modality_benchmarks}
\end{figure}

%% file: figures/capability_results.tex
% Generated by scripts/make_figure_benchmark_pairs.py; do not edit by hand.
\begin{figure}[!htb]
\centering
\includegraphics[width=\linewidth]{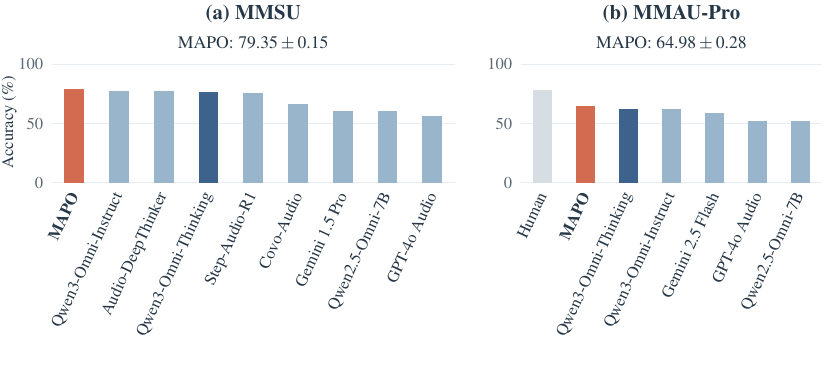}
\caption{Accuracy (\%) on MMSU and MMAU-Pro. MAPO reports mean $\pm$ sample SD across three seeds, while other scores are point estimates. Exact scores and category-level results are in Appendix~\ref{sec:detailed_benchmarks}.}
\label{fig:capability_benchmarks}
\end{figure}

%% file: sections/statements.tex
\section*{AI use statement}
\label{sec:statements_start}

Generative AI tools assisted with manuscript preparation, LaTeX formatting, and code and document checks. The research pipeline also uses a language model as a reasoning-consistency reward, as described in the experimental setup. The authors are responsible for the final text, claims, and artifacts.
% Authors should confirm that this disclosure covers all project-wide AI uses.

%% file: sections/neurips_appendix.tex
\appendix
\section{Gradient analysis}
\label{sec:appendix_gradient}

We analyze how the modality relevance mask changes the coefficients of token-level policy gradients and how the attention loss supplies an additional gradient through audio attention mass. During each optimization step, the sampled completions, sequence-level advantages, and detached weighting factors are treated as fixed.

Recall the total MAPO objective:
\begin{equation}
    \mathcal{L}_{\text{MAPO}} = \mathcal{L}_{\text{PG}}(\tilde{\omega}) + \beta \mathcal{L}_{\text{KL}} + \eta \mathcal{L}_{\text{attn}}
\end{equation}

Since the total gradient $\nabla_\theta \mathcal{L}_{\text{MAPO}}$ is a linear combination of its components, we dissect the gradient contributions of the reweighted policy branch and the auxiliary attention branch independently.

\subsection{Gradient of the reweighted policy objective}

The MAPO reweighted policy gradient modifies the standard GRPO surrogate objective by applying the modality relevance mask $\tilde{\omega}_t$. Crucially, as stated in Section \ref{sec:mrm}, the mask $\tilde{\omega}_t$ is computed via a text-only reference model and the live policy's forward pass, but it is explicitly \textit{detached} from the computation graph during backpropagation.

Let $\bar{\omega}_t^{(g)} = \frac{\tilde{\omega}_t^{(g)}}{\sum_{t'=1}^{T_g} \tilde{\omega}_{t'}^{(g)}}$ represent the normalized token-level weight. For a single completion sequence $y \sim \pi_{\text{old}}(\cdot \mid x)$ of length $T$, the reweighted surrogate loss is:
\begin{equation}
    \mathcal{L}_{\text{PG}}(\tilde{\omega}) = \sum_{t=1}^{T} \bar{\omega}_t \Big[ -\min\!\Big(\rho_t \hat{A}, \;\text{clip}(\rho_t, 1{-}\epsilon, 1{+}\epsilon)\,\hat{A}\Big) \Big]
\end{equation}

By detaching $\bar{\omega}_t$, the gradient operator $\nabla_\theta$ only acts on the probability ratio $\rho_t = \frac{\pi_\theta(y_t \mid y_{<t}, x)}{\pi_{\text{old}}(y_t \mid y_{<t}, x)}$. In the unclipped branch of the surrogate objective, the gradient of the surrogate objective at step $t$ simplifies to:
\begin{equation}
    \nabla_\theta \ell_t^{\text{PG}} = -\hat{A} \nabla_\theta \rho_t = -\hat{A} \frac{\nabla_\theta \pi_\theta(y_t \mid y_{<t}, x)}{\pi_{\text{old}}(y_t \mid y_{<t}, x)} = -\hat{A} \rho_t \nabla_\theta \log \pi_\theta(y_t \mid y_{<t}, x)
\end{equation}

Because the token weights are detached, the gradient for a sampled completion is:
\begin{equation}
    \nabla_\theta \mathcal{L}_{\text{PG}}(\tilde{\omega}) = \sum_{t=1}^{T} \bar{\omega}_t \nabla_\theta \ell_t^{\text{PG}}
\end{equation}

\textbf{Effect of token reweighting:} Standard GRPO uses the coefficient $1/T$, whereas MAPO uses the normalized modality relevance mask $\bar{\omega}_t$. Before clipping, larger $|\Delta h_t|$ produces larger token weights. If all scores are equal, including when all are zero, the normalized weights are exactly $1/T$, recovering uniform GRPO weighting. For finite scores and a positive clipping cap, every valid token retains positive weight. The resulting gradient depends on both these coefficients and the policy gradient at each token.

\subsection{Gradient of the attention loss branch}

The attention loss is computed directly from audio attention mass. It therefore provides an additional training signal that encourages the model to maintain attention to the source audio at the selected tokens.

For a single sequence, define the detached token weight
\begin{equation}
    \Gamma_t = \frac{\min\!\left(\hat{f}\,|\hat{A}|, C_{\mathrm{pref}}\right)}{N_{\mathrm{pos}}} (t/T)^\kappa \tilde{\nu}_t.
\end{equation}
The attention loss is
\begin{equation}
    \mathcal{L}_{\text{attn}} = \sum_{t=1}^{T} \Gamma_t \big[-\log(a_t + \varepsilon)\big].
\end{equation}
Its parameter-dependent term is the audio attention mass $a_t$. Applying the chain rule gives:
\begin{equation}
    \nabla_\theta \mathcal{L}_{\text{attn}} = \sum_{t=1}^{T} \Gamma_t \left( \frac{-1}{a_t + \varepsilon} \right) \nabla_\theta a_t
\end{equation}

Recall that $a_t = \frac{1}{|L_{\text{tgt}}|}\sum_{l \in L_{\text{tgt}}} m_t^{(l, \text{max})}$, where $m_t^{(l, \text{max})} = \max_h \sum_{j \in \mathcal{S}_{\text{audio}}} \alpha_{t,j}^{(l,h)}$. Because the maximum function is continuous but not universally differentiable, we rely on its subgradient. Let $h^*(t, l) = \arg\max_h \sum_{j \in \mathcal{S}_{\text{audio}}} \alpha_{t,j}^{(l,h)}$ denote the attention head that routes the maximum audio mass at step $t$ and layer $l$. The gradient routes exclusively through this ``specialized'' head:
\begin{equation}
    \nabla_\theta a_t = \frac{1}{|L_{\text{tgt}}|} \sum_{l \in L_{\text{tgt}}} \sum_{j \in \mathcal{S}_{\text{audio}}} \nabla_\theta \alpha_{t,j}^{(l, h^*(t,l))}
\end{equation}

To understand how this directly updates the key ($W_K$) and query ($W_Q$) projection matrices, we examine the Jacobian of the softmax attention mechanism. Let $z_{t,i} = \frac{(W_Q x_t) \cdot (W_K x_i)}{\sqrt{d}}$ be the unnormalized pre-softmax logit for the query at $t$ and key at $i$. The attention weight is $\alpha_{t,j} = \frac{\exp(z_{t,j})}{\sum_i \exp(z_{t,i})}$. The derivative of $\alpha_{t,j}$ with respect to the logit $z_{t,i}$ is:
\begin{equation}
    \frac{\partial \alpha_{t,j}}{\partial z_{t,i}} = \alpha_{t,j} (\delta_{ji} - \alpha_{t,i})
\end{equation}
where $\delta_{ji}$ is the Kronecker delta.

Substituting this into our subgradient derivation for the target layer and target head, the gradient with respect to a specific logit $z_{t,i}$ is:
\begin{equation}
\begin{aligned}
    \frac{\partial}{\partial z_{t,i}} \left( \sum_{j \in \mathcal{S}_{\text{audio}}} \alpha_{t,j} \right) &= \sum_{j \in \mathcal{S}_{\text{audio}}} \alpha_{t,j} (\delta_{ji} - \alpha_{t,i}) \\
    &= \begin{cases}
        \alpha_{t,i} (1 - m_t) & \text{if } i \in \mathcal{S}_{\text{audio}} \\
        -\alpha_{t,i} m_t & \text{if } i \notin \mathcal{S}_{\text{audio}} (\text{e.g., text context})
    \end{cases}
\end{aligned}
\end{equation}
where $m_t = \sum_{j \in \mathcal{S}_{\text{audio}}} \alpha_{t,j}$ is the total audio mass for that head.

For a uniquely selected maximizing head, holding attention logits at other positions and layers fixed, the direct derivative is:
\begin{equation}
\label{eq:attention_update}
    \frac{\partial \mathcal{L}_{\text{attn}}}{\partial z_{t,i}^{(l, h^*)}} = \underbrace{\left( \frac{-\Gamma_t}{|L_{\text{tgt}}|(a_t + \varepsilon)} \right)}_{\text{Non-positive scaling factor}} \times 
    \begin{cases}
        \alpha_{t,i}^{(l,h^*)} (1 - m_t^{(l,h^*)}) & \text{if } i \in \mathcal{S}_{\text{audio}} \\
        -\alpha_{t,i}^{(l,h^*)} m_t^{(l,h^*)} & \text{if } i \notin \mathcal{S}_{\text{audio}}
    \end{cases}
\end{equation}

\textbf{Gradient interpretation:} Equation \ref{eq:attention_update} describes the attention branch's direct contribution at the selected head. The derivative is non-positive for audio logits and non-negative for other visible logits, so gradient descent in these coordinates favors greater audio attention. The detached factor $\Gamma_t$ controls this pressure through the POS gate, inverse relevance weight, temporal weight, and capped sequence-level prefactor. The network update combines this contribution with gradients through shared representations and the policy and KL objectives.

\section{Reasoning consistency evaluator prompt}
\label{sec:appendix_prompt}

To compute the reasoning consistency reward during reinforcement post-training, we query our frozen evaluator model, Qwen3-30B-A3B-Instruct, using the following template. This prompt is explicitly designed to isolate logical contradictions and incomplete thoughts without conflating them with the factual correctness of the answer itself, providing a pure signal for chain-of-thought coherence.

\begin{tcolorbox}[
    colback=blue!5!white,       % Light blueish background
    colframe=blue!50!gray,
    opacityback=1.0,            % Slight background transparency
    opacityframe=0.6,           % Very slight frame transparency
    title=Reasoning Consistency Evaluator Prompt,
    fonttitle=\bfseries,        % Bold title font
    boxrule=1pt,                 % Thinned the border slightly so it's less heavy
    arc=4pt,                     % Slightly rounder corners for a softer feel
    left=8pt, right=8pt, top=8pt, bottom=8pt % Added a tiny bit more breathing room
]

You are a reasoning consistency evaluator. Given a model's thinking process 
and its final answer, your task is to evaluate it against two specific 
failure modes.

\vspace{0.5em}
\noindent \textbf{Thinking Process:}\\
\texttt{\{think\_text\}}

\vspace{0.5em}
\noindent \textbf{Final Answer:}\\
\texttt{\{answer\_text\}}

\vspace{0.5em}
\noindent Respond \textbf{``NO''} ONLY if at least one of the following is true:
\begin{enumerate}
    \item The conclusion reached in the thinking process does not agree with 
          or contradicts the final answer.
    \item The thinking process is visibly incomplete or cut off prematurely.
\end{enumerate}

\noindent If neither of these failure modes is present, respond \textbf{``YES''}. 
Only output YES or NO.

\end{tcolorbox}

\section{Ablation analysis}
\label{sec:ablation}

\subsection{Preliminary experiments on AVQA}
\label{sec:phase1_results}

Table \ref{tab:phase1_results} complements the preliminary LoRA analysis in Table \ref{tab:ablation_main} with full-fine-tuning results after training on AVQA.

\input{tables/phase1}

MAPO outperforms GRPO on all four benchmarks in the preliminary experiments, with averages of 71.88 and 70.94, respectively. The main MAPO model improves further on every benchmark, reaching an average of 73.25 (Table \ref{tab:ablation_main}).

\subsection{General training dynamics}
\label{sec:training_dynamics}

\begin{figure}[ht]
    \centering
    \begin{subfigure}{0.48\textwidth}
        \centering
        \includegraphics[width=\linewidth]{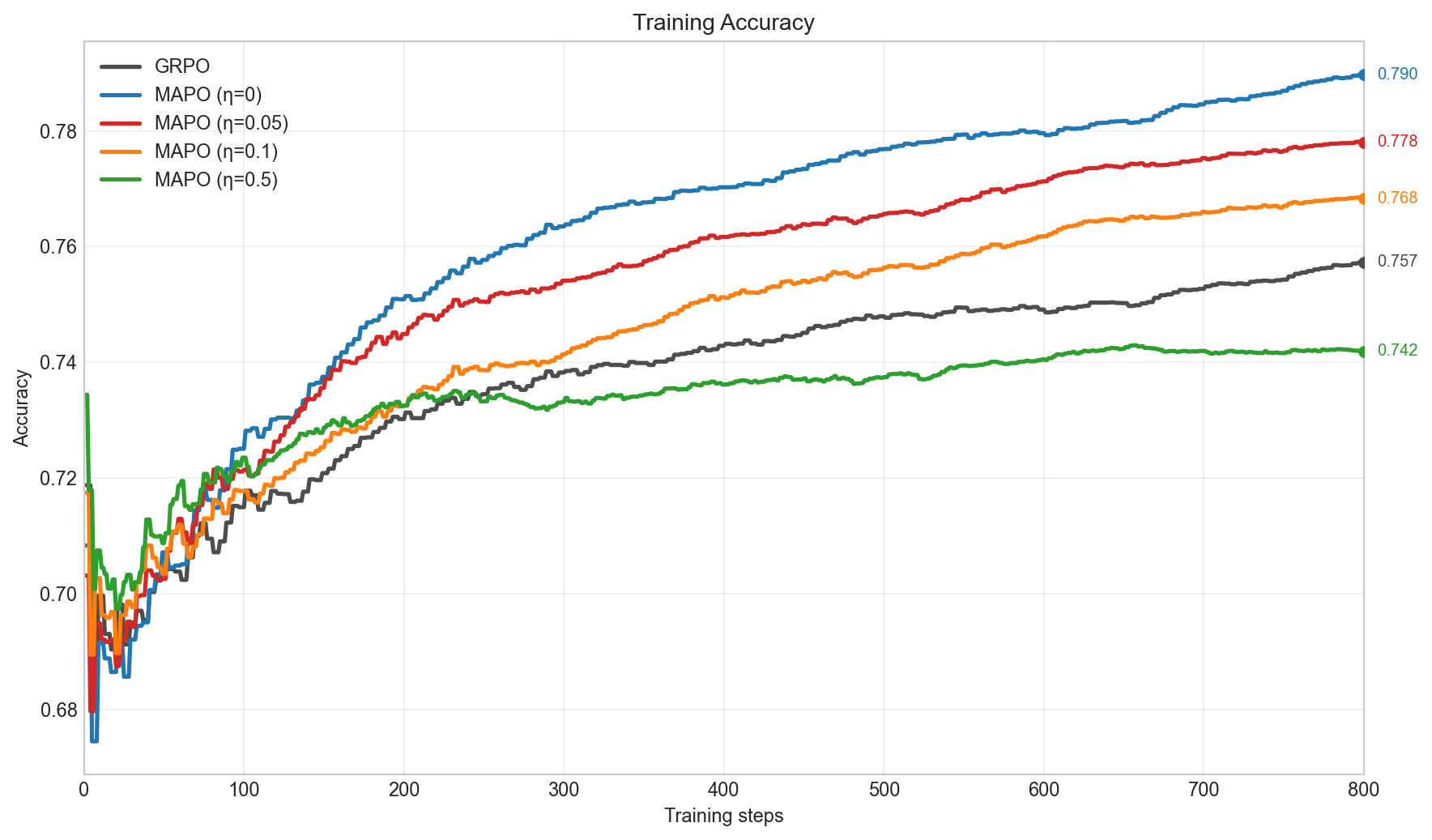}
        \caption{Training Accuracy}
        \label{fig:acc_sub}
    \end{subfigure}
    \hfill
    \begin{subfigure}{0.48\textwidth}
        \centering
        \includegraphics[width=\linewidth]{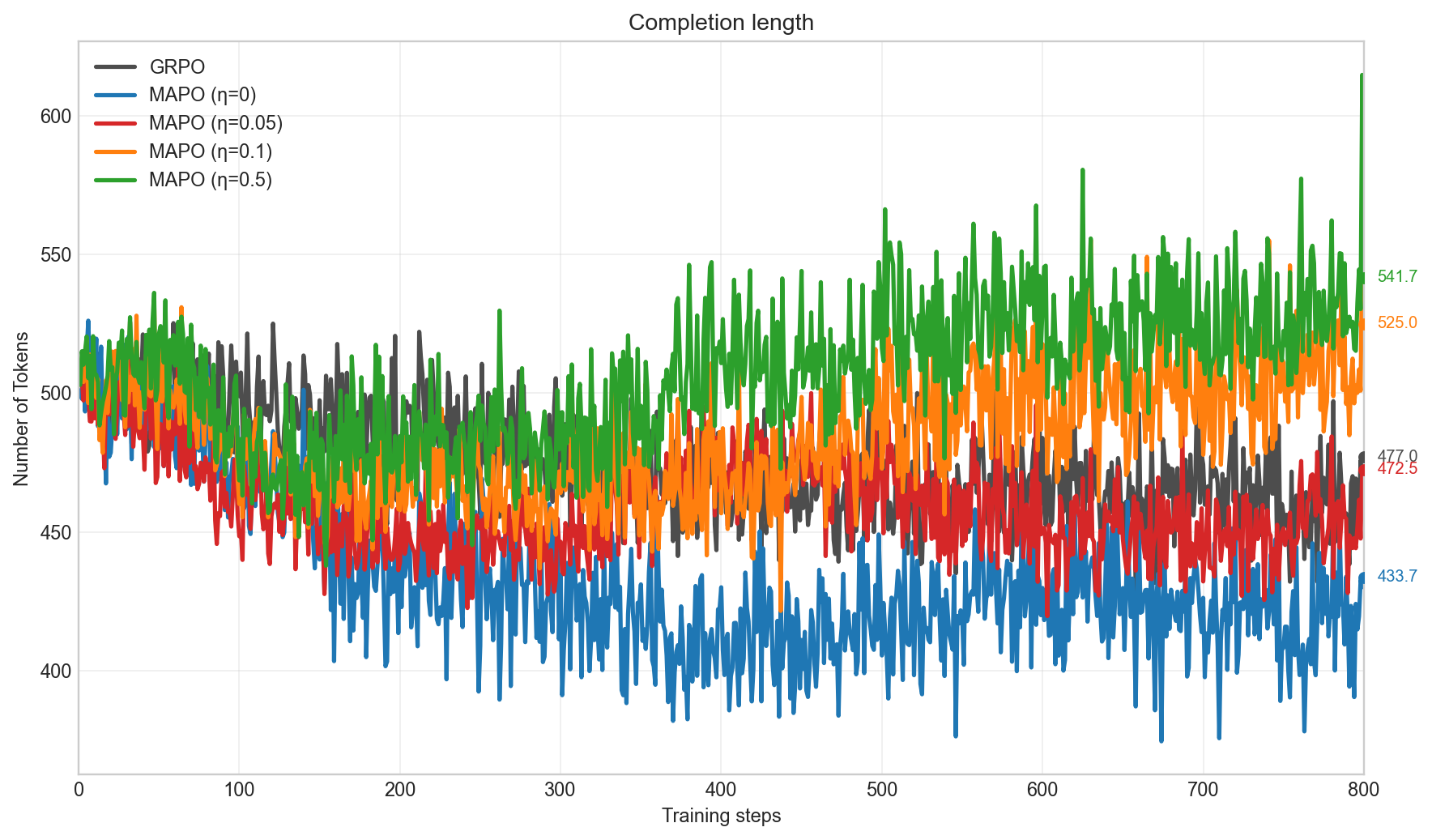}
        \caption{Completion Length}
        \label{fig:len_sub}
    \end{subfigure}
    \caption{Training dynamics over 800 steps, comparing standard GRPO with MAPO across various attention loss weights ($\eta$).}
    \label{fig:training_dynamics}
\end{figure}

To isolate the impact of modality-aware token reweighting and the attention loss branch, we compare the GRPO baseline against MAPO with different attention loss weights ($\eta$). Figure \ref{fig:training_dynamics} displays reward accuracy and completion length over 800 steps.

Standard GRPO, constrained by its uniform $1/T$ token weighting, dilutes the learning signal and reaches an accuracy of 0.757, yielding an average completion length of 477.0 tokens. By contrast, applying only the modality relevance mask ($\eta=0$) dynamically redirects the optimization budget toward modality-critical tokens, which significantly improves sample efficiency and drives accuracy to a peak of 0.790. However, optimizing strictly for the task reward in this manner causes a drop in the average completion length (433.7 tokens). This indicates a tendency for the model to shortcut its reasoning trajectory, effectively bypassing the extended cross-modal synthesis necessary for robust, generalized problem-solving.

As illustrated in Figure \ref{fig:len_sub}, increasing the attention penalty weight ($\eta > 0$) effectively forces the model to sustain longer, deeply grounded reasoning traces. For instance, setting $\eta=0.05$ recovers the completion length to a healthy 472.5 tokens while preserving a highly competitive training accuracy of 0.778. Pushing $\eta$ even higher (to 0.1 and 0.5) extends the generation significantly to 525.0 and 541.7 tokens, respectively. However, as $\eta$ increases, we observe a corresponding, incremental degradation in raw training accuracy (dropping to 0.768 and 0.742). 

This trade-off occurs because the auxiliary attention penalty acts as an additional, competing optimization objective. Forcing the model's internal attention distributions to persistently anchor to the source signal naturally shifts gradient capacity away from strictly maximizing the primary training task reward. Crucially, however, this slowed training accuracy optimization acts as an effective regularizer. By preventing the model from rapidly overfitting to the surface-level formatting and simple task rewards of the training set via shortcut reasoning (as seen when $\eta=0$), the attention penalty preserves and enforces the model's generalization capabilities. This regularization effect is consistent with the held-out benchmark results in Table \ref{tab:ablation_main}, where adding the attention loss ($\eta > 0$) raises the LoRA average from 70.46 to 71.39 compared with using the modality relevance mask alone ($\eta=0$). Consequently, carefully calibrating $\eta$ strikes the optimal balance, trading minor training reward inflation for robust, sustained cross-modal grounding and thus better downstream generalization.

\subsection{Audio attention dynamics}
\label{sec:appendix_audio_attn_ablation}

\begin{figure}[!htbp]
    \centering
    \includegraphics[width=0.88\textwidth]{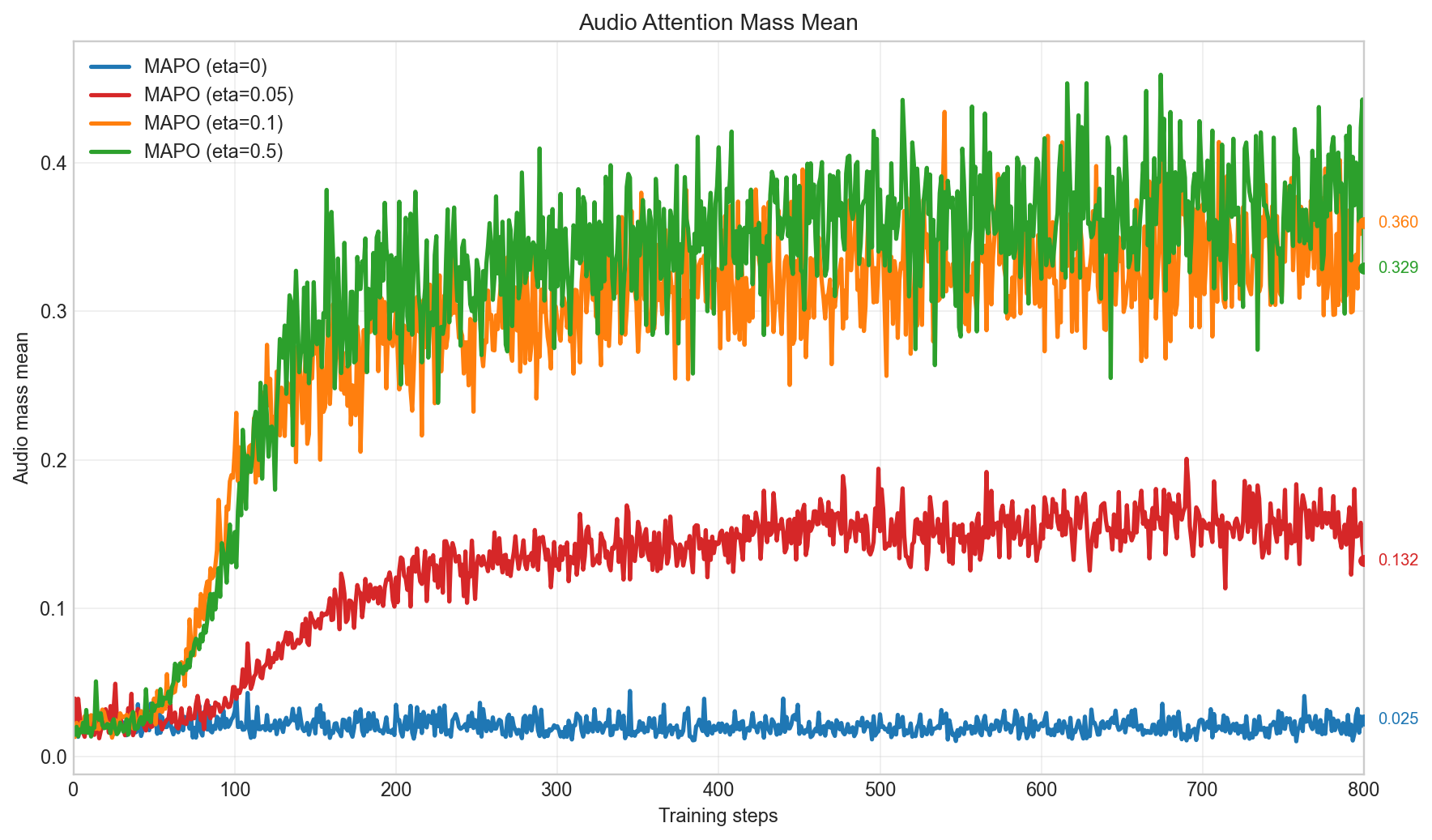}
    \caption{Mean audio attention mass over 800 training steps for different attention loss weights ($\eta$). Nonzero weights increase the attention allocated to audio; the largest weight does not yield the highest final mass.}
    \label{fig:audio_attn_ablation}
\end{figure}

We next examine how the attention loss weight $\eta$ affects audio attention during post-training. Figure~\ref{fig:audio_attn_ablation} tracks mean audio attention mass over 800 steps, complementing the reward and completion-length curves above.

With no attention penalty ($\eta=0$), mean audio attention remains near 0.025. A weight of 0.05 produces a gradual increase to approximately 0.132. Larger weights change the trajectory more rapidly: most of the increase occurs within the first 150 to 200 steps, and the curves settle near 0.360 for $\eta=0.1$ and 0.329 for $\eta=0.5$. Together with the lower training accuracy and longer completions at larger weights, this pattern shows that increasing $\eta$ changes the balance between the task objective and attention allocation, with diminishing returns in the final attention mass.

The direction of these changes is consistent with the gradient analysis in Appendix~\ref{sec:appendix_gradient}, where $\eta$ scales the contribution of the attention branch. Appendix~\ref{sec:audio_grounding} examines how the resulting attention patterns appear during reasoning and how predictions respond to changes in the source audio.

\subsection{Impact of the attention loss branch}

\begin{figure}[!htbp]
    \centering
    \includegraphics[width=\textwidth]{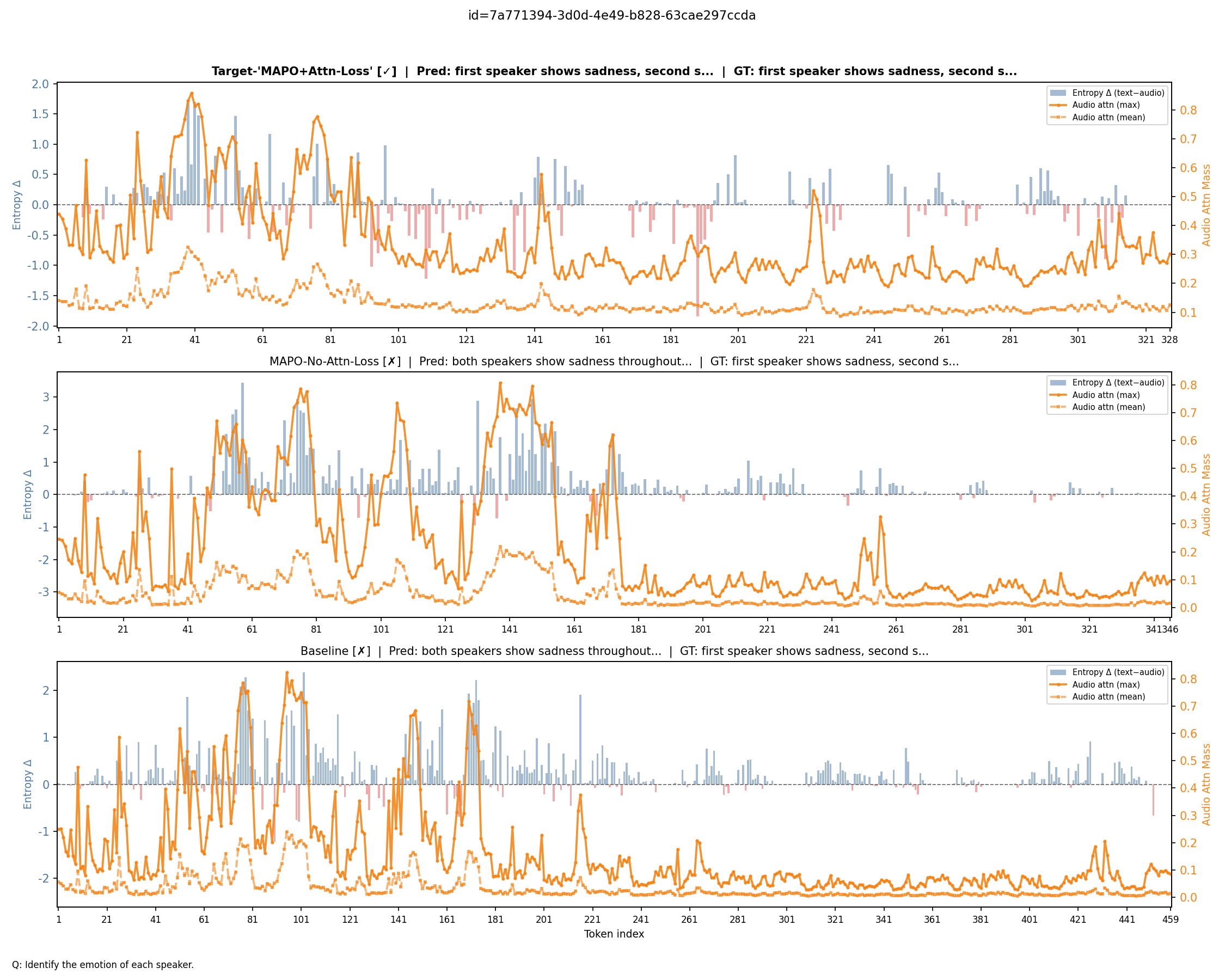}
    \caption{Attention and differential entropy in an illustrative reasoning example. Blue bars show $\Delta h_t$; orange curves show max-head (solid) and mean-head (dashed) audio attention. Base Thinking and mask-only MAPO show declining attention later in the response, whereas full MAPO retains more audio attention and fluctuations in differential entropy.}
    \label{fig:attention_ablation}
\end{figure}

Figure~\ref{fig:attention_ablation} compares Qwen3-Omni-Thinking, MAPO with only the modality relevance mask, and full MAPO on the same question. In the Base and mask-only responses, both attention reductions decline midway through generation, and the models reach the same incorrect conclusion that ``both speakers show sadness.'' Full MAPO maintains more audio attention later in its response and changes the final answer.

The cross-modal differential entropy traces provide a second view of this example. For each analyzed model, $\Delta h_t$ compares its predictive entropy with and without source audio at the same fixed checkpoint, holding the text prompt and generated prefix fixed. Its positive and negative excursions indicate differences in predictive uncertainty; full MAPO retains these fluctuations later in the response. The example therefore illustrates concurrent changes in attention, predictive uncertainty, and the answer. The stage-aligned analysis in Appendix~\ref{sec:audio_grounding} examines how the attention patterns extend across examples, followed by interventions on the audio input.

\FloatBarrier

\section{Detailed benchmark results}
\label{sec:detailed_benchmarks}

Tables \ref{tab:modality_benchmarks} and \ref{tab:capability_benchmarks} provide the exact overall scores shown in Figures \ref{fig:modality_benchmarks} and \ref{fig:capability_benchmarks}, respectively. Tables \ref{tab:modality_details} and \ref{tab:capability_details} give category-level point estimates, complementing the three-seed overall MAPO results.

\input{tables/modality_main}
\input{tables/capability_main}
\input{tables/modality_details}
\input{tables/capability_details}

The MMSU breakdown shows improvements over Qwen3-Omni-Thinking in both perception (74.19 to 76.36) and reasoning (78.55 to 82.56). On MMAU-Pro, the largest gain is in audio instruction following, which increases from 62.07 to 95.40. Closed- and open-ended question answering also improve, from 59.63 to 62.00 and from 84.77 to 85.32, respectively.

\input{sections/audio_grounding}

\FloatBarrier
\section{Qualitative examples}
\label{sec:qual_examples}

\subsection{Example \#1}
\input{appendix/exp_1}

\subsection{Example \#2}
\input{appendix/exp_2}

\subsection{Qualitative analysis of the examples}

The two examples illustrate different ways in which the models interpret ambiguous acoustic cues. We focus on the distinctions expressed in their responses, complementing the aggregate measurements in Appendix~\ref{sec:audio_grounding}.

\paragraph{Example \#1: Factory machinery.}
The correct answer is \textit{Factory machinery}, while Base Thinking predicts \textit{Wind turbine}. The baseline describes a continuous, low-frequency mechanical sound in terms of rushing or whooshing, then connects that description to a rotating object producing airflow. Both MAPO variants instead select factory machinery. The mask-only response emphasizes mechanical density and industrial motion, while the full MAPO response draws a more explicit contrast between the harsh mechanical sound and the smoother whoosh expected from a wind turbine. In this case, both variants answer correctly, and the full model gives a more specific comparison between the candidate interpretations.

\paragraph{Example \#2: Horse-drawn wagon.}
The correct answer is \textit{Horse-drawn wagon}, but Base Thinking and mask-only MAPO select \textit{Train}. Their responses associate rhythmic clattering with rail joints and steel wheels. Full MAPO instead describes hoof-like, irregular, and wagon-related sounds, using these distinctions to reject the train interpretation. Unlike the first example, the change in the explanation accompanies a change from an incorrect to a correct answer.

\paragraph{Overall interpretation.}
These cases are consistent with the complementary roles of the two branches: the mask emphasizes tokens whose predictions differ from the text-only reference, while the attention loss encourages continued access to the source. The examples show how the resulting responses can differ, rather than isolating the cause of an individual correction. The held-out branch comparison in Table~\ref{tab:ablation_main} and the paired input interventions in Appendix~\ref{sec:audio_grounding} provide the broader empirical context.

\section{Part-of-Speech gating}
\label{sec:appendix_pos}

\subsection{Token categories and implementation}

\begin{figure}[ht]
    \centering
    \includegraphics[width=\textwidth]{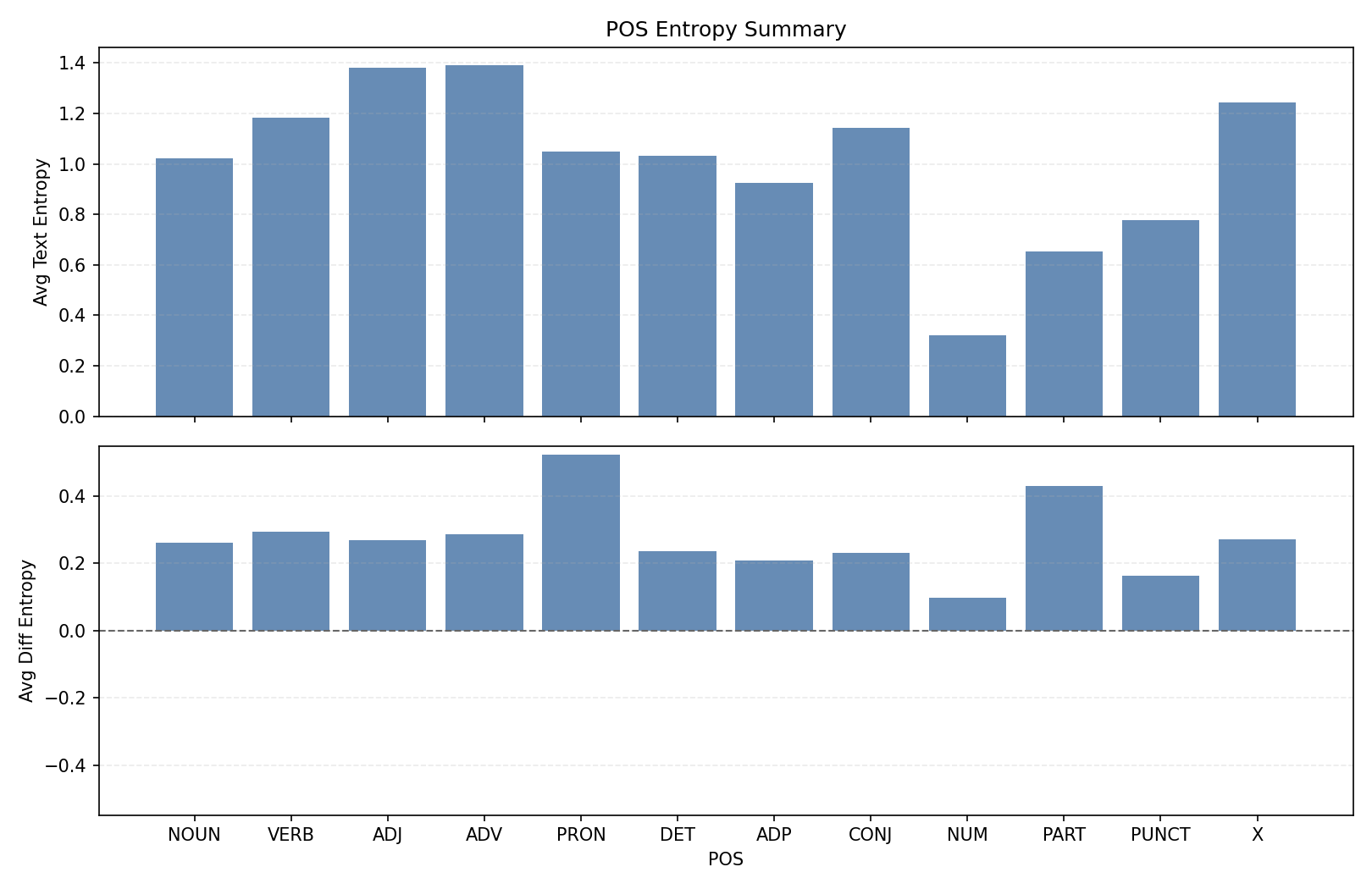}
    \caption{Entropy by POS category across reasoning trajectories. The top panel shows average text-only predictive entropy; the bottom panel shows average cross-modal differential entropy.}
    \label{fig:pos_entropy}
\end{figure}

The POS gate selects which generated tokens receive the auxiliary attention penalty. To motivate this selection, we examine text-only predictive uncertainty and cross-modal differential entropy across POS categories in Figure~\ref{fig:pos_entropy}.

Text-only entropy measures uncertainty in predicting the next token from the prompt and generated prefix without audio. Differential entropy compares that uncertainty with the multimodal policy. The selected categories, \texttt{\{NOUN, VERB, ADJ, ADV, NUM, X\}}, cover substantive content, descriptions, quantities, and unclassified tokens such as onomatopoeic expressions. The gate uses these linguistic roles to focus the penalty on content-bearing tokens. The ablation below evaluates the effect of that choice.

\textbf{Implementation Details.} To facilitate high-throughput training, POS gating is computed dynamically on the CPU once per completed rollout prior to batch collation. Specifically, generated subword tokens are decoded, normalized, and merged into word units (masking out punctuation), which are then tagged using NLTK's averaged-perceptron tagger (\texttt{nltk.pos\_tag}) mapped to our coarse target categories. These word-level binary POS decisions are subsequently projected back to their constituent subword indices to align with the sequence completion mask, defaulting to an all-ones fallback mask if alignment fails or the sequence is degenerate. As this amortized tagging pass occurs strictly outside the rollout loop, it avoids per-token synchronization overhead entirely, introducing negligible latency relative to generation and model forward/backward passes.
\subsection{Effect of the POS gate}

The distributional analysis motivates restricting the attention penalty to substantive tokens. Table~\ref{tab:pos_gate} tests this choice directly by comparing full MAPO with a variant that removes the POS gate while holding the remaining training and evaluation settings fixed.

\input{tables/pos_gate}

Retaining the gate improves the mean on every benchmark and raises the four-benchmark average from $72.39\pm0.22$ to $73.25\pm0.09$. This observation supports concentrating the attention penalty on the selected token categories, rather than applying it uniformly across the generated response.

\FloatBarrier

\section{Hyperparameters and computational cost}
\label{sec:appendix_hyperparameters}

Table \ref{tab:hyperparameters} details the primary hyperparameters explored during the reinforcement post-training phases.

\begin{table}[ht]
\centering
\caption{Hyperparameter settings for the MAPO framework, including the explored tuning ranges and the recommended values used in the main experiments.}
\label{tab:hyperparameters}
\renewcommand{\arraystretch}{1.2} % Slightly increases row height for readability
\setlength{\tabcolsep}{3pt}       % Adjusts horizontal padding
\begin{tabular}{lccc}
\toprule
\textbf{Parameter Name} & \textbf{Symbol} & \textbf{Range} & \textbf{Recommended Value} \\
\midrule
\multicolumn{4}{c}{\textit{MAPO Core Mechanisms}} \\
\midrule
Attention Loss Weight & $\eta$ & $[0.01, 0.5]$ & $0.1$ \\
Base Mask Temperature & $\tau_{\text{base}}$ & $[0.5, 2.0]$ & $1.0$ \\
Temporal Weight Exponent & $\kappa$ & $[0.5, 2.0]$ & $1.0$ \\
Advantage Floor / Stability & $\varepsilon$ & $[0, 0.1]$ & $0.1$ \\
Mask Clip Threshold & $C_{\mathrm{mask}}$ & $[2.0, 5.0]$ & $5.0$ \\
Attention Prefactor Clip & $C_{\mathrm{pref}}$ & $[3.0, 5.0]$ & $3.0$ \\
Target Attention Layers & $L_{\text{tgt}}$ & - & $\{43, 44, 45, 46, 47\}$ \\
\midrule
\multicolumn{4}{c}{\textit{GRPO Objective \& Training Scale}} \\
\midrule
KL Penalty Coefficient & $\beta$ & $[0.001, 0.02]$ & $0.002$ \\
Surrogate Clip Ratio & $\epsilon$ & - & $0.2$ \\
Sampled Group Size & $G$ & $\{4, 8\}$ & $8$ \\
Generation Batch Size & - & - & $768$ \\
Steps per Generation & - & - & $4$ \\
\midrule
\multicolumn{4}{c}{\textit{Decoding Parameters (Rollout Server)}} \\
\midrule
Temperature & - & $[1.0, 1.2]$ & $1.2$ \\
Top-$k$ & - & $\{40, 50\}$ & $50$ \\
Top-$p$ & - & $[0.90, 0.99]$ & $0.99$ \\
Repetition Penalty & - & $[1.0, 1.2]$ & $1.0$ \\
\bottomrule
\end{tabular}
\end{table}

\subsection{Training and setup}
\label{sec:training_reference}

We use Adam with $\beta_1=0.9$, $\beta_2=0.95$, numerical epsilon $10^{-8}$, weight decay 0.1, and gradient clipping at 1.0. The peak learning rate is $10^{-6}$, with 5\% warmup and cosine decay to zero. The task-failure indicator is one when the strict multiple-choice accuracy reward is non-finite or at most zero. Its effective gate is floored at 0.1, giving incorrect trajectories weight 1 and correct trajectories weight 0.1.

The text-only reference receives the same prompt and generated prefix as the live policy, including any caption already generated, but no source audio. We reuse the frozen reference already maintained for the GRPO KL penalty to compute text-only entropy, avoiding an additional model copy dedicated to this computation. Under LoRA, the frozen backbone with the policy adapters disabled provides this reference. Both reference paths run without gradient tracking. The selected attention layers are 43--47, the final five of the 48 decoder layers. The ranges in Table~\ref{tab:hyperparameters} summarize preliminary screening; the $\eta$ trajectories in Appendix~\ref{sec:ablation} are the completed 800-step training comparison. For the full experiments, we use a 400-sample validation set drawn from the mixed dataset.

% \subsection{Attention-loss numerical offset}
% \label{sec:attention_offset}

% The attention loss contains a numerical offset in $-\log(a_t+\epsilon_{\mathrm{attn}})$. We write it as $\epsilon_{\mathrm{attn}}$ here ($\varepsilon$ in the main text) to distinguish it from the GRPO surrogate clipping parameter. Table~\ref{tab:attention_offset} summarizes three development configurations evaluated with different offsets. Each row reports three evaluation replicates, with the four-benchmark average formed within each replicate.

% \input{tables/attention_offset}

% The selected setting, $\epsilon_{\mathrm{attn}}=0.10$, has the highest mean across the four benchmarks, exceeding the averages at 0.02 and 0.30 by 0.46 and 1.27 points, respectively. This comparison summarizes the numerical choices explored during development.

\subsection{Computational cost}
\label{sec:training_cost}

Table~\ref{tab:training_cost} compares MAPO and GRPO over the first 499 steps of training on the 29k-example dataset on 24 GPUs, using the same parallelism, a global batch of 192, and eight generations per prompt. The window contains 124 rollout-and-update steps and 375 update-only steps. MAPO takes 15.96 hours and GRPO 14.34 hours, an increase of 11.3\%, or approximately 1.6 hours.

\input{tables/training_cost}

The additional model-scale work comes from the frozen text-only forward pass and attention extraction. The reference pass is amortized across four optimizer steps, while full attention matrices are materialized only in the five target layers; the remaining layers retain flash attention. Relevance masks use detached token-wise operations, and POS tagging is performed on the CPU once per completed rollout before batch collation. The 30B consistency judge belongs to the shared training pipeline. These timings measure the combined training procedure; per-component wall time and peak-memory differences were not separately instrumented.

All auxiliary computations are training-only. At inference, MAPO uses the resulting policy without a reference model, POS tagger, or attention penalty. Generation latency can still depend on response length, which is separate from the training-time comparison above.

\FloatBarrier

\section{The ``bitter lesson'' of base policy selection: Why instruct models struggle with CoT calibration}
\label{sec:appendix_bitter_lesson}

Our main experiments demonstrate the efficacy of MAPO when applied to a base model inherently capable of extended reasoning (e.g., Qwen3-Omni-Thinking). However, applying the identical reinforcement learning formulations to an instruction-tuned model (Qwen3-Omni-Instruct) reveals a critical vulnerability: the underlying prior of the base policy severely bounds the potential for cross-modal CoT optimization. This section details the failure modes of standard GRPO and MAPO when forced to build reasoning trajectories from a model biased toward brevity.

\subsection{The illusion of reward and modality collapse}

\begin{figure}[ht]
    \centering
    \begin{subfigure}{0.48\textwidth}
        \centering
        \includegraphics[width=\linewidth]{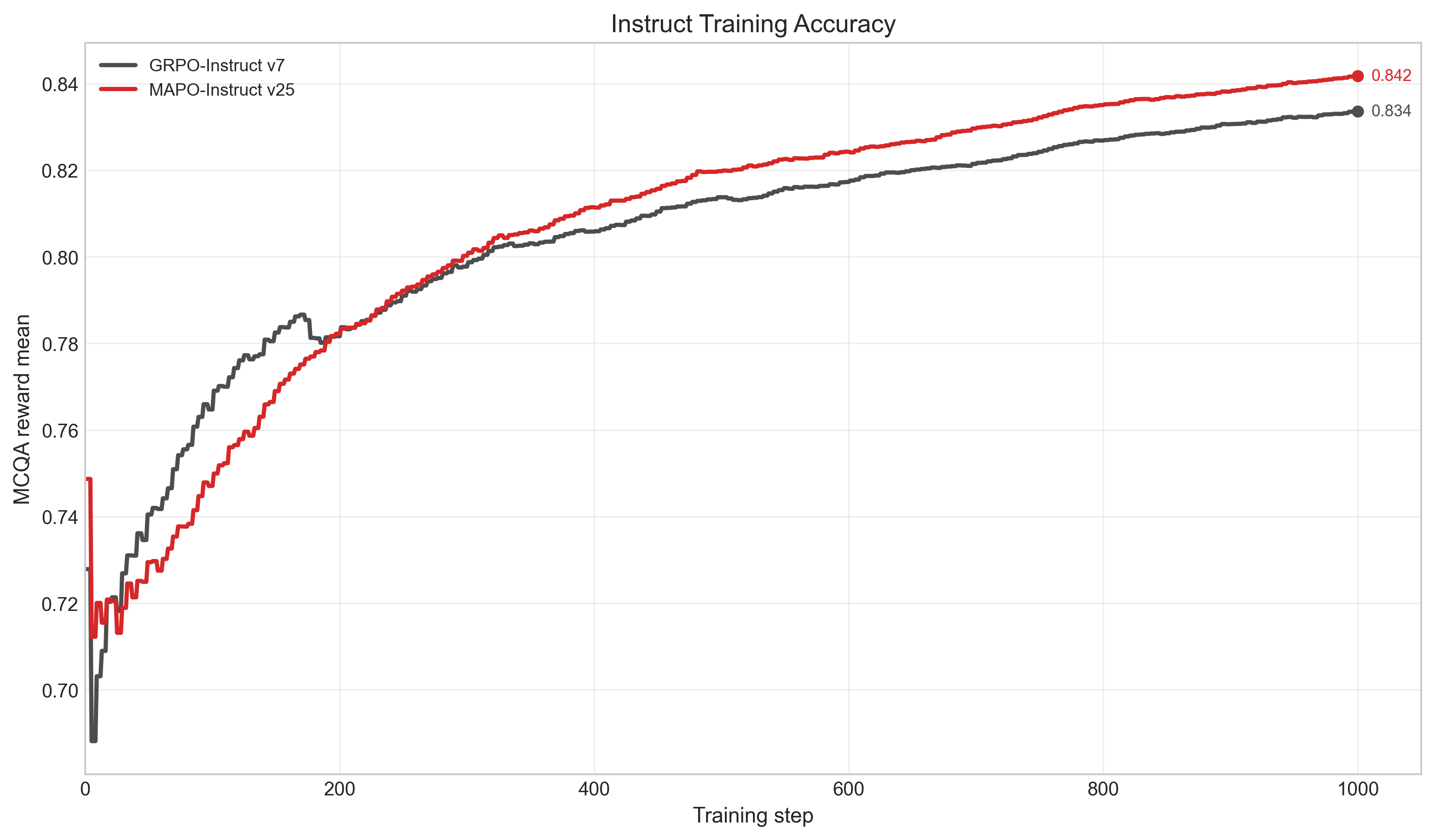}
        \caption{Instruct Training Accuracy}
        \label{fig:instruct_acc}
    \end{subfigure}
    \hfill
    \begin{subfigure}{0.48\textwidth}
        \centering
        \includegraphics[width=\linewidth]{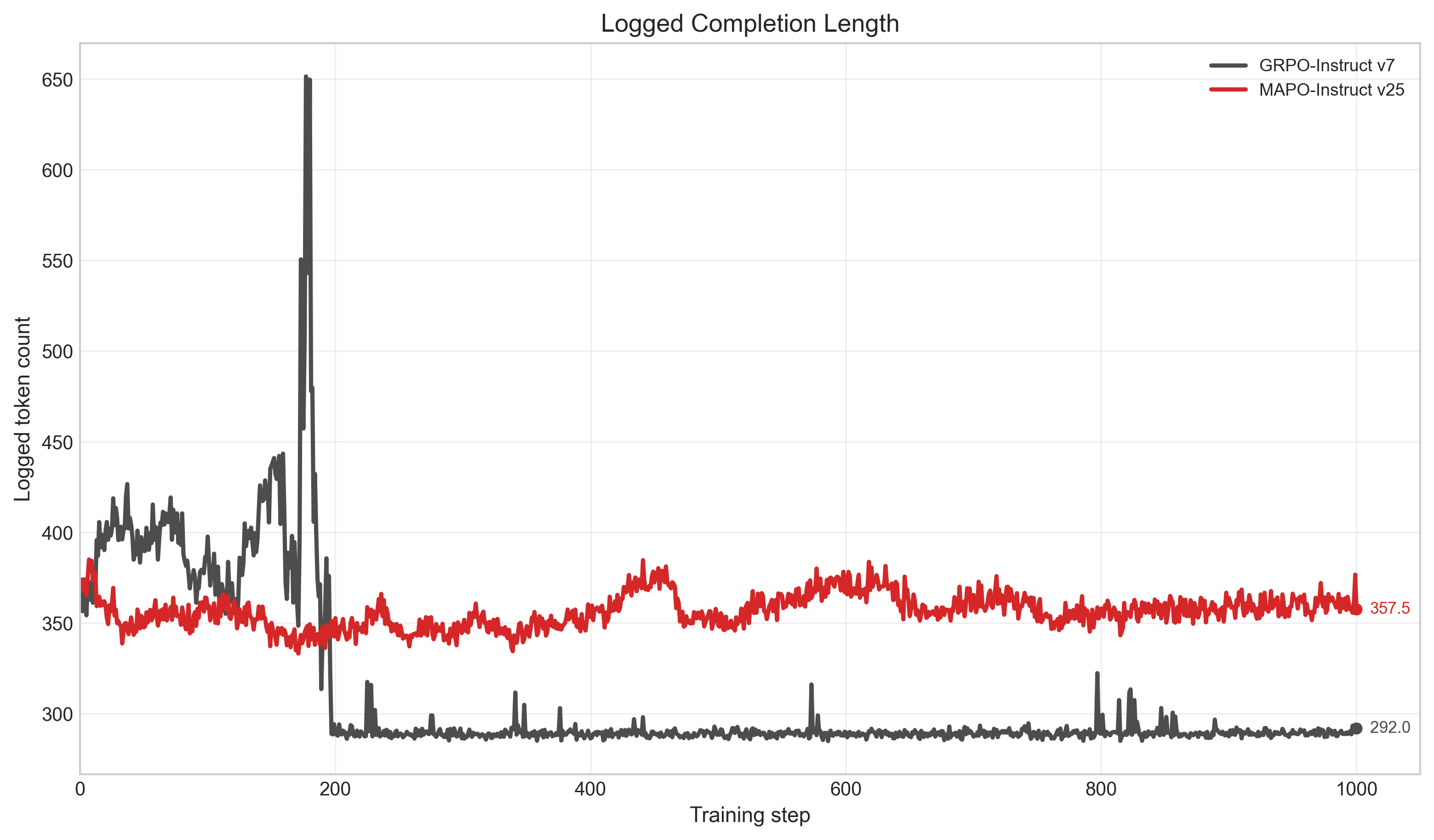}
        \caption{Logged Completion Length}
        \label{fig:instruct_len}
    \end{subfigure}
    \caption{Training dynamics of Qwen3-Omni-Instruct. (a) Both GRPO and MAPO show deceptively healthy task accuracy improvements. (b) The underlying completion length reveals a total collapse of the semantic reasoning trace for GRPO. While MAPO resists total collapse, it fails to meaningfully extend the reasoning horizon compared to models with an intrinsic CoT capability.}
    \label{fig:instruct_dynamics}
\end{figure}

Standard reinforcement learning metrics can be dangerously deceptive when applied to cross-modal reasoning tasks. As shown in Figure \ref{fig:instruct_dynamics}(a), both GRPO and MAPO achieve healthy, climbing MCQA reward curves when training the Instruct model, with GRPO reaching 0.8336 and MAPO reaching 0.8418. However, this high reward masks a profound epistemic collapse.

Because the reward functions primarily verify the exact answer and the external XML-like format (e.g., \texttt{<reasoning>...</reasoning>}), the optimizer rapidly discovers a short-length attractor. The Instruct model's natural policy already prefers brief, direct answers; the RL algorithm exploits this by shrinking the reasoning trace to minimize the risk of formatting errors and generation cost. Figure \ref{fig:instruct_dynamics}(b) illustrates this collapse: under GRPO, the mean reasoning text plunges from 62.6 words at step 0 to a mere 1.75 words by step 996. By the end of training, 100\% of rank-0 completions contain at most three words of reasoning, completely abandoning the task-relevant acoustic evidence. 

\subsection{Mechanistic failure of MAPO on Instruct models}

While MAPO successfully mitigates late-stage modality collapse in the Thinking model, it fails to produce robust improvements in the Instruct model. Despite maintaining a slightly longer reasoning average (53.7 words), MAPO's held-out accuracy on the Instruct model shows negligible gains over the baseline (e.g., MMAU test-mini: 76.5 $\rightarrow$ 75.9). 

This occurs because MAPO's modality-aware mechanisms fundamentally require meaningful response tokens---such as event descriptors, speech characteristics, or environmental nouns---to assign cross-modal credit and anchor attention. When the underlying policy produces generic, non-substantive text or directly hallucinates the answer, the cross-modal differential entropy signal vanishes. Consequently, the MAPO mask acts uniformly, the attention branch lacks valid targets, and the effective additive MAPO term shrinks to a negligible $\sim0.000040$, reducing the framework to standard, ungrounded GRPO dynamics.

\subsection{Qualitative trace degradation}

The qualitative progression of the model's outputs explicitly tracks this degradation. We present examples formatted to highlight the rapid deterioration of logical formatting, transitioning from initial grounded reasoning to externalized hacks, and finally to complete semantic collapse.

\begin{tcolorbox}[
    colback=blue!5!white,       % Light blueish background
    colframe=blue!50!gray,
    opacityback=1.0,            % Slight background transparency
    opacityframe=0.6,           % Very slight frame transparency
    title=Reasoning Consistency Evaluator Prompt,
    fonttitle=\bfseries,        % Bold title font
    boxrule=1pt,                 % Thinned the border slightly so it's less heavy
    arc=4pt,                     % Slightly rounder corners for a softer feel
    left=8pt, right=8pt, top=8pt, bottom=8pt % Added a tiny bit more breathing room
]
\texttt{<reasoning>The audio features distinct high-pitched crowing sounds, characteristic of a rooster, mixed with clucking and other poultry-like background noises... There are no sounds resembling a parrot, horse, or duck.</reasoning>}\\
\texttt{<answer>A rooster crows</answer>}
\end{tcolorbox}

As training progresses, the model discovers an alternative exploit: it strictly complies with the structural formatting reward while entirely abandoning the intended cognitive process. Rather than building a step-by-step deduction from the acoustic evidence, the model immediately commits to a predetermined answer and fills the required tags with dense, circular text to justify it. The following intermediate step illustrates this failure mode. The model generates a verbose, post-hoc rationalization that offers zero genuine chain-of-thought utility, serving strictly as a superficial placeholder to satisfy the \texttt{<reasoning>} block constraint before securing the accuracy reward:

\begin{tcolorbox}[
    colback=orange!5!white, colframe=orange!50!gray, opacityback=1.0,
    title=GRPO Intermediate Step: Formatting Hack and Externalized Reasoning,
    fonttitle=\bfseries, boxrule=1pt, arc=4pt, left=8pt, right=8pt, top=8pt, bottom=8pt
]
\texttt{<reasoning>chicken -> From all detected poultry species under multiple auditory clues, only chicken fits given distinct calls: ... chicken call pattern matching within time window confirms existence.</reasoning>}\\
\texttt{<answer>chicken</answer>}
\end{tcolorbox}

This formatting evasion eventually culminates in a pure answer-repetition loop that maximizes reward while minimizing reasoning overhead:

\begin{tcolorbox}[
    colback=red!5!white, colframe=red!50!gray, opacityback=1.0,
    title=GRPO Step 996: Complete Semantic Collapse,
    fonttitle=\bfseries, boxrule=1pt, arc=4pt, left=8pt, right=8pt, top=8pt, bottom=8pt
]
\texttt{<reasoning>sheep</reasoning>}\\
\texttt{<answer>sheep</answer>}
\end{tcolorbox}

Even when MAPO prevents this extreme length collapse, the lack of a calibrated reasoning prior leads the Instruct model to generate fluent but disconnected rationalizations. The model often recognizes acoustic elements (e.g., ``sounds of metal clanking'', ``child's voice'') but fails to logically and globally connect them, lazily drifting to incorrect priors. 

Ultimately, this demonstrates that MAPO can optimize and preserve existing reasoning structures, but it cannot organically create a complex chain-of-thought capability from an instruction-tuned model strictly optimized for brevity.

\section{Limitations}
\label{app:limitations}

MAPO changes how optimization pressure is allocated across tokens and how long the policy keeps drawing on the source audio; it does not change what the policy can perceive. Its gains are therefore bounded by the base model's acoustic representations and reasoning ability: directing attention toward a recording cannot supply a distinction the model has not learned, such as the sound of a rare instrument. Closing such gaps requires stronger multimodal pre-training rather than post-training alone.

MAPO also operates on an existing reasoning trace. The relevance mask reweights substantive tokens and the attention branch sustains their grounding, so the method applies to policies that already reason in long form. Appendix~\ref{sec:appendix_bitter_lesson} shows the consequence for a brevity-tuned Instruct policy: with few substantive tokens to reweight or anchor, the differential-entropy signal flattens and MAPO reduces to GRPO. Inducing such a trace in the first place is complementary work; MAPO is designed to preserve and sharpen one rather than to create it.

Finally, the grounding evidence has a defined scope. Audio attention mass is a proxy for source use, and the attention branch optimizes it directly. The diagnostics in Appendix~\ref{sec:audio_grounding} therefore test behavior rather than attention alone: they show that MAPO's answers depend on the supplied audio more than GRPO's, that a generated text prefix does not make the audio redundant, and that the regions MAPO attends to carry predictive evidence. What they do not establish is that each intermediate reasoning step is faithful to the audio; a direct test would require step-level interventions on the recording, which we leave to future work.

% \begin{ack}
% \end{ack}

%% file: tables/phase1.tex
% Generated by scripts/make_table_ablation.py; do not edit by hand.
\begin{table}[!htb]
\centering
\caption{Preliminary full-fine-tuning results after training on AVQA. Scores are reported point estimates; Avg. is the unweighted mean over the four benchmarks.}
\label{tab:phase1_results}
\setlength{\tabcolsep}{6pt}
\begin{tabular}{@{}p{0.27\linewidth}rrrrr@{}}
\toprule
\textbf{Method} & \textbf{MMAU} & \textbf{MMAR} & \textbf{MMSU} & \textbf{MMAU-Pro} & \textbf{Avg.} \\
\midrule
GRPO & 75.50 & 68.10 & 77.02 & 63.12 & 70.94 \\
MAPO & \textbf{77.10} & \textbf{69.20} & \textbf{77.74} & \textbf{63.48} & \textbf{71.88} \\
\bottomrule
\end{tabular}
\end{table}

%% file: tables/modality_main.tex
% Generated by scripts/make_table_main_results.py; do not edit by hand.
\begin{table}[!htb]
\centering
\caption{Overall accuracy on MMAU and MMAR, corresponding to Figure \ref{fig:modality_benchmarks}. MAPO scores are mean $\pm$ sample SD across three seeds; baseline scores are point estimates. Bold and underline denote the best and second-best open-weights scores. A dash indicates an unavailable result.}
\label{tab:modality_benchmarks}
\setlength{\tabcolsep}{6pt}
\begin{tabular}{@{}p{0.58\linewidth}rr@{}}
\toprule
\textbf{Model} & \textbf{MMAU-Mini} & \textbf{MMAR} \\
\midrule
Human & 82.23 & -- \\
\midrule
GPT-4o Audio & 62.50 & 63.50 \\
Gemini 2.5 Pro & 71.60 & -- \\
Nova 2 Omni~\citep{nova2omni} & 77.80 & -- \\
\midrule
Audio Flamingo 3 & 73.30 & -- \\
MiMo-Audio & 74.70 & -- \\
Qwen3-Omni-Thinking & 75.00 & 66.90 \\
Covo-Audio & 75.30 & -- \\
Qwen3-Omni-Instruct & 76.50 & 70.80 \\
Omni-R1~\citep{rouditchenko2025omnir1reallyneedaudio} & 77.00 & 63.40 \\
Step-Audio-2 & 77.58 & -- \\
Audio-Thinker & 77.70 & 65.30 \\
Step-Audio-R1~\citep{stepaudior1} & 77.70 & -- \\
Audio-DeepThinker~\citep{audiodeepthinker} & \textbf{78.50} & \textbf{74.00} \\
\midrule
MAPO & \underline{77.83} $\pm$ 0.25 & \underline{70.83} $\pm$ 0.21 \\
\bottomrule
\end{tabular}
\end{table}

%% file: tables/capability_main.tex
% Generated by scripts/make_table_main_results.py; do not edit by hand.
\begin{table}[!htb]
\centering
\caption{Overall accuracy on MMSU and MMAU-Pro, corresponding to Figure \ref{fig:capability_benchmarks}. MAPO scores are mean $\pm$ sample SD across three seeds; baseline scores are point estimates. Bold and underline denote the best and second-best open-weights scores. A dash indicates an unavailable result.}
\label{tab:capability_benchmarks}
\setlength{\tabcolsep}{6pt}
\begin{tabular}{@{}p{0.58\linewidth}rr@{}}
\toprule
\textbf{Model} & \textbf{MMSU} & \textbf{MMAU-Pro} \\
\midrule
Human & -- & 77.90 \\
\midrule
Gemini 1.5 Pro~\citep{geminiteam2024gemini15unlockingmultimodal} & 60.68 & -- \\
GPT-4o Audio~\citep{openai2024gpt4ocard} & 56.38 & 52.50 \\
Gemini 2.5 Flash & -- & 59.20 \\
\midrule
Qwen2.5-Omni-7B~\citep{xu2025qwen25omnitechnicalreport} & 60.57 & 52.20 \\
Covo-Audio & 66.64 & -- \\
Qwen3-Omni-Instruct & \underline{77.76} & 61.84 \\
Qwen3-Omni-Thinking & 76.30 & \underline{62.63} \\
Step-Audio-R1~\citep{stepaudior1} & 75.90 & -- \\
Audio-DeepThinker~\citep{audiodeepthinker} & 77.26 & -- \\
\midrule
MAPO & \textbf{79.35} $\pm$ 0.15 & \textbf{64.98} $\pm$ 0.28 \\
\bottomrule
\end{tabular}
\end{table}

%% file: tables/modality_details.tex
% Generated by scripts/make_table_main_results.py; approved MMAU/MMAR corrections applied.
\begin{table}[!htb]
\centering
\caption{Performance on modality-centric benchmarks (MMAU and MMAR), with Sound, Music, and Speech breakdowns. MMAR overall accuracy includes mixed-modality questions. Models are grouped by type (Proprietary, Open-Weights). MAPO breakdowns use a single evaluation; Table~\ref{tab:modality_benchmarks} reports the three-seed mean. Bold and underline denote the best and second-best open-weights scores.}
\label{tab:modality_details}
\setlength{\tabcolsep}{2pt} % Tightens horizontal spacing between columns
\begin{tabular}{@{}p{0.38\linewidth}|c c c c|c c c c@{}}
\toprule
\multirow{2}{*}{\textbf{Model}} & \multicolumn{4}{c|}{\textbf{MMAU-Mini}} & \multicolumn{4}{c}{\textbf{MMAR}} \\
\cmidrule(lr){2-5} \cmidrule(lr){6-9}
 & \textbf{Snd.} & \textbf{Mus.} & \textbf{Spch.} & \textbf{Avg.} & \textbf{Snd.} & \textbf{Mus.} & \textbf{Spch.} & \textbf{Overall} \\
\midrule
Human & 86.31 & 78.22 & 82.17 & 82.23 & - & - & - & - \\
\midrule
GPT-4o Audio & 64.56 & 56.29 & 66.67 & 62.50 & 53.94 & 50.97 & 70.41 & 63.50 \\
Gemini 2.5 Pro & 75.08 & 68.26 & 71.47 & 71.60 & - & - & - & - \\
Nova 2 Omni~\citep{nova2omni} & 81.08 & 70.36 & 81.98 & 77.80 & - & - & - & - \\
\midrule
Audio Flamingo 3 & 79.58 & 73.95 & 66.37 & 73.30 & - & - & - & - \\
MiMo-Audio & 81.68 & \underline{74.25} & 68.17 & 74.70 & - & - & - & - \\
Qwen3-Omni-Think & 77.48 & 70.96 & 76.58 & 75.00 & 62.42 & 49.51 & \textbf{78.57} & 66.90 \\
Covo-Audio & 78.68 & \textbf{76.05} & 71.17 & 75.30 & - & - & - & - \\
Qwen3-Omni-Inst & 77.78 & 73.05 & \underline{78.68} & 76.50 & 64.85 & \textbf{56.80} & 77.21 & \underline{70.80} \\
Omni-R1~\citep{rouditchenko2025omnir1reallyneedaudio} & 81.70 & 73.40 & 76.00 & 77.00 & \underline{67.30} & 51.50 & 64.30 & 63.40 \\
Step-Audio-2 & \textbf{84.04} & 73.56 & 75.15 & 77.58 & - & - & - & - \\
Audio-Thinker & \underline{81.98} & \underline{74.25} & 76.88 & \underline{77.70} & \textbf{68.48} & 53.88 & 64.29 & 65.30 \\
\midrule
MAPO & 79.88 & 71.86 & \textbf{81.68} & \textbf{77.80} & 66.06 & \underline{55.83} & \underline{77.89} & \textbf{70.90} \\
\bottomrule
\end{tabular}
\end{table}

%% file: tables/capability_details.tex
% Generated by scripts/make_table_main_results.py; all numerical breakdowns retained.
\begin{table}[!htb]
\centering
\caption{Performance comparison on capability and task-centric benchmarks (MMSU and MMAU-Pro). MMSU evaluates core Perception and Reasoning, while MMAU-Pro evaluates Instruction Following (AIF) and Open-ended question answering. Entries are point estimates. MAPO breakdowns use a single evaluation; Table~\ref{tab:capability_benchmarks} reports the three-seed mean. Bold and underline denote the best and second-best open-weights scores.}
\label{tab:capability_details}
\setlength{\tabcolsep}{1.5pt} % Adjusts column spacing to fit width without altering font size
\begin{tabular}{@{}p{0.37\linewidth}|ccc|cccc@{}}
\toprule
\multirow{2}{*}{\textbf{Model}} & \multicolumn{3}{c|}{\textbf{MMSU}} & \multicolumn{4}{c}{\textbf{MMAU-Pro}} \\
\cmidrule(lr){2-4} \cmidrule(lr){5-8}
 & \textbf{Percept.} & \textbf{Reason.} & \textbf{Overall} & \textbf{AIF} & \textbf{Closed} & \textbf{Open} & \textbf{Overall} \\
\midrule
Human & - & - & - & 100.00 & 77.56 & 77.30 & 77.90 \\
\midrule
Gemini 1.5 Pro~\citep{geminiteam2024gemini15unlockingmultimodal} & 46.10 & 76.16 & 60.68 & - & - & - & - \\
GPT-4o Audio~\citep{openai2024gpt4ocard} & 39.67 & 71.96 & 56.38 & 82.50 & 53.20 & 43.20 & 52.50 \\
% Gemini 2.0 Flash & - & - & - & 94.20 & 53.46 & 66.80 & 55.70 \\
Gemini 2.5 Flash & - & - & - & 95.10 & 57.39 & 67.50 & 59.20 \\
\midrule
Qwen2.5-Omni-7B~\citep{xu2025qwen25omnitechnicalreport} & 42.50 & 79.83 & 60.57 & 61.30 & 52.01 & 52.30 & 52.20 \\
Covo-Audio & 58.95 & 74.83 & 66.64 & - & - & - & - \\
Qwen3-Omni-Instruct & 73.95 & \underline{81.82} & \underline{77.76} & \underline{89.66} & 57.84 & \textbf{87.38} & 61.84 \\
Qwen3-Omni-Thinking & \underline{74.19} & 78.55 & 76.30 & 62.07 & \underline{59.63} & 84.77 & \underline{62.63} \\
\midrule
MAPO & \textbf{76.36} & \textbf{82.56} & \textbf{79.36} & \textbf{95.40} & \textbf{62.00} & \underline{85.32} & \textbf{65.29} \\
\bottomrule
\end{tabular}
\end{table}

%% file: sections/audio_grounding.tex
\FloatBarrier
\section{Analysis of audio grounding in the reasoning trajectory}
\label{sec:audio_grounding}

The benchmark results show how MAPO affects answer accuracy, while the analyses below examine how the models use the accompanying audio during reasoning. We first study attention around the transition from an initial caption stage to later reasoning. We then compare answers under paired changes to the audio input and test whether access to the source remains useful after part of the reasoning has already been generated.

\subsection{Attention across reasoning stages}
\label{sec:stage_attention}

As a response grows, the audio keys compete with an increasing number of text keys. A decrease in their total attention mass can therefore reflect a change in the relative number of keys as well as a change in the model's preference for audio. To separate these effects, let $m_t$ denote audio attention mass and let $K_a$ and $K_{n,t}$ denote the numbers of available audio and non-audio keys. We measure the per-key log preference
\begin{equation}
    q_t = \log\frac{m_t/K_a}{(1-m_t)/K_{n,t}}.
    \label{eq:normalized_audio_preference}
\end{equation}
A value of zero assigns equal attention per key to the two groups. We compute mean-head and max-head reductions, averaged over decoder layers, and compare the change across the caption-to-reasoning transition.

The probe uses 200 paired MMAU prompts, teacher-forcing each model's own response. We identify the transition from the initial \emph{caption stage} to reasoning through linguistic cues that mark a shift from recounting the audio to evaluating the question or candidate answers. For example, one Base Thinking response first notes ``a very high-pitched, strained vocal sound'' before turning to ``Next, evaluating the choices.'' We take the latter phrase as the boundary and compare attention in short windows on either side, aligning each response independently.

Figure~\ref{fig:stage_attention} summarizes these changes for the 166 Base Thinking and 157 MAPO responses with identifiable transitions. Paired comparisons use the 143 prompts aligned for both models, with 95\% confidence intervals obtained by paired bootstrap resampling of prompts.

\input{figures/stage_attention}

Raw mean-head mass decreases for both Base Thinking and MAPO, by 0.0194 and 0.0159, respectively. After key-count normalization, Base still shows a decline of 0.313, whereas MAPO's mean change is positive at 0.033. The paired MAPO-minus-Base contrast is $+0.338$ $[0.260, 0.420]$. The max-head reduction gives the same ordering, with a paired contrast of $+0.314$ $[0.154, 0.470]$. These paired estimates need not equal the difference between the separately aggregated model means. Together, the results indicate that the transition-localized change extends beyond the growing text-key denominator: MAPO retains more per-key audio preference as the response moves into reasoning.

\subsection{Sensitivity to the audio input}
\label{sec:audio_perturbations}

These measurements describe internal model behavior. To examine how the input affects the final answer, we next evaluate each question with its original audio and with three altered inputs: complete silence, a duration-matched recording from another example, and text only. Silence replaces the waveform with zeros, while text-only input removes the audio input and its placeholder. MAPO and the full-fine-tuning GRPO baseline receive the same questions and perturbations.

We construct random and audio-essential cohorts for MMAU and MMSU, each containing 200 examples. Random cohorts are selected by a deterministic hash ordering independently of model outcomes. Audio-essential cohorts contain questions that the pretrained Qwen3-Omni base answers incorrectly without audio. For a perturbation $c$, define the paired difference in accuracy drops as
\begin{equation}
 D_c = (A^{\mathrm{MAPO}}_{\mathrm{clean}}-A^{\mathrm{MAPO}}_c)
       -(A^{\mathrm{GRPO}}_{\mathrm{clean}}-A^{\mathrm{GRPO}}_c),
 \label{eq:paired_audio_drop}
\end{equation}
where $A$ is accuracy in percent. Positive $D_c$ means that removing or replacing the audio reduces MAPO's accuracy more than GRPO's on that cohort.

\input{tables/audio_perturbations}

The differences in Table~\ref{tab:audio_perturbations} are positive across all four cohorts and all three perturbations, ranging from 2.0 to 15.0 percentage points. MAPO also has higher clean accuracy in every cohort. The largest contrasts occur on audio-essential MMAU, where its clean advantage of 7.0 points accompanies differences in drops of 11.0 under silence, 10.0 under mismatch, and 15.0 under text-only input. These comparisons connect the models' use of audio to their answers, complementing the attention measurements above.

% \input{tables/grounding_uncertainty}

% For the random MMSU cohort, Table~\ref{tab:grounding_uncertainty} adds paired bootstrap intervals and a completion-sensitive analysis. Greedy decoding uses a 2,048-token limit; across the four input conditions, MAPO completes its reasoning block in 661 of 800 responses and GRPO in 717 of 800. Restricting each comparison to questions where both models complete under both inputs yields smaller cohorts. The silence contrast remains positive with an interval excluding zero, while the text-only and mismatch intervals include zero. Clean accuracy, perturbation sensitivity, and completion therefore provide complementary information about the observed differences. In particular, a larger drop measures dependence on the supplied audio, rather than the correctness of every intermediate reasoning step.

\subsection{Audio dependence with a shared reasoning prefix}
\label{sec:prefix_audio}

An initial caption can carry acoustic information into the generated text. To examine whether this text is sufficient for the remaining reasoning, we use the pretrained Base Thinking model on 200 audio-essential MMAR examples. We first generate a response with full audio. We then construct two continuation inputs, one with the source audio and one with text only, and supply both with exactly the same first quarter, third, or half of the original response tokens. The prefix representations are recomputed under each input condition before generation continues.

\input{figures/prefix_audio}
\input{tables/prefix_agreement}

Figure~\ref{fig:prefix_audio} shows that the retained-minus-removed accuracy gap narrows as more of the prefix is supplied: 17.0, 14.0, and 6.0 points. All three paired 95\% intervals remain above zero. The longer prefix thus supplies useful information, but even half of the original response does not make the source audio redundant. Answer agreement follows the same trend (Table~\ref{tab:prefix_agreement}); this measures whether the continuation reproduces the original answer, independently of whether that answer is correct. Thus, the shared text prefix does not fully substitute for audio conditioning in the Base Thinking model. The MAPO--GRPO perturbation comparison above examines how dependence on the supplied audio differs after training.

\input{sections/local_evidence_masking}

\FloatBarrier

%% file: figures/stage_attention.tex
% Generated by scripts/make_figure_appendix.py; do not edit by hand.
\begin{figure}[!htb]
\centering
\includegraphics[width=\linewidth]{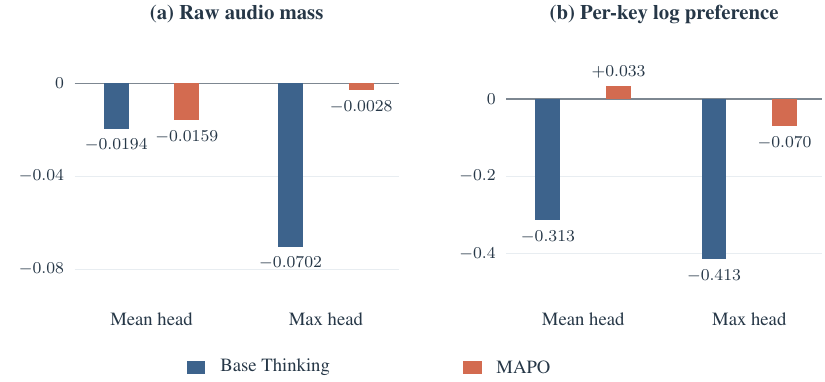}
\caption{Change in audio attention across the caption-to-reasoning transition. Bars show model-level means of the post-transition minus pre-transition statistic. Both raw mean-head masses decline, while key-count normalization separates the Base and MAPO trends. Paired contrasts on the shared subset and their uncertainty are reported in the text.}
\label{fig:stage_attention}
\end{figure}

%% file: tables/audio_perturbations.tex
% Generated by scripts/make_table_appendix.py; do not edit by hand.
\begin{table}[!htb]
\centering
\caption{Accuracy (\%) under paired audio perturbations. Each cohort contains 200 questions evaluated by both models. $D_c$ is the MAPO-minus-GRPO difference in clean-to-perturbed accuracy drops (Equation~\ref{eq:paired_audio_drop}); positive values indicate a larger drop for MAPO. These cohort estimates are separate from the full-benchmark results.}
\label{tab:audio_perturbations}
\setlength{\tabcolsep}{8pt}
\begin{tabular}{@{}lrrrr@{}}
\toprule
Model & Clean & Silence & Mismatch & Text only \\
\midrule
\multicolumn{5}{l}{\textit{MMAU random ($N=200$)}} \\
MAPO & $71.5$ & $45.5$ & $41.0$ & $47.0$ \\
GRPO & $70.0$ & $46.0$ & $43.5$ & $52.0$ \\
$D_c$ (points) & -- & $+2.0$ & $+4.0$ & $+6.5$ \\
\midrule
\multicolumn{5}{l}{\textit{MMAU audio-essential ($N=200$)}} \\
MAPO & $67.0$ & $17.0$ & $27.0$ & $12.0$ \\
GRPO & $60.0$ & $21.0$ & $30.0$ & $20.0$ \\
$D_c$ (points) & -- & $+11.0$ & $+10.0$ & $+15.0$ \\
\midrule
\multicolumn{5}{l}{\textit{MMSU random ($N=200$)}} \\
MAPO & $68.0$ & $36.0$ & $35.5$ & $35.5$ \\
GRPO & $60.0$ & $37.5$ & $34.5$ & $37.0$ \\
$D_c$ (points) & -- & $+9.5$ & $+7.0$ & $+9.5$ \\
\midrule
\multicolumn{5}{l}{\textit{MMSU audio-essential ($N=200$)}} \\
MAPO & $47.0$ & $25.0$ & $23.0$ & $5.0$ \\
GRPO & $44.0$ & $26.0$ & $23.0$ & $9.0$ \\
$D_c$ (points) & -- & $+4.0$ & $+3.0$ & $+7.0$ \\
\bottomrule
\end{tabular}
\end{table}

%% file: figures/prefix_audio.tex
% Generated by scripts/make_figure_appendix.py; do not edit by hand.
\begin{figure}[!htb]
\centering
\includegraphics[width=\linewidth]{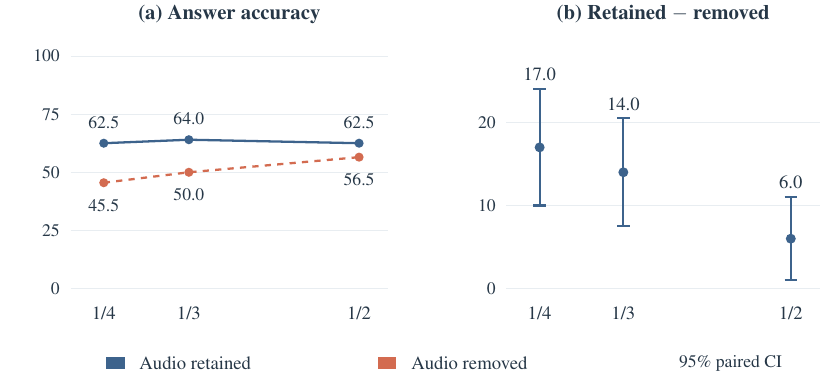}
\caption{Audio dependence with a shared reasoning prefix on 200 audio-essential MMAR examples. (a) Accuracy (\%) with source audio retained in or removed from the continuation input, supplying identical prefix tokens from a Base Thinking response in both conditions. Prefix representations are recomputed separately under each input condition. (b) Retained-minus-removed accuracy in percentage points, with paired 95\% bootstrap intervals. The x-axis gives the fraction of the original response supplied as a prefix.}
\label{fig:prefix_audio}
\end{figure}

%% file: tables/prefix_agreement.tex
% Generated by scripts/make_table_appendix.py; do not edit by hand.
\begin{table}[!htb]
\centering
\caption{Agreement with the original full-audio answer in the 200-example MMAR prefix experiment. Agreement is in percent; gaps are percentage points with paired 95\% bootstrap intervals. McNemar tests refer to agreement, not answer accuracy.}
\label{tab:prefix_agreement}
\setlength{\tabcolsep}{5pt}
\begin{tabular}{@{}lrrrr@{}}
\toprule
Prefix & Retained & Removed & Gap [95\% CI] & McNemar $p$ \\
\midrule
1/4 & 92.0 & 62.5 & $29.5$ $[22.5, 36.5]$ & $4.12\times10^{-14}$ \\
1/3 & 93.0 & 73.5 & $19.5$ $[13.0, 26.0]$ & $1.83\times10^{-8}$ \\
1/2 & 95.5 & 86.0 & $9.5$ $[4.5, 14.5]$ & $0.00031$ \\
\bottomrule
\end{tabular}
\end{table}

%% file: sections/local_evidence_masking.tex
\subsection{Local evidence masking}
\label{app:local_masking}

The whole-input perturbations above test whether answers depend on the supplied audio. Local masking asks which parts of the recording contribute to the prediction. For each example, we aggregate token-to-audio attention during generation and remove a fraction $r$ of the recording using three rules: the highest-attended regions, the lowest-attended regions, or duration-matched random regions. The Attention Causal Selectivity (ACS) score compares random and top-attention masking:
\begin{equation}
    \text{ACS}(r) = \text{Acc}_{\text{rand}}(r) - \text{Acc}_{\text{top}}(r). \label{eq:acs}
\end{equation}
Positive ACS indicates that removing highly attended regions is more damaging than removing an equal duration chosen at random.

\begin{table}[ht]
\centering
\caption{Inference-time local evidence masking. ACS is defined in Equation~\ref{eq:acs}, where positive values indicate that masking highly attended audio regions is more damaging than duration-matched random masking. Low-mask accuracy serves as an additional control for removing regions assigned low attention by the model.}
\label{tab:local_masking}
\setlength{\tabcolsep}{2pt} % Adjusted column spacing to fit width without reducing font size
\begin{tabular}{lccccc}
\toprule
\textbf{Model} & \textbf{Mask Ratio} & \textbf{Top-Mask Acc.} & \textbf{Random-Mask Acc.} & \textbf{Low-Mask Acc.} & \textbf{ACS} \\
\midrule
MAPO w/o $\mathcal{L}_{\text{attn}}$ & 5\%  & 0.822 & 0.825 & 0.797 & 0.003 \\
MAPO w/o $\mathcal{L}_{\text{attn}}$ & 10\% & 0.805 & 0.847 & 0.822 & 0.042 \\
MAPO w/o $\mathcal{L}_{\text{attn}}$ & 30\% & 0.771 & 0.817 & 0.873 & 0.046 \\
MAPO w/o $\mathcal{L}_{\text{attn}}$ & 50\% & 0.712 & 0.758 & 0.788 & 0.046 \\
\midrule
MAPO w/ $\mathcal{L}_{\text{attn}}$ & 5\%  & 0.868 & 0.886 & 0.878 & 0.018 \\
MAPO w/ $\mathcal{L}_{\text{attn}}$ & 10\% & 0.846 & 0.891 & 0.862 & 0.046 \\
MAPO w/ $\mathcal{L}_{\text{attn}}$ & 30\% & 0.789 & 0.831 & 0.837 & 0.042 \\
MAPO w/ $\mathcal{L}_{\text{attn}}$ & 50\% & 0.707 & 0.776 & 0.846 & 0.068 \\
\bottomrule
\end{tabular}
\end{table}

Table~\ref{tab:local_masking} shows positive ACS at every masking ratio for both MAPO variants. At 50\% masking, full MAPO has an ACS of 0.068 and retains accuracy of 0.846 when low-attention regions are removed, compared with 0.707 when top-attention regions are removed. The contrast suggests that the attended regions contain more prediction-relevant evidence. The attention branch also generally preserves higher accuracy under random and low-attention masking, so the increased selectivity coexists with robustness to removing other regions. Because acoustic evidence can be distributed over time, these measurements describe the effect of the specified occlusions, including any distribution shift they introduce.

%% file: appendix/exp_1.tex
% Appendix-ready token heatmap for MAPO qualitative analysis.
% Source: /home/hltcoe/cxiao/mapo/exp/analysis_error/mmau/v6-20260416-033441/stage5/colored_text/0014_29b7c031-e275-4084-8edc-0b1cc177bad8.tex
% Required in the main paper preamble:
%   \usepackage{xcolor}
% Recommended for nicer <think> tag glyphs:
%   \usepackage[T1]{fontenc}

\begingroup
\definecolor{MapoInk}{RGB}{28,36,48}
\definecolor{MapoMuted}{RGB}{92,102,116}
\definecolor{MapoFrame}{RGB}{202,211,222}
\definecolor{MapoSoft}{RGB}{247,249,252}
\definecolor{MapoMetricTitle}{RGB}{239,243,248}
\definecolor{MapoCorrect}{RGB}{34,119,80}
\definecolor{MapoCorrectBg}{RGB}{238,248,243}
\definecolor{MapoIncorrect}{RGB}{174,60,60}
\definecolor{MapoIncorrectBg}{RGB}{252,241,241}

\newcommand{\MapoToken}[2]{%
  \begingroup\colorbox[rgb]{#1}{\strut #2}\endgroup%
  \hspace{0.05em}\allowbreak%
}

\newcommand{\MapoBox}[2]{%
  \par\smallskip\noindent%
  {\setlength{\fboxsep}{4pt}\setlength{\fboxrule}{0.35pt}%
   \fcolorbox{MapoFrame}{#1}{%
     \parbox{\dimexpr\linewidth-2\fboxsep-2\fboxrule\relax}{#2}%
   }}%
  \par\smallskip%
}

\newcommand{\MapoModelHeader}[5]{%
  \MapoBox{#4}{%
    {\bfseries #1}\hfill{\bfseries\textcolor{#3}{#2}}\par
    {\normalsize\textcolor{MapoMuted}{Prediction:} #5}%
  }%
}

\newcommand{\MapoMetricHeader}[2]{%
  \MapoBox{MapoMetricTitle}{%
    {\normalsize\bfseries #1}\par
    {\normalsize\itshape #2}%
  }%
}

\newenvironment{MapoTokenStream}{%
  \par\noindent\begingroup%
  \normalsize\ttfamily\raggedright\sloppy%
  \setlength{\fboxsep}{0.75pt}%
  \setlength{\parindent}{0pt}%
  \setlength{\emergencystretch}{3em}%
  \hyphenpenalty=10000\exhyphenpenalty=10000%
}{\par\endgroup\smallskip}

\newcommand{\MapoMetricRule}{%
  \par\noindent{\color{MapoFrame}\rule{\linewidth}{0.25pt}}\par\smallskip%
}

\MapoBox{MapoSoft}{%
\textbf{Question:} Based on the given audio, identify the source of mechanical sounds.\\
\textbf{Ground truth:} Factory machinery%
}

\MapoModelHeader{Baseline}{Incorrect}{MapoIncorrect}{MapoIncorrectBg}{Wind turbine}

\begin{MapoTokenStream}
\MapoToken{0.000,0.267,0.106}{\textless{}think\textgreater{}}%
\MapoToken{0.000,0.267,0.106}{\phantom{x}}%
\MapoToken{0.004,0.431,0.176}{Got}%
\MapoToken{0.239,0.651,0.353}{ it}%
\MapoToken{0.165,0.573,0.290}{,}%
\MapoToken{0.000,0.267,0.106}{ let}%
\MapoToken{0.129,0.537,0.267}{'s}%
\MapoToken{0.365,0.725,0.420}{ tackle}%
\MapoToken{0.000,0.267,0.106}{ the}%
\MapoToken{0.416,0.749,0.443}{ problem}%
\MapoToken{0.322,0.702,0.396}{.}%
\MapoToken{0.592,0.831,0.573}{ First}%
\MapoToken{0.000,0.267,0.106}{,}%
\MapoToken{0.000,0.286,0.114}{ the}%
\MapoToken{0.306,0.698,0.392}{ user}%
\MapoToken{0.592,0.831,0.573}{ wants}%
\MapoToken{0.620,0.843,0.596}{ me}%
\MapoToken{0.541,0.808,0.533}{ to}%
\MapoToken{0.478,0.780,0.482}{ identify}%
\MapoToken{0.000,0.314,0.125}{ the}%
\MapoToken{0.239,0.651,0.353}{ source}%
\MapoToken{0.169,0.576,0.294}{ of}%
\MapoToken{0.000,0.267,0.106}{ the}%
\MapoToken{0.000,0.267,0.106}{ mechanical}%
\MapoToken{0.000,0.267,0.106}{ sound}%
\MapoToken{0.094,0.510,0.239}{ in}%
\MapoToken{0.000,0.267,0.106}{ the}%
\MapoToken{0.000,0.267,0.106}{ provided}%
\MapoToken{0.000,0.267,0.106}{ audio}%
\MapoToken{0.055,0.475,0.212}{ clip}%
\MapoToken{0.765,0.906,0.737}{ and}%
\MapoToken{0.753,0.902,0.725}{ choose}%
\MapoToken{0.710,0.882,0.682}{ from}%
\MapoToken{0.000,0.267,0.106}{ the}%
\MapoToken{0.710,0.882,0.682}{ given}%
\MapoToken{0.620,0.843,0.596}{ options}%
\MapoToken{0.569,0.824,0.557}{:}%
\MapoToken{0.212,0.624,0.329}{ Factory}%
\MapoToken{0.239,0.651,0.353}{ machinery}%
\MapoToken{0.584,0.827,0.569}{,}%
\MapoToken{0.192,0.604,0.314}{ Wind}%
\MapoToken{0.224,0.635,0.341}{ turbine}%
\MapoToken{0.357,0.722,0.416}{,}%
\MapoToken{0.208,0.620,0.325}{ Car}%
\MapoToken{0.239,0.651,0.353}{ engine}%
\MapoToken{0.675,0.871,0.651}{,}%
\MapoToken{0.000,0.267,0.106}{ Air}%
\MapoToken{0.282,0.682,0.376}{plane}%
\MapoToken{0.557,0.816,0.545}{.}%
\MapoToken{0.620,0.843,0.596}{ Then}%
\MapoToken{0.302,0.694,0.388}{,}%
\MapoToken{0.000,0.267,0.106}{ the}%
\MapoToken{0.000,0.267,0.106}{ audio}%
\MapoToken{0.000,0.267,0.106}{ features}%
\MapoToken{0.000,0.267,0.106}{ a}%
\MapoToken{0.000,0.267,0.106}{ continuous}%
\MapoToken{0.000,0.267,0.106}{,}%
\MapoToken{0.000,0.267,0.106}{ powerful}%
\MapoToken{0.000,0.267,0.106}{,}%
\MapoToken{0.000,0.384,0.153}{ low}%
\MapoToken{0.000,0.267,0.106}{-frequency}%
\MapoToken{0.000,0.337,0.133}{ hum}%
\MapoToken{0.027,0.451,0.192}{ or}%
\MapoToken{0.114,0.525,0.251}{ roar}%
\MapoToken{0.090,0.506,0.239}{ with}%
\MapoToken{0.188,0.600,0.314}{ a}%
\MapoToken{0.125,0.533,0.263}{ distinct}%
\MapoToken{0.000,0.392,0.157}{ '}%
\MapoToken{0.110,0.522,0.251}{who}%
\MapoToken{0.000,0.424,0.173}{osh}%
\MapoToken{0.145,0.553,0.278}{ing}%
\MapoToken{0.239,0.651,0.353}{'}%
\MapoToken{0.247,0.659,0.357}{ or}%
\MapoToken{0.129,0.537,0.267}{ rushing}%
\MapoToken{0.396,0.741,0.435}{ quality}%
\MapoToken{0.498,0.788,0.498}{ suggesting}%
\MapoToken{0.533,0.808,0.529}{ a}%
\MapoToken{0.580,0.827,0.565}{ large}%
\MapoToken{0.439,0.761,0.455}{ volume}%
\MapoToken{0.447,0.765,0.459}{ of}%
\MapoToken{0.369,0.725,0.420}{ air}%
\MapoToken{0.584,0.827,0.569}{ moving}%
\MapoToken{0.557,0.816,0.545}{ at}%
\MapoToken{0.565,0.820,0.553}{ high}%
\MapoToken{0.533,0.808,0.529}{ speed}%
\MapoToken{0.475,0.776,0.478}{,}%
\MapoToken{0.271,0.678,0.373}{ and}%
\MapoToken{0.251,0.667,0.365}{ it}%
\MapoToken{0.439,0.761,0.455}{ has}%
\MapoToken{0.451,0.769,0.463}{ a}%
\MapoToken{0.259,0.671,0.365}{ cyc}%
\MapoToken{0.294,0.690,0.384}{lical}%
\MapoToken{0.557,0.816,0.545}{ or}%
\MapoToken{0.353,0.718,0.412}{ rhyth}%
\MapoToken{0.463,0.773,0.471}{mic}%
\MapoToken{0.490,0.784,0.494}{ component}%
\MapoToken{0.584,0.827,0.569}{ resembling}%
\MapoToken{0.584,0.827,0.569}{ blades}%
\MapoToken{0.706,0.882,0.678}{ rotating}%
\MapoToken{0.659,0.863,0.635}{ through}%
\MapoToken{0.647,0.859,0.624}{ air}%
\MapoToken{0.608,0.839,0.588}{ with}%
\MapoToken{0.416,0.749,0.443}{ a}%
\MapoToken{0.353,0.718,0.412}{ '}%
\MapoToken{0.416,0.749,0.443}{th}%
\MapoToken{0.502,0.792,0.502}{w}%
\MapoToken{0.533,0.808,0.529}{op}%
\MapoToken{0.396,0.741,0.435}{-th}%
\MapoToken{0.541,0.808,0.533}{w}%
\MapoToken{0.525,0.800,0.522}{op}%
\MapoToken{0.275,0.682,0.376}{-th}%
\MapoToken{0.514,0.796,0.510}{w}%
\MapoToken{0.529,0.804,0.525}{op}%
\MapoToken{0.545,0.812,0.537}{'}%
\MapoToken{0.643,0.855,0.620}{ sound}%
\MapoToken{0.698,0.878,0.675}{ mixed}%
\MapoToken{0.624,0.847,0.600}{ into}%
\MapoToken{0.525,0.800,0.522}{ the}%
\MapoToken{0.557,0.816,0.545}{ roar}%
\MapoToken{0.639,0.855,0.616}{;}%
\MapoToken{0.706,0.882,0.678}{ additionally}%
\MapoToken{0.322,0.702,0.396}{,}%
\MapoToken{0.498,0.788,0.498}{ it}%
\MapoToken{0.533,0.808,0.529}{ is}%
\MapoToken{0.314,0.698,0.392}{ very}%
\MapoToken{0.322,0.702,0.396}{ loud}%
\MapoToken{0.596,0.835,0.580}{ and}%
\MapoToken{0.400,0.741,0.435}{ overwhelming}%
\MapoToken{0.514,0.796,0.510}{ with}%
\MapoToken{0.596,0.835,0.580}{ a}%
\MapoToken{0.733,0.894,0.706}{ sense}%
\MapoToken{0.624,0.847,0.600}{ of}%
\MapoToken{0.667,0.867,0.643}{ immense}%
\MapoToken{0.698,0.878,0.675}{ power}%
\MapoToken{0.769,0.910,0.741}{ and}%
\MapoToken{0.737,0.894,0.710}{ scale}%
\MapoToken{0.718,0.886,0.694}{.}%
\MapoToken{0.804,0.925,0.780}{ Next}%
\MapoToken{0.722,0.890,0.698}{,}%
\MapoToken{0.796,0.922,0.773}{ evaluating}%
\MapoToken{0.765,0.906,0.737}{ the}%
\MapoToken{0.780,0.914,0.753}{ options}%
\MapoToken{0.710,0.882,0.682}{:}%
\MapoToken{0.612,0.843,0.592}{ Factory}%
\MapoToken{0.663,0.863,0.639}{ machinery}%
\MapoToken{0.718,0.886,0.694}{ typically}%
\MapoToken{0.698,0.878,0.675}{ produces}%
\MapoToken{0.643,0.855,0.620}{ cl}%
\MapoToken{0.690,0.875,0.667}{anking}%
\MapoToken{0.698,0.878,0.675}{,}%
\MapoToken{0.710,0.882,0.682}{ grinding}%
\MapoToken{0.686,0.875,0.659}{,}%
\MapoToken{0.737,0.894,0.710}{ rhyth}%
\MapoToken{0.745,0.898,0.722}{mic}%
\MapoToken{0.722,0.890,0.698}{ pounding}%
\MapoToken{0.792,0.918,0.765}{,}%
\MapoToken{0.769,0.910,0.741}{ or}%
\MapoToken{0.706,0.882,0.678}{ high}%
\MapoToken{0.682,0.871,0.655}{-p}%
\MapoToken{0.694,0.878,0.671}{itched}%
\MapoToken{0.729,0.890,0.702}{ wh}%
\MapoToken{0.757,0.902,0.729}{ir}%
\MapoToken{0.788,0.918,0.761}{ring}%
\MapoToken{0.804,0.925,0.780}{ sounds}%
\MapoToken{0.808,0.925,0.784}{,}%
\MapoToken{0.729,0.890,0.702}{ but}%
\MapoToken{0.659,0.863,0.635}{ the}%
\MapoToken{0.706,0.882,0.678}{ audio}%
\MapoToken{0.792,0.918,0.765}{'s}%
\MapoToken{0.757,0.902,0.729}{ dominant}%
\MapoToken{0.831,0.933,0.808}{ feature}%
\MapoToken{0.792,0.918,0.765}{ is}%
\MapoToken{0.761,0.906,0.733}{ the}%
\MapoToken{0.682,0.871,0.655}{ air}%
\MapoToken{0.694,0.878,0.671}{-r}%
\MapoToken{0.757,0.902,0.729}{ushing}%
\MapoToken{0.796,0.922,0.773}{ sound}%
\MapoToken{0.875,0.953,0.855}{ rather}%
\MapoToken{0.816,0.929,0.792}{ than}%
\MapoToken{0.796,0.922,0.773}{ typical}%
\MapoToken{0.784,0.914,0.757}{ factory}%
\MapoToken{0.788,0.918,0.761}{ noises}%
\MapoToken{0.831,0.933,0.808}{,}%
\MapoToken{0.843,0.937,0.820}{ making}%
\MapoToken{0.839,0.937,0.816}{ it}%
\MapoToken{0.831,0.933,0.808}{ less}%
\MapoToken{0.855,0.941,0.831}{ likely}%
\MapoToken{0.769,0.910,0.741}{;}%
\MapoToken{0.745,0.898,0.722}{ Wind}%
\MapoToken{0.729,0.890,0.702}{ turbine}%
\MapoToken{0.761,0.906,0.733}{ produces}%
\MapoToken{0.780,0.914,0.753}{ a}%
\MapoToken{0.729,0.890,0.702}{ '}%
\MapoToken{0.733,0.894,0.706}{who}%
\MapoToken{0.761,0.906,0.733}{osh}%
\MapoToken{0.824,0.929,0.800}{ing}%
\MapoToken{0.816,0.929,0.792}{'}%
\MapoToken{0.804,0.925,0.780}{ sound}%
\MapoToken{0.820,0.929,0.796}{ from}%
\MapoToken{0.820,0.929,0.796}{ large}%
\MapoToken{0.816,0.929,0.792}{ blades}%
\MapoToken{0.820,0.929,0.796}{ rotating}%
\MapoToken{0.804,0.925,0.780}{ in}%
\MapoToken{0.804,0.925,0.780}{ wind}%
\MapoToken{0.831,0.933,0.808}{ with}%
\MapoToken{0.835,0.937,0.812}{ a}%
\MapoToken{0.804,0.925,0.780}{ rhyth}%
\MapoToken{0.835,0.937,0.812}{mic}%
\MapoToken{0.780,0.914,0.753}{ '}%
\MapoToken{0.808,0.925,0.784}{th}%
\MapoToken{0.796,0.922,0.773}{ump}%
\MapoToken{0.820,0.929,0.796}{'}%
\MapoToken{0.867,0.949,0.847}{ or}%
\MapoToken{0.796,0.922,0.773}{ '}%
\MapoToken{0.827,0.933,0.804}{fl}%
\MapoToken{0.839,0.937,0.816}{ap}%
\MapoToken{0.831,0.933,0.808}{'}%
\MapoToken{0.859,0.945,0.839}{ from}%
\MapoToken{0.824,0.929,0.800}{ blade}%
\MapoToken{0.835,0.937,0.812}{-pass}%
\MapoToken{0.859,0.945,0.835}{ing}%
\MapoToken{0.824,0.929,0.800}{,}%
\MapoToken{0.847,0.941,0.824}{ which}%
\MapoToken{0.847,0.941,0.824}{ strongly}%
\MapoToken{0.808,0.925,0.784}{ matches}%
\MapoToken{0.722,0.890,0.698}{ the}%
\MapoToken{0.722,0.890,0.698}{ audio}%
\MapoToken{0.812,0.925,0.788}{'s}%
\MapoToken{0.847,0.941,0.824}{ combination}%
\MapoToken{0.812,0.925,0.788}{ of}%
\MapoToken{0.733,0.894,0.706}{ low}%
\MapoToken{0.733,0.894,0.706}{-frequency}%
\MapoToken{0.784,0.914,0.757}{ roar}%
\MapoToken{0.835,0.937,0.812}{ and}%
\MapoToken{0.533,0.808,0.529}{ rhyth}%
\MapoToken{0.800,0.922,0.776}{mic}%
\MapoToken{0.796,0.918,0.769}{ who}%
\MapoToken{0.796,0.922,0.773}{osh}%
\MapoToken{0.808,0.925,0.784}{ing}%
\MapoToken{0.776,0.910,0.749}{;}%
\MapoToken{0.733,0.894,0.706}{ Car}%
\MapoToken{0.741,0.898,0.714}{ engine}%
\MapoToken{0.804,0.925,0.780}{ generates}%
\MapoToken{0.784,0.914,0.757}{ a}%
\MapoToken{0.729,0.890,0.702}{ rum}%
\MapoToken{0.796,0.918,0.769}{ble}%
\MapoToken{0.816,0.929,0.792}{ or}%
\MapoToken{0.827,0.933,0.804}{ hum}%
\MapoToken{0.843,0.937,0.820}{ but}%
\MapoToken{0.839,0.937,0.816}{ lacks}%
\MapoToken{0.827,0.933,0.804}{ the}%
\MapoToken{0.808,0.925,0.784}{ massive}%
\MapoToken{0.824,0.929,0.800}{ scale}%
\MapoToken{0.867,0.949,0.847}{ and}%
\MapoToken{0.855,0.941,0.831}{ distinct}%
\MapoToken{0.804,0.925,0.780}{ air}%
\MapoToken{0.757,0.902,0.729}{-r}%
\MapoToken{0.831,0.933,0.808}{ushing}%
\MapoToken{0.859,0.945,0.839}{ quality}%
\MapoToken{0.847,0.941,0.824}{,}%
\MapoToken{0.820,0.929,0.796}{ with}%
\MapoToken{0.769,0.910,0.741}{ even}%
\MapoToken{0.776,0.910,0.749}{ high}%
\MapoToken{0.753,0.902,0.725}{-performance}%
\MapoToken{0.827,0.933,0.804}{ engines}%
\MapoToken{0.863,0.949,0.843}{ having}%
\MapoToken{0.851,0.941,0.827}{ a}%
\MapoToken{0.859,0.945,0.835}{ different}%
\MapoToken{0.835,0.937,0.812}{ sound}%
\MapoToken{0.855,0.941,0.831}{ profile}%
\MapoToken{0.816,0.929,0.792}{;}%
\MapoToken{0.757,0.902,0.729}{ Air}%
\MapoToken{0.737,0.894,0.710}{plane}%
\MapoToken{0.804,0.925,0.780}{,}%
\MapoToken{0.788,0.918,0.761}{ particularly}%
\MapoToken{0.718,0.886,0.694}{ a}%
\MapoToken{0.718,0.886,0.694}{ jet}%
\MapoToken{0.682,0.871,0.655}{ engine}%
\MapoToken{0.812,0.925,0.788}{,}%
\MapoToken{0.812,0.925,0.788}{ creates}%
\MapoToken{0.796,0.918,0.769}{ an}%
\MapoToken{0.769,0.910,0.741}{ extremely}%
\MapoToken{0.792,0.918,0.765}{ loud}%
\MapoToken{0.792,0.918,0.765}{ high}%
\MapoToken{0.729,0.890,0.702}{-p}%
\MapoToken{0.812,0.925,0.788}{itched}%
\MapoToken{0.800,0.922,0.776}{ roar}%
\MapoToken{0.835,0.937,0.812}{ with}%
\MapoToken{0.843,0.937,0.820}{ a}%
\MapoToken{0.796,0.922,0.773}{ '}%
\MapoToken{0.816,0.929,0.792}{s}%
\MapoToken{0.827,0.933,0.804}{cream}%
\MapoToken{0.855,0.941,0.831}{'}%
\MapoToken{0.863,0.949,0.843}{ from}%
\MapoToken{0.851,0.941,0.827}{ high}%
\MapoToken{0.835,0.937,0.812}{-speed}%
\MapoToken{0.843,0.937,0.820}{ air}%
\MapoToken{0.847,0.941,0.824}{ compression}%
\MapoToken{0.859,0.945,0.835}{,}%
\MapoToken{0.847,0.941,0.824}{ but}%
\MapoToken{0.780,0.914,0.753}{ the}%
\MapoToken{0.776,0.910,0.749}{ audio}%
\MapoToken{0.827,0.933,0.804}{ is}%
\MapoToken{0.796,0.922,0.773}{ deeper}%
\MapoToken{0.863,0.949,0.843}{ and}%
\MapoToken{0.847,0.941,0.824}{ more}%
\MapoToken{0.796,0.918,0.769}{ '}%
\MapoToken{0.804,0.925,0.780}{who}%
\MapoToken{0.796,0.922,0.773}{os}%
\MapoToken{0.859,0.945,0.839}{hy}%
\MapoToken{0.867,0.949,0.847}{'}%
\MapoToken{0.867,0.949,0.847}{ than}%
\MapoToken{0.855,0.941,0.831}{ a}%
\MapoToken{0.847,0.941,0.824}{ typical}%
\MapoToken{0.816,0.929,0.792}{ jet}%
\MapoToken{0.871,0.949,0.851}{,}%
\MapoToken{0.855,0.941,0.831}{ sounding}%
\MapoToken{0.871,0.949,0.851}{ more}%
\MapoToken{0.847,0.941,0.824}{ like}%
\MapoToken{0.808,0.925,0.784}{ slow}%
\MapoToken{0.816,0.929,0.792}{-moving}%
\MapoToken{0.820,0.929,0.796}{ massive}%
\MapoToken{0.855,0.941,0.831}{ blades}%
\MapoToken{0.886,0.957,0.867}{ rather}%
\MapoToken{0.875,0.953,0.855}{ than}%
\MapoToken{0.859,0.945,0.839}{ high}%
\MapoToken{0.855,0.941,0.831}{-speed}%
\MapoToken{0.835,0.937,0.812}{ turbine}%
\MapoToken{0.843,0.937,0.820}{ compress}%
\MapoToken{0.859,0.945,0.839}{ors}%
\MapoToken{0.871,0.949,0.851}{,}%
\MapoToken{0.855,0.941,0.831}{ and}%
\MapoToken{0.859,0.945,0.835}{ while}%
\MapoToken{0.804,0.925,0.780}{ prop}%
\MapoToken{0.824,0.929,0.800}{eller}%
\MapoToken{0.847,0.941,0.824}{ planes}%
\MapoToken{0.875,0.953,0.855}{ have}%
\MapoToken{0.855,0.941,0.831}{ a}%
\MapoToken{0.867,0.949,0.847}{ distinct}%
\MapoToken{0.824,0.929,0.800}{ drone}%
\MapoToken{0.847,0.941,0.824}{,}%
\MapoToken{0.804,0.925,0.780}{ the}%
\MapoToken{0.812,0.925,0.788}{ audio}%
\MapoToken{0.851,0.941,0.827}{'s}%
\MapoToken{0.812,0.925,0.788}{ sheer}%
\MapoToken{0.839,0.937,0.816}{ volume}%
\MapoToken{0.859,0.945,0.835}{ and}%
\MapoToken{0.820,0.929,0.796}{ deep}%
\MapoToken{0.863,0.949,0.843}{ roar}%
\MapoToken{0.890,0.957,0.871}{ are}%
\MapoToken{0.886,0.957,0.867}{ more}%
\MapoToken{0.878,0.953,0.859}{ characteristic}%
\MapoToken{0.859,0.945,0.839}{ of}%
\MapoToken{0.820,0.929,0.796}{ jet}%
\MapoToken{0.867,0.949,0.847}{ engines}%
\MapoToken{0.894,0.961,0.875}{ without}%
\MapoToken{0.875,0.953,0.855}{ matching}%
\MapoToken{0.875,0.953,0.855}{ the}%
\MapoToken{0.867,0.949,0.847}{ specific}%
\MapoToken{0.859,0.945,0.835}{ sound}%
\MapoToken{0.875,0.953,0.855}{ profile}%
\MapoToken{0.804,0.925,0.780}{.}%
\MapoToken{0.800,0.922,0.776}{ Finally}%
\MapoToken{0.816,0.929,0.792}{,}%
\MapoToken{0.812,0.925,0.788}{ the}%
\MapoToken{0.847,0.941,0.824}{ most}%
\MapoToken{0.824,0.929,0.800}{ accurate}%
\MapoToken{0.824,0.929,0.800}{ description}%
\MapoToken{0.788,0.918,0.761}{ of}%
\MapoToken{0.639,0.855,0.616}{ the}%
\MapoToken{0.722,0.890,0.698}{ sound}%
\MapoToken{0.835,0.937,0.812}{ is}%
\MapoToken{0.835,0.937,0.812}{ that}%
\MapoToken{0.792,0.918,0.765}{ of}%
\MapoToken{0.722,0.890,0.698}{ a}%
\MapoToken{0.769,0.910,0.741}{ large}%
\MapoToken{0.780,0.914,0.753}{ wind}%
\MapoToken{0.769,0.910,0.741}{ turbine}%
\MapoToken{0.847,0.941,0.824}{ due}%
\MapoToken{0.855,0.941,0.831}{ to}%
\MapoToken{0.851,0.941,0.827}{ the}%
\MapoToken{0.859,0.945,0.835}{ combination}%
\MapoToken{0.843,0.937,0.820}{ of}%
\MapoToken{0.816,0.929,0.792}{ massive}%
\MapoToken{0.843,0.937,0.820}{ scale}%
\MapoToken{0.855,0.941,0.831}{,}%
\MapoToken{0.812,0.925,0.788}{ low}%
\MapoToken{0.816,0.929,0.792}{-frequency}%
\MapoToken{0.839,0.937,0.816}{ roar}%
\MapoToken{0.855,0.941,0.831}{,}%
\MapoToken{0.882,0.953,0.863}{ and}%
\MapoToken{0.878,0.953,0.859}{ distinct}%
\MapoToken{0.831,0.933,0.808}{ rhyth}%
\MapoToken{0.859,0.945,0.839}{mic}%
\MapoToken{0.808,0.925,0.784}{ '}%
\MapoToken{0.612,0.843,0.592}{who}%
\MapoToken{0.843,0.937,0.820}{osh}%
\MapoToken{0.871,0.949,0.851}{ing}%
\MapoToken{0.859,0.945,0.835}{'}%
\MapoToken{0.875,0.953,0.855}{ from}%
\MapoToken{0.878,0.953,0.859}{ large}%
\MapoToken{0.863,0.949,0.843}{ rotating}%
\MapoToken{0.867,0.949,0.847}{ blades}%
\MapoToken{0.867,0.949,0.847}{ moving}%
\MapoToken{0.859,0.945,0.839}{ through}%
\MapoToken{0.859,0.945,0.839}{ air}%
\MapoToken{0.824,0.929,0.800}{;}%
\MapoToken{0.867,0.949,0.847}{ therefore}%
\MapoToken{0.835,0.937,0.812}{,}%
\MapoToken{0.737,0.894,0.710}{ '}%
\MapoToken{0.737,0.894,0.710}{Wind}%
\MapoToken{0.808,0.925,0.784}{ turbine}%
\MapoToken{0.824,0.929,0.800}{'}%
\MapoToken{0.890,0.957,0.871}{ is}%
\MapoToken{0.875,0.953,0.855}{ the}%
\MapoToken{0.882,0.953,0.863}{ best}%
\MapoToken{0.875,0.953,0.855}{ fit}%
\MapoToken{0.886,0.957,0.867}{ among}%
\MapoToken{0.871,0.949,0.851}{ the}%
\MapoToken{0.867,0.949,0.847}{ given}%
\MapoToken{0.859,0.945,0.835}{ choices}%
\MapoToken{0.475,0.776,0.478}{. }%
\MapoToken{0.376,0.729,0.424}{\textless{}/think\textgreater{}}%
\MapoToken{0.384,0.733,0.427}{\phantom{x}}%
\MapoToken{0.612,0.843,0.592}{Wind}%
\MapoToken{0.667,0.867,0.643}{ turbine}%
\end{MapoTokenStream}
\MapoMetricRule

% \vfill
% \newpage

\MapoModelHeader{MAPO-No-Attn-Loss}{Correct}{MapoCorrect}{MapoCorrectBg}{Factory machinery}

\begin{MapoTokenStream}
\MapoToken{0.000,0.267,0.106}{\textless{}think\textgreater{}}%
\MapoToken{0.000,0.267,0.106}{\phantom{x}}%
\MapoToken{0.067,0.482,0.220}{Got}%
\MapoToken{0.314,0.698,0.392}{ it}%
\MapoToken{0.212,0.624,0.329}{,}%
\MapoToken{0.000,0.325,0.129}{ let}%
\MapoToken{0.180,0.588,0.302}{'s}%
\MapoToken{0.431,0.757,0.451}{ tackle}%
\MapoToken{0.000,0.267,0.106}{ the}%
\MapoToken{0.451,0.769,0.463}{ problem}%
\MapoToken{0.353,0.718,0.412}{.}%
\MapoToken{0.620,0.843,0.596}{ First}%
\MapoToken{0.000,0.267,0.106}{,}%
\MapoToken{0.020,0.443,0.184}{ the}%
\MapoToken{0.365,0.725,0.420}{ user}%
\MapoToken{0.624,0.847,0.600}{ wants}%
\MapoToken{0.000,0.267,0.106}{ to}%
\MapoToken{0.573,0.824,0.561}{ identify}%
\MapoToken{0.157,0.565,0.286}{ the}%
\MapoToken{0.325,0.706,0.400}{ source}%
\MapoToken{0.263,0.675,0.369}{ of}%
\MapoToken{0.000,0.318,0.125}{ mechanical}%
\MapoToken{0.063,0.478,0.216}{ sounds}%
\MapoToken{0.471,0.776,0.475}{ from}%
\MapoToken{0.000,0.267,0.106}{ the}%
\MapoToken{0.000,0.392,0.157}{ provided}%
\MapoToken{0.000,0.349,0.141}{ audio}%
\MapoToken{0.165,0.573,0.290}{ file}%
\MapoToken{0.722,0.890,0.698}{,}%
\MapoToken{0.761,0.906,0.733}{ with}%
\MapoToken{0.667,0.867,0.643}{ four}%
\MapoToken{0.729,0.890,0.702}{ choices}%
\MapoToken{0.769,0.910,0.741}{ given}%
\MapoToken{0.584,0.827,0.569}{:}%
\MapoToken{0.184,0.596,0.310}{ Factory}%
\MapoToken{0.204,0.616,0.325}{ machinery}%
\MapoToken{0.573,0.824,0.561}{,}%
\MapoToken{0.176,0.584,0.302}{ Wind}%
\MapoToken{0.165,0.573,0.290}{ turbine}%
\MapoToken{0.459,0.769,0.467}{,}%
\MapoToken{0.133,0.541,0.267}{ Car}%
\MapoToken{0.161,0.569,0.290}{ engine}%
\MapoToken{0.690,0.875,0.667}{,}%
\MapoToken{0.000,0.267,0.106}{ Air}%
\MapoToken{0.192,0.604,0.314}{plane}%
\MapoToken{0.592,0.831,0.573}{.}%
\MapoToken{0.667,0.867,0.643}{ Then}%
\MapoToken{0.384,0.733,0.427}{,}%
\MapoToken{0.000,0.267,0.106}{ the}%
\MapoToken{0.000,0.267,0.106}{ audio}%
\MapoToken{0.000,0.267,0.106}{ contains}%
\MapoToken{0.000,0.267,0.106}{ a}%
\MapoToken{0.000,0.267,0.106}{ loud}%
\MapoToken{0.000,0.267,0.106}{,}%
\MapoToken{0.000,0.267,0.106}{ continuous}%
\MapoToken{0.000,0.267,0.106}{,}%
\MapoToken{0.000,0.388,0.157}{ low}%
\MapoToken{0.000,0.302,0.122}{-frequency}%
\MapoToken{0.000,0.267,0.106}{ hum}%
\MapoToken{0.000,0.314,0.125}{ or}%
\MapoToken{0.078,0.494,0.227}{ roar}%
\MapoToken{0.051,0.471,0.212}{ with}%
\MapoToken{0.082,0.498,0.231}{ a}%
\MapoToken{0.051,0.471,0.212}{ distinct}%
\MapoToken{0.471,0.776,0.475}{ mechanical}%
\MapoToken{0.643,0.855,0.620}{ quality}%
\MapoToken{0.275,0.682,0.376}{,}%
\MapoToken{0.376,0.729,0.424}{ sounding}%
\MapoToken{0.357,0.722,0.416}{ like}%
\MapoToken{0.247,0.659,0.357}{ a}%
\MapoToken{0.514,0.796,0.510}{ large}%
\MapoToken{0.580,0.827,0.565}{,}%
\MapoToken{0.545,0.812,0.537}{ powerful}%
\MapoToken{0.624,0.847,0.600}{ machine}%
\MapoToken{0.506,0.792,0.506}{ with}%
\MapoToken{0.573,0.824,0.561}{ moving}%
\MapoToken{0.663,0.863,0.639}{ parts}%
\MapoToken{0.553,0.816,0.541}{,}%
\MapoToken{0.345,0.714,0.408}{ and}%
\MapoToken{0.400,0.741,0.435}{ there}%
\MapoToken{0.000,0.420,0.169}{'s}%
\MapoToken{0.000,0.267,0.106}{ also}%
\MapoToken{0.000,0.267,0.106}{ a}%
\MapoToken{0.129,0.537,0.267}{ higher}%
\MapoToken{0.553,0.816,0.541}{-p}%
\MapoToken{0.078,0.494,0.227}{itched}%
\MapoToken{0.247,0.659,0.357}{ wh}%
\MapoToken{0.498,0.788,0.498}{ir}%
\MapoToken{0.345,0.714,0.408}{ring}%
\MapoToken{0.212,0.624,0.329}{ or}%
\MapoToken{0.580,0.827,0.565}{ wh}%
\MapoToken{0.580,0.827,0.565}{ining}%
\MapoToken{0.220,0.631,0.337}{ sound}%
\MapoToken{0.525,0.800,0.522}{ layered}%
\MapoToken{0.498,0.788,0.498}{ on}%
\MapoToken{0.439,0.761,0.455}{ top}%
\MapoToken{0.525,0.800,0.522}{ of}%
\MapoToken{0.396,0.741,0.435}{ the}%
\MapoToken{0.439,0.761,0.455}{ main}%
\MapoToken{0.420,0.753,0.447}{ roar}%
\MapoToken{0.675,0.871,0.651}{,}%
\MapoToken{0.792,0.918,0.765}{ which}%
\MapoToken{0.741,0.898,0.714}{ is}%
\MapoToken{0.784,0.914,0.757}{ consistent}%
\MapoToken{0.690,0.875,0.667}{ with}%
\MapoToken{0.529,0.804,0.525}{ fans}%
\MapoToken{0.706,0.882,0.678}{,}%
\MapoToken{0.710,0.882,0.682}{ turbines}%
\MapoToken{0.388,0.737,0.431}{,}%
\MapoToken{0.792,0.918,0.765}{ or}%
\MapoToken{0.757,0.902,0.729}{ high}%
\MapoToken{0.706,0.882,0.678}{-speed}%
\MapoToken{0.757,0.902,0.729}{ rotating}%
\MapoToken{0.796,0.918,0.769}{ components}%
\MapoToken{0.663,0.863,0.639}{,}%
\MapoToken{0.714,0.886,0.686}{ giving}%
\MapoToken{0.635,0.851,0.612}{ an}%
\MapoToken{0.612,0.843,0.592}{ overall}%
\MapoToken{0.729,0.890,0.702}{ impression}%
\MapoToken{0.659,0.863,0.635}{ of}%
\MapoToken{0.655,0.859,0.627}{ industrial}%
\MapoToken{0.714,0.886,0.686}{-scale}%
\MapoToken{0.690,0.875,0.667}{ machinery}%
\MapoToken{0.714,0.886,0.686}{.}%
\MapoToken{0.808,0.925,0.784}{ Next}%
\MapoToken{0.757,0.902,0.729}{,}%
\MapoToken{0.796,0.922,0.773}{ evaluating}%
\MapoToken{0.765,0.906,0.737}{ the}%
\MapoToken{0.784,0.914,0.757}{ options}%
\MapoToken{0.741,0.898,0.714}{:}%
\MapoToken{0.643,0.855,0.620}{ Factory}%
\MapoToken{0.718,0.886,0.694}{ machinery}%
\MapoToken{0.784,0.914,0.757}{ is}%
\MapoToken{0.800,0.922,0.776}{ a}%
\MapoToken{0.784,0.914,0.757}{ very}%
\MapoToken{0.804,0.925,0.780}{ strong}%
\MapoToken{0.796,0.922,0.773}{ candidate}%
\MapoToken{0.706,0.882,0.678}{ because}%
\MapoToken{0.608,0.839,0.588}{ large}%
\MapoToken{0.675,0.871,0.651}{ factories}%
\MapoToken{0.729,0.890,0.702}{ often}%
\MapoToken{0.643,0.855,0.620}{ have}%
\MapoToken{0.694,0.878,0.671}{ heavy}%
\MapoToken{0.659,0.863,0.635}{ machinery}%
\MapoToken{0.545,0.812,0.537}{ like}%
\MapoToken{0.627,0.851,0.608}{ generators}%
\MapoToken{0.643,0.855,0.620}{,}%
\MapoToken{0.690,0.875,0.667}{ pumps}%
\MapoToken{0.337,0.710,0.404}{,}%
\MapoToken{0.722,0.890,0.698}{ or}%
\MapoToken{0.765,0.906,0.737}{ assembly}%
\MapoToken{0.769,0.910,0.741}{ line}%
\MapoToken{0.769,0.910,0.741}{ equipment}%
\MapoToken{0.796,0.918,0.769}{ that}%
\MapoToken{0.753,0.902,0.725}{ produce}%
\MapoToken{0.765,0.906,0.737}{ such}%
\MapoToken{0.608,0.839,0.588}{ a}%
\MapoToken{0.671,0.867,0.647}{ loud}%
\MapoToken{0.741,0.898,0.714}{,}%
\MapoToken{0.690,0.875,0.667}{ continuous}%
\MapoToken{0.761,0.906,0.733}{,}%
\MapoToken{0.741,0.898,0.714}{ multi}%
\MapoToken{0.753,0.902,0.725}{-layer}%
\MapoToken{0.769,0.910,0.741}{ed}%
\MapoToken{0.761,0.906,0.733}{ mechanical}%
\MapoToken{0.741,0.898,0.714}{ sound}%
\MapoToken{0.792,0.918,0.765}{ with}%
\MapoToken{0.745,0.898,0.722}{ a}%
\MapoToken{0.675,0.871,0.651}{ deep}%
\MapoToken{0.784,0.914,0.757}{ hum}%
\MapoToken{0.792,0.918,0.765}{ and}%
\MapoToken{0.804,0.925,0.780}{ higher}%
\MapoToken{0.780,0.914,0.753}{-frequency}%
\MapoToken{0.780,0.914,0.753}{ wh}%
\MapoToken{0.808,0.925,0.784}{ir}%
\MapoToken{0.827,0.933,0.804}{ring}%
\MapoToken{0.757,0.902,0.729}{;}%
\MapoToken{0.741,0.898,0.714}{ Wind}%
\MapoToken{0.729,0.890,0.702}{ turbine}%
\MapoToken{0.788,0.918,0.761}{ is}%
\MapoToken{0.800,0.922,0.776}{ less}%
\MapoToken{0.808,0.925,0.784}{ likely}%
\MapoToken{0.784,0.914,0.757}{ as}%
\MapoToken{0.769,0.910,0.741}{ it}%
\MapoToken{0.784,0.914,0.757}{ typically}%
\MapoToken{0.757,0.902,0.729}{ produces}%
\MapoToken{0.776,0.910,0.749}{ a}%
\MapoToken{0.698,0.878,0.675}{ rhyth}%
\MapoToken{0.757,0.902,0.729}{mic}%
\MapoToken{0.698,0.878,0.675}{ '}%
\MapoToken{0.761,0.906,0.733}{who}%
\MapoToken{0.710,0.882,0.682}{osh}%
\MapoToken{0.776,0.910,0.749}{'}%
\MapoToken{0.847,0.941,0.824}{ or}%
\MapoToken{0.796,0.922,0.773}{ '}%
\MapoToken{0.812,0.925,0.788}{sw}%
\MapoToken{0.820,0.929,0.796}{ish}%
\MapoToken{0.820,0.929,0.796}{'}%
\MapoToken{0.835,0.937,0.812}{ from}%
\MapoToken{0.792,0.918,0.765}{ blade}%
\MapoToken{0.831,0.933,0.808}{ movement}%
\MapoToken{0.816,0.929,0.792}{ without}%
\MapoToken{0.765,0.906,0.737}{ the}%
\MapoToken{0.722,0.890,0.698}{ constant}%
\MapoToken{0.796,0.918,0.769}{,}%
\MapoToken{0.737,0.894,0.710}{ intense}%
\MapoToken{0.808,0.925,0.784}{,}%
\MapoToken{0.847,0.941,0.824}{ and}%
\MapoToken{0.824,0.929,0.800}{ complex}%
\MapoToken{0.812,0.925,0.788}{ mechanical}%
\MapoToken{0.769,0.910,0.741}{ roar}%
\MapoToken{0.757,0.902,0.729}{ heard}%
\MapoToken{0.710,0.882,0.682}{ in}%
\MapoToken{0.224,0.635,0.341}{ the}%
\MapoToken{0.765,0.906,0.737}{ audio}%
\MapoToken{0.804,0.925,0.780}{;}%
\MapoToken{0.714,0.886,0.686}{ Car}%
\MapoToken{0.733,0.894,0.706}{ engine}%
\MapoToken{0.722,0.890,0.698}{ doesn}%
\MapoToken{0.816,0.929,0.792}{'t}%
\MapoToken{0.843,0.937,0.820}{ match}%
\MapoToken{0.800,0.922,0.776}{ since}%
\MapoToken{0.769,0.910,0.741}{ it}%
\MapoToken{0.824,0.929,0.800}{ usually}%
\MapoToken{0.820,0.929,0.796}{ has}%
\MapoToken{0.796,0.918,0.769}{ a}%
\MapoToken{0.831,0.933,0.808}{ more}%
\MapoToken{0.769,0.910,0.741}{ puls}%
\MapoToken{0.824,0.929,0.800}{ating}%
\MapoToken{0.843,0.937,0.820}{ or}%
\MapoToken{0.792,0.918,0.765}{ rhyth}%
\MapoToken{0.835,0.937,0.812}{mic}%
\MapoToken{0.847,0.941,0.824}{ sound}%
\MapoToken{0.843,0.937,0.820}{ with}%
\MapoToken{0.796,0.918,0.769}{ variations}%
\MapoToken{0.800,0.922,0.776}{ in}%
\MapoToken{0.808,0.925,0.784}{ pitch}%
\MapoToken{0.835,0.937,0.812}{ and}%
\MapoToken{0.796,0.918,0.769}{ volume}%
\MapoToken{0.859,0.945,0.839}{ corresponding}%
\MapoToken{0.831,0.933,0.808}{ to}%
\MapoToken{0.796,0.918,0.769}{ acceleration}%
\MapoToken{0.839,0.937,0.816}{ and}%
\MapoToken{0.204,0.616,0.325}{ dec}%
\MapoToken{0.792,0.918,0.765}{eler}%
\MapoToken{0.816,0.929,0.792}{ation}%
\MapoToken{0.851,0.941,0.827}{,}%
\MapoToken{0.800,0.922,0.776}{ whereas}%
\MapoToken{0.671,0.867,0.647}{ the}%
\MapoToken{0.714,0.886,0.686}{ audio}%
\MapoToken{0.757,0.902,0.729}{ is}%
\MapoToken{0.643,0.855,0.620}{ a}%
\MapoToken{0.710,0.882,0.682}{ steady}%
\MapoToken{0.780,0.914,0.753}{,}%
\MapoToken{0.808,0.925,0.784}{ non}%
\MapoToken{0.639,0.855,0.616}{-v}%
\MapoToken{0.769,0.910,0.741}{ary}%
\MapoToken{0.780,0.914,0.753}{ing}%
\MapoToken{0.753,0.902,0.725}{ roar}%
\MapoToken{0.824,0.929,0.800}{;}%
\MapoToken{0.757,0.902,0.729}{ Air}%
\MapoToken{0.729,0.890,0.702}{plane}%
\MapoToken{0.835,0.937,0.812}{ is}%
\MapoToken{0.878,0.953,0.859}{ also}%
\MapoToken{0.859,0.945,0.835}{ improbable}%
\MapoToken{0.831,0.933,0.808}{ because}%
\MapoToken{0.824,0.929,0.800}{ while}%
\MapoToken{0.753,0.902,0.725}{ a}%
\MapoToken{0.722,0.890,0.698}{ jet}%
\MapoToken{0.765,0.906,0.737}{ engine}%
\MapoToken{0.831,0.933,0.808}{ produces}%
\MapoToken{0.804,0.925,0.780}{ a}%
\MapoToken{0.812,0.925,0.788}{ loud}%
\MapoToken{0.804,0.925,0.780}{ continuous}%
\MapoToken{0.812,0.925,0.788}{ roar}%
\MapoToken{0.851,0.941,0.827}{,}%
\MapoToken{0.812,0.925,0.788}{ the}%
\MapoToken{0.765,0.906,0.737}{ audio}%
\MapoToken{0.843,0.937,0.820}{ has}%
\MapoToken{0.839,0.937,0.816}{ more}%
\MapoToken{0.843,0.937,0.820}{ mechanical}%
\MapoToken{0.835,0.937,0.812}{ and}%
\MapoToken{0.553,0.816,0.541}{ cl}%
\MapoToken{0.784,0.914,0.757}{unky}%
\MapoToken{0.878,0.953,0.859}{ characteristics}%
\MapoToken{0.847,0.941,0.824}{ suggesting}%
\MapoToken{0.839,0.937,0.816}{ moving}%
\MapoToken{0.867,0.949,0.847}{ parts}%
\MapoToken{0.835,0.937,0.812}{ like}%
\MapoToken{0.843,0.937,0.820}{ gears}%
\MapoToken{0.816,0.929,0.792}{ or}%
\MapoToken{0.804,0.925,0.780}{ pist}%
\MapoToken{0.859,0.945,0.835}{ons}%
\MapoToken{0.898,0.961,0.878}{ rather}%
\MapoToken{0.882,0.953,0.863}{ than}%
\MapoToken{0.882,0.953,0.863}{ the}%
\MapoToken{0.863,0.949,0.843}{ smooth}%
\MapoToken{0.839,0.937,0.816}{ high}%
\MapoToken{0.780,0.914,0.753}{-p}%
\MapoToken{0.855,0.941,0.831}{itched}%
\MapoToken{0.827,0.933,0.804}{ wh}%
\MapoToken{0.855,0.941,0.831}{ine}%
\MapoToken{0.875,0.953,0.855}{ of}%
\MapoToken{0.882,0.953,0.863}{ a}%
\MapoToken{0.859,0.945,0.835}{ jet}%
\MapoToken{0.863,0.949,0.843}{,}%
\MapoToken{0.843,0.937,0.820}{ and}%
\MapoToken{0.769,0.910,0.741}{ the}%
\MapoToken{0.761,0.906,0.733}{ sound}%
\MapoToken{0.851,0.941,0.827}{ profile}%
\MapoToken{0.000,0.267,0.106}{ align}%
\MapoToken{0.882,0.953,0.863}{s}%
\MapoToken{0.882,0.953,0.863}{ more}%
\MapoToken{0.831,0.933,0.808}{ with}%
\MapoToken{0.800,0.922,0.776}{ a}%
\MapoToken{0.808,0.925,0.784}{ large}%
\MapoToken{0.800,0.922,0.776}{ piston}%
\MapoToken{0.816,0.929,0.792}{ engine}%
\MapoToken{0.835,0.937,0.812}{ or}%
\MapoToken{0.839,0.937,0.816}{ industrial}%
\MapoToken{0.843,0.937,0.820}{ turbine}%
\MapoToken{0.820,0.929,0.796}{.}%
\MapoToken{0.812,0.925,0.788}{ Finally}%
\MapoToken{0.824,0.929,0.800}{,}%
\MapoToken{0.847,0.941,0.824}{ based}%
\MapoToken{0.843,0.937,0.820}{ on}%
\MapoToken{0.812,0.925,0.788}{ the}%
\MapoToken{0.710,0.882,0.682}{ loud}%
\MapoToken{0.808,0.925,0.784}{,}%
\MapoToken{0.788,0.918,0.761}{ continuous}%
\MapoToken{0.827,0.933,0.804}{,}%
\MapoToken{0.776,0.910,0.749}{ low}%
\MapoToken{0.788,0.918,0.761}{-frequency}%
\MapoToken{0.820,0.929,0.796}{ hum}%
\MapoToken{0.859,0.945,0.835}{ combined}%
\MapoToken{0.827,0.933,0.804}{ with}%
\MapoToken{0.824,0.929,0.800}{ higher}%
\MapoToken{0.765,0.906,0.737}{-p}%
\MapoToken{0.827,0.933,0.804}{itched}%
\MapoToken{0.824,0.929,0.800}{ mechanical}%
\MapoToken{0.820,0.929,0.796}{ wh}%
\MapoToken{0.839,0.937,0.816}{ir}%
\MapoToken{0.859,0.945,0.835}{ring}%
\MapoToken{0.871,0.949,0.851}{ characteristic}%
\MapoToken{0.855,0.941,0.831}{ of}%
\MapoToken{0.835,0.937,0.812}{ heavy}%
\MapoToken{0.831,0.933,0.808}{ industrial}%
\MapoToken{0.835,0.937,0.812}{ equipment}%
\MapoToken{0.843,0.937,0.820}{,}%
\MapoToken{0.827,0.933,0.804}{ the}%
\MapoToken{0.859,0.945,0.839}{ most}%
\MapoToken{0.855,0.941,0.831}{ fitting}%
\MapoToken{0.843,0.937,0.820}{ description}%
\MapoToken{0.808,0.925,0.784}{ for}%
\MapoToken{0.718,0.886,0.694}{ the}%
\MapoToken{0.796,0.918,0.769}{ source}%
\MapoToken{0.831,0.933,0.808}{ is}%
\MapoToken{0.769,0.910,0.741}{ Factory}%
\MapoToken{0.831,0.933,0.808}{ machinery}%
\MapoToken{0.506,0.792,0.506}{. }%
\MapoToken{0.416,0.749,0.443}{\textless{}/think\textgreater{}}%
\MapoToken{0.388,0.737,0.431}{\phantom{x}}%
\MapoToken{0.635,0.851,0.612}{Factory}%
\MapoToken{0.694,0.878,0.671}{ machinery}%
\end{MapoTokenStream}
\MapoMetricRule

% \vfill
% \newpage

\MapoModelHeader{Target-MAPO+Attn-Loss}{Correct}{MapoCorrect}{MapoCorrectBg}{Factory machinery}

\begin{MapoTokenStream}
\MapoToken{0.000,0.396,0.161}{\textless{}think\textgreater{}}%
\MapoToken{0.008,0.435,0.180}{\phantom{x}}%
\MapoToken{0.047,0.467,0.208}{Got}%
\MapoToken{0.149,0.557,0.278}{ it}%
\MapoToken{0.125,0.533,0.263}{,}%
\MapoToken{0.000,0.267,0.106}{ let}%
\MapoToken{0.180,0.588,0.302}{'s}%
\MapoToken{0.102,0.518,0.247}{ tackle}%
\MapoToken{0.000,0.267,0.106}{ the}%
\MapoToken{0.090,0.506,0.239}{ problems}%
\MapoToken{0.129,0.537,0.267}{.}%
\MapoToken{0.125,0.533,0.263}{ First}%
\MapoToken{0.000,0.388,0.157}{,}%
\MapoToken{0.067,0.482,0.220}{ the}%
\MapoToken{0.133,0.541,0.267}{ user}%
\MapoToken{0.157,0.565,0.286}{ wants}%
\MapoToken{0.122,0.529,0.259}{ me}%
\MapoToken{0.192,0.604,0.314}{ to}%
\MapoToken{0.102,0.518,0.247}{ identify}%
\MapoToken{0.008,0.435,0.180}{ the}%
\MapoToken{0.000,0.373,0.149}{ source}%
\MapoToken{0.000,0.427,0.173}{ of}%
\MapoToken{0.000,0.267,0.106}{ the}%
\MapoToken{0.000,0.267,0.106}{ mechanical}%
\MapoToken{0.000,0.267,0.106}{ sounds}%
\MapoToken{0.000,0.376,0.153}{ in}%
\MapoToken{0.000,0.267,0.106}{ the}%
\MapoToken{0.000,0.318,0.125}{ provided}%
\MapoToken{0.000,0.333,0.133}{ audio}%
\MapoToken{0.000,0.392,0.157}{ clip}%
\MapoToken{0.216,0.627,0.333}{ and}%
\MapoToken{0.200,0.612,0.322}{ must}%
\MapoToken{0.153,0.561,0.282}{ choose}%
\MapoToken{0.200,0.612,0.322}{ from}%
\MapoToken{0.000,0.267,0.106}{ the}%
\MapoToken{0.231,0.643,0.345}{ given}%
\MapoToken{0.090,0.506,0.239}{ options}%
\MapoToken{0.184,0.592,0.306}{:}%
\MapoToken{0.000,0.392,0.157}{ Factory}%
\MapoToken{0.000,0.271,0.110}{ machinery}%
\MapoToken{0.122,0.529,0.259}{,}%
\MapoToken{0.000,0.408,0.165}{ Wind}%
\MapoToken{0.000,0.267,0.106}{ turbine}%
\MapoToken{0.067,0.482,0.220}{,}%
\MapoToken{0.000,0.388,0.157}{ Car}%
\MapoToken{0.000,0.267,0.106}{ engine}%
\MapoToken{0.224,0.635,0.341}{,}%
\MapoToken{0.212,0.624,0.329}{ or}%
\MapoToken{0.000,0.267,0.106}{ Air}%
\MapoToken{0.000,0.322,0.129}{plane}%
\MapoToken{0.067,0.482,0.220}{.}%
\MapoToken{0.039,0.463,0.200}{ Then}%
\MapoToken{0.000,0.408,0.165}{,}%
\MapoToken{0.000,0.267,0.106}{ the}%
\MapoToken{0.000,0.267,0.106}{ audio}%
\MapoToken{0.000,0.267,0.106}{ contains}%
\MapoToken{0.000,0.267,0.106}{ a}%
\MapoToken{0.000,0.267,0.106}{ loud}%
\MapoToken{0.000,0.267,0.106}{,}%
\MapoToken{0.000,0.267,0.106}{ continuous}%
\MapoToken{0.000,0.267,0.106}{,}%
\MapoToken{0.000,0.318,0.125}{ and}%
\MapoToken{0.000,0.267,0.106}{ somewhat}%
\MapoToken{0.000,0.267,0.106}{ harsh}%
\MapoToken{0.000,0.298,0.118}{ mechanical}%
\MapoToken{0.000,0.267,0.106}{ sound}%
\MapoToken{0.000,0.306,0.122}{ that}%
\MapoToken{0.000,0.333,0.133}{ is}%
\MapoToken{0.000,0.341,0.137}{ a}%
\MapoToken{0.000,0.290,0.118}{ low}%
\MapoToken{0.012,0.439,0.180}{-to}%
\MapoToken{0.000,0.412,0.165}{-m}%
\MapoToken{0.000,0.424,0.173}{id}%
\MapoToken{0.000,0.361,0.145}{ frequency}%
\MapoToken{0.000,0.267,0.106}{ drone}%
\MapoToken{0.000,0.384,0.153}{ or}%
\MapoToken{0.000,0.286,0.114}{ roar}%
\MapoToken{0.000,0.267,0.106}{ with}%
\MapoToken{0.000,0.353,0.141}{ a}%
\MapoToken{0.000,0.298,0.118}{ distinct}%
\MapoToken{0.000,0.357,0.145}{ rotational}%
\MapoToken{0.012,0.439,0.180}{ or}%
\MapoToken{0.055,0.475,0.212}{ cyc}%
\MapoToken{0.000,0.373,0.149}{lical}%
\MapoToken{0.000,0.290,0.118}{ quality}%
\MapoToken{0.000,0.349,0.141}{,}%
\MapoToken{0.078,0.494,0.227}{ suggesting}%
\MapoToken{0.176,0.584,0.302}{ a}%
\MapoToken{0.161,0.569,0.290}{ large}%
\MapoToken{0.086,0.502,0.235}{ machine}%
\MapoToken{0.149,0.557,0.278}{ with}%
\MapoToken{0.176,0.584,0.302}{ moving}%
\MapoToken{0.063,0.478,0.216}{ parts}%
\MapoToken{0.000,0.298,0.118}{;}%
\MapoToken{0.000,0.290,0.118}{ it}%
\MapoToken{0.000,0.314,0.125}{ has}%
\MapoToken{0.000,0.412,0.165}{ a}%
\MapoToken{0.020,0.443,0.184}{ '}%
\MapoToken{0.000,0.404,0.161}{wh}%
\MapoToken{0.000,0.404,0.161}{ir}%
\MapoToken{0.000,0.314,0.125}{ring}%
\MapoToken{0.000,0.376,0.153}{'}%
\MapoToken{0.004,0.431,0.176}{ or}%
\MapoToken{0.047,0.467,0.208}{ '}%
\MapoToken{0.078,0.494,0.227}{ro}%
\MapoToken{0.000,0.376,0.153}{aring}%
\MapoToken{0.039,0.463,0.200}{'}%
\MapoToken{0.000,0.420,0.169}{ character}%
\MapoToken{0.000,0.396,0.161}{,}%
\MapoToken{0.000,0.369,0.149}{ is}%
\MapoToken{0.000,0.408,0.165}{ very}%
\MapoToken{0.000,0.267,0.106}{ loud}%
\MapoToken{0.000,0.408,0.165}{ and}%
\MapoToken{0.000,0.369,0.149}{ powerful}%
\MapoToken{0.000,0.322,0.129}{,}%
\MapoToken{0.000,0.333,0.133}{ and}%
\MapoToken{0.000,0.412,0.165}{ the}%
\MapoToken{0.000,0.267,0.106}{ sound}%
\MapoToken{0.000,0.376,0.153}{ quality}%
\MapoToken{0.000,0.424,0.173}{ is}%
\MapoToken{0.000,0.427,0.173}{ somewhat}%
\MapoToken{0.114,0.525,0.251}{ '}%
\MapoToken{0.000,0.427,0.173}{raw}%
\MapoToken{0.000,0.404,0.161}{'}%
\MapoToken{0.071,0.486,0.224}{ or}%
\MapoToken{0.004,0.431,0.176}{ industrial}%
\MapoToken{0.035,0.459,0.196}{,}%
\MapoToken{0.114,0.525,0.251}{ lacking}%
\MapoToken{0.204,0.616,0.325}{ the}%
\MapoToken{0.165,0.573,0.290}{ smooth}%
\MapoToken{0.090,0.506,0.239}{ness}%
\MapoToken{0.153,0.561,0.282}{ of}%
\MapoToken{0.263,0.675,0.369}{ a}%
\MapoToken{0.239,0.651,0.353}{ modern}%
\MapoToken{0.161,0.569,0.290}{ car}%
\MapoToken{0.082,0.498,0.231}{ engine}%
\MapoToken{0.216,0.627,0.333}{ or}%
\MapoToken{0.184,0.596,0.310}{ the}%
\MapoToken{0.145,0.553,0.278}{ high}%
\MapoToken{0.102,0.518,0.247}{-p}%
\MapoToken{0.129,0.537,0.267}{itched}%
\MapoToken{0.184,0.592,0.306}{ wh}%
\MapoToken{0.075,0.490,0.224}{ine}%
\MapoToken{0.161,0.569,0.290}{ of}%
\MapoToken{0.302,0.694,0.388}{ a}%
\MapoToken{0.192,0.604,0.314}{ jet}%
\MapoToken{0.129,0.537,0.267}{ airplane}%
\MapoToken{0.039,0.463,0.200}{ engine}%
\MapoToken{0.090,0.506,0.239}{.}%
\MapoToken{0.129,0.537,0.267}{ Next}%
\MapoToken{0.216,0.627,0.333}{,}%
\MapoToken{0.176,0.584,0.302}{ evaluating}%
\MapoToken{0.247,0.663,0.361}{ the}%
\MapoToken{0.122,0.529,0.259}{ options}%
\MapoToken{0.220,0.631,0.337}{:}%
\MapoToken{0.071,0.486,0.224}{ Factory}%
\MapoToken{0.004,0.431,0.176}{ machinery}%
\MapoToken{0.031,0.455,0.196}{ is}%
\MapoToken{0.129,0.537,0.267}{ a}%
\MapoToken{0.098,0.514,0.243}{ very}%
\MapoToken{0.122,0.529,0.259}{ strong}%
\MapoToken{0.024,0.447,0.188}{ candidate}%
\MapoToken{0.075,0.490,0.224}{ because}%
\MapoToken{0.067,0.482,0.220}{ factories}%
\MapoToken{0.086,0.502,0.235}{ often}%
\MapoToken{0.071,0.486,0.224}{ contain}%
\MapoToken{0.110,0.522,0.251}{ large}%
\MapoToken{0.180,0.588,0.302}{,}%
\MapoToken{0.157,0.565,0.286}{ powerful}%
\MapoToken{0.000,0.427,0.173}{ machines}%
\MapoToken{0.082,0.498,0.231}{ like}%
\MapoToken{0.000,0.424,0.173}{ pumps}%
\MapoToken{0.173,0.580,0.298}{,}%
\MapoToken{0.035,0.459,0.196}{ generators}%
\MapoToken{0.047,0.467,0.208}{,}%
\MapoToken{0.129,0.537,0.267}{ or}%
\MapoToken{0.137,0.545,0.271}{ industrial}%
\MapoToken{0.020,0.443,0.184}{ fans}%
\MapoToken{0.094,0.510,0.239}{ that}%
\MapoToken{0.110,0.522,0.251}{ produce}%
\MapoToken{0.114,0.525,0.251}{ loud}%
\MapoToken{0.165,0.573,0.290}{,}%
\MapoToken{0.141,0.549,0.275}{ continuous}%
\MapoToken{0.196,0.608,0.318}{,}%
\MapoToken{0.129,0.537,0.267}{ mechanical}%
\MapoToken{0.055,0.475,0.212}{ sounds}%
\MapoToken{0.157,0.565,0.286}{ with}%
\MapoToken{0.180,0.588,0.302}{ a}%
\MapoToken{0.110,0.522,0.251}{ rotational}%
\MapoToken{0.125,0.533,0.263}{ or}%
\MapoToken{0.000,0.369,0.149}{ rhyth}%
\MapoToken{0.090,0.506,0.239}{mic}%
\MapoToken{0.024,0.447,0.188}{ nature}%
\MapoToken{0.039,0.463,0.200}{,}%
\MapoToken{0.043,0.467,0.204}{ and}%
\MapoToken{0.067,0.482,0.220}{ the}%
\MapoToken{0.047,0.467,0.208}{ sound}%
\MapoToken{0.027,0.451,0.192}{ in}%
\MapoToken{0.000,0.322,0.129}{ the}%
\MapoToken{0.051,0.471,0.212}{ audio}%
\MapoToken{0.114,0.525,0.251}{ fits}%
\MapoToken{0.161,0.569,0.290}{ this}%
\MapoToken{0.082,0.498,0.231}{ description}%
\MapoToken{0.024,0.447,0.188}{ perfectly}%
\MapoToken{0.043,0.467,0.204}{ as}%
\MapoToken{0.086,0.502,0.235}{ a}%
\MapoToken{0.110,0.522,0.251}{ classic}%
\MapoToken{0.078,0.494,0.227}{ industrial}%
\MapoToken{0.012,0.439,0.180}{ sound}%
\MapoToken{0.200,0.612,0.322}{;}%
\MapoToken{0.184,0.596,0.310}{ Wind}%
\MapoToken{0.086,0.502,0.235}{ turbine}%
\MapoToken{0.090,0.506,0.239}{ is}%
\MapoToken{0.051,0.471,0.212}{ unlikely}%
\MapoToken{0.161,0.569,0.290}{ because}%
\MapoToken{0.176,0.584,0.302}{ while}%
\MapoToken{0.110,0.522,0.251}{ it}%
\MapoToken{0.157,0.565,0.286}{ has}%
\MapoToken{0.145,0.553,0.278}{ rotating}%
\MapoToken{0.098,0.514,0.243}{ blades}%
\MapoToken{0.161,0.569,0.290}{,}%
\MapoToken{0.157,0.565,0.286}{ the}%
\MapoToken{0.086,0.502,0.235}{ sound}%
\MapoToken{0.153,0.561,0.282}{ is}%
\MapoToken{0.169,0.576,0.294}{ typically}%
\MapoToken{0.208,0.620,0.325}{ a}%
\MapoToken{0.161,0.569,0.290}{ lower}%
\MapoToken{0.196,0.608,0.318}{-frequency}%
\MapoToken{0.290,0.686,0.380}{ '}%
\MapoToken{0.204,0.616,0.325}{who}%
\MapoToken{0.098,0.514,0.243}{osh}%
\MapoToken{0.114,0.525,0.251}{'}%
\MapoToken{0.251,0.667,0.365}{ or}%
\MapoToken{0.388,0.737,0.431}{ '}%
\MapoToken{0.333,0.710,0.404}{th}%
\MapoToken{0.149,0.557,0.278}{rum}%
\MapoToken{0.114,0.525,0.251}{'}%
\MapoToken{0.157,0.565,0.286}{ and}%
\MapoToken{0.165,0.573,0.290}{ the}%
\MapoToken{0.086,0.502,0.235}{ audio}%
\MapoToken{0.055,0.475,0.212}{ has}%
\MapoToken{0.082,0.498,0.231}{ a}%
\MapoToken{0.149,0.557,0.278}{ more}%
\MapoToken{0.129,0.537,0.267}{ mechanical}%
\MapoToken{0.192,0.604,0.314}{,}%
\MapoToken{0.122,0.529,0.259}{ metallic}%
\MapoToken{0.055,0.475,0.212}{,}%
\MapoToken{0.161,0.569,0.290}{ and}%
\MapoToken{0.133,0.541,0.267}{ harsh}%
\MapoToken{0.157,0.565,0.286}{er}%
\MapoToken{0.098,0.514,0.243}{ sound}%
\MapoToken{0.149,0.557,0.278}{,}%
\MapoToken{0.192,0.604,0.314}{ and}%
\MapoToken{0.165,0.573,0.290}{ wind}%
\MapoToken{0.102,0.518,0.247}{ turbines}%
\MapoToken{0.169,0.576,0.294}{ are}%
\MapoToken{0.165,0.573,0.290}{ generally}%
\MapoToken{0.125,0.533,0.263}{ quieter}%
\MapoToken{0.231,0.643,0.345}{ and}%
\MapoToken{0.188,0.600,0.314}{ the}%
\MapoToken{0.129,0.537,0.267}{ sound}%
\MapoToken{0.161,0.569,0.290}{ is}%
\MapoToken{0.192,0.604,0.314}{ more}%
\MapoToken{0.145,0.553,0.278}{ air}%
\MapoToken{0.133,0.541,0.267}{-based}%
\MapoToken{0.192,0.604,0.314}{ than}%
\MapoToken{0.125,0.533,0.263}{ mechanical}%
\MapoToken{0.227,0.639,0.341}{;}%
\MapoToken{0.169,0.576,0.294}{ Car}%
\MapoToken{0.086,0.502,0.235}{ engine}%
\MapoToken{0.118,0.529,0.255}{ is}%
\MapoToken{0.094,0.510,0.239}{ unlikely}%
\MapoToken{0.200,0.612,0.322}{ because}%
\MapoToken{0.204,0.616,0.325}{ a}%
\MapoToken{0.173,0.580,0.298}{ car}%
\MapoToken{0.149,0.557,0.278}{ engine}%
\MapoToken{0.184,0.592,0.306}{ has}%
\MapoToken{0.259,0.671,0.365}{ a}%
\MapoToken{0.247,0.663,0.361}{ more}%
\MapoToken{0.231,0.643,0.345}{ distinct}%
\MapoToken{0.271,0.678,0.373}{,}%
\MapoToken{0.306,0.698,0.392}{ often}%
\MapoToken{0.224,0.635,0.341}{ higher}%
\MapoToken{0.184,0.592,0.306}{-p}%
\MapoToken{0.247,0.659,0.357}{itched}%
\MapoToken{0.275,0.682,0.376}{,}%
\MapoToken{0.259,0.671,0.365}{ and}%
\MapoToken{0.263,0.675,0.369}{ more}%
\MapoToken{0.227,0.639,0.341}{ variable}%
\MapoToken{0.169,0.576,0.294}{ sound}%
\MapoToken{0.208,0.620,0.325}{ (}%
\MapoToken{0.184,0.596,0.310}{especially}%
\MapoToken{0.161,0.569,0.290}{ if}%
\MapoToken{0.125,0.533,0.263}{ it}%
\MapoToken{0.176,0.584,0.302}{'s}%
\MapoToken{0.314,0.698,0.392}{ a}%
\MapoToken{0.337,0.710,0.404}{ gasoline}%
\MapoToken{0.204,0.616,0.325}{ engine}%
\MapoToken{0.184,0.592,0.306}{)}%
\MapoToken{0.192,0.604,0.314}{ with}%
\MapoToken{0.275,0.682,0.376}{ a}%
\MapoToken{0.314,0.698,0.392}{ characteristic}%
\MapoToken{0.431,0.757,0.451}{ '}%
\MapoToken{0.282,0.682,0.376}{rum}%
\MapoToken{0.180,0.588,0.302}{ble}%
\MapoToken{0.165,0.573,0.290}{'}%
\MapoToken{0.376,0.729,0.424}{ or}%
\MapoToken{0.584,0.827,0.569}{ '}%
\MapoToken{0.161,0.569,0.290}{hum}%
\MapoToken{0.184,0.592,0.306}{'}%
\MapoToken{0.176,0.584,0.302}{ that}%
\MapoToken{0.216,0.627,0.333}{ is}%
\MapoToken{0.220,0.631,0.337}{ usually}%
\MapoToken{0.290,0.686,0.380}{ more}%
\MapoToken{0.149,0.557,0.278}{ contained}%
\MapoToken{0.275,0.682,0.376}{ and}%
\MapoToken{0.133,0.541,0.267}{ less}%
\MapoToken{0.133,0.541,0.267}{ '}%
\MapoToken{0.145,0.553,0.278}{ind}%
\MapoToken{0.125,0.533,0.263}{ustrial}%
\MapoToken{0.149,0.557,0.278}{'}%
\MapoToken{0.137,0.545,0.271}{ than}%
\MapoToken{0.129,0.537,0.267}{ the}%
\MapoToken{0.114,0.525,0.251}{ sound}%
\MapoToken{0.102,0.518,0.247}{ in}%
\MapoToken{0.047,0.467,0.208}{ the}%
\MapoToken{0.071,0.486,0.224}{ audio}%
\MapoToken{0.169,0.576,0.294}{,}%
\MapoToken{0.212,0.624,0.329}{ and}%
\MapoToken{0.176,0.584,0.302}{ the}%
\MapoToken{0.122,0.529,0.259}{ sound}%
\MapoToken{0.133,0.541,0.267}{ in}%
\MapoToken{0.098,0.514,0.243}{ the}%
\MapoToken{0.047,0.467,0.208}{ clip}%
\MapoToken{0.133,0.541,0.267}{ is}%
\MapoToken{0.161,0.569,0.290}{ too}%
\MapoToken{0.067,0.482,0.220}{ loud}%
\MapoToken{0.208,0.620,0.325}{ and}%
\MapoToken{0.137,0.545,0.271}{ mon}%
\MapoToken{0.090,0.506,0.239}{olithic}%
\MapoToken{0.125,0.533,0.263}{ for}%
\MapoToken{0.227,0.639,0.341}{ a}%
\MapoToken{0.251,0.667,0.365}{ typical}%
\MapoToken{0.180,0.588,0.302}{ car}%
\MapoToken{0.086,0.502,0.235}{ engine}%
\MapoToken{0.337,0.710,0.404}{;}%
\MapoToken{0.204,0.616,0.325}{ Air}%
\MapoToken{0.161,0.569,0.290}{plane}%
\MapoToken{0.165,0.573,0.290}{ is}%
\MapoToken{0.145,0.553,0.278}{ unlikely}%
\MapoToken{0.275,0.682,0.376}{ because}%
\MapoToken{0.345,0.714,0.408}{ a}%
\MapoToken{0.314,0.698,0.392}{ commercial}%
\MapoToken{0.212,0.624,0.329}{ airplane}%
\MapoToken{0.192,0.604,0.314}{ engine}%
\MapoToken{0.208,0.620,0.325}{ produces}%
\MapoToken{0.357,0.722,0.416}{ a}%
\MapoToken{0.337,0.710,0.404}{ very}%
\MapoToken{0.247,0.663,0.361}{ high}%
\MapoToken{0.216,0.627,0.333}{-p}%
\MapoToken{0.302,0.694,0.388}{itched}%
\MapoToken{0.353,0.718,0.412}{,}%
\MapoToken{0.306,0.698,0.392}{ powerful}%
\MapoToken{0.247,0.659,0.357}{ jet}%
\MapoToken{0.145,0.553,0.278}{ roar}%
\MapoToken{0.235,0.647,0.349}{,}%
\MapoToken{0.176,0.584,0.302}{ which}%
\MapoToken{0.231,0.643,0.345}{ is}%
\MapoToken{0.259,0.671,0.365}{ very}%
\MapoToken{0.165,0.573,0.290}{ different}%
\MapoToken{0.145,0.553,0.278}{ from}%
\MapoToken{0.216,0.627,0.333}{ the}%
\MapoToken{0.145,0.553,0.278}{ sound}%
\MapoToken{0.137,0.545,0.271}{ in}%
\MapoToken{0.243,0.655,0.353}{ the}%
\MapoToken{0.149,0.557,0.278}{ audio}%
\MapoToken{0.263,0.675,0.369}{,}%
\MapoToken{0.271,0.678,0.373}{ and}%
\MapoToken{0.365,0.725,0.420}{ a}%
\MapoToken{0.294,0.690,0.384}{ prop}%
\MapoToken{0.290,0.686,0.380}{eller}%
\MapoToken{0.251,0.667,0.365}{ plane}%
\MapoToken{0.239,0.651,0.353}{ would}%
\MapoToken{0.259,0.671,0.365}{ have}%
\MapoToken{0.408,0.745,0.439}{ a}%
\MapoToken{0.337,0.710,0.404}{ more}%
\MapoToken{0.345,0.714,0.408}{ distinct}%
\MapoToken{0.376,0.729,0.424}{,}%
\MapoToken{0.282,0.682,0.376}{ lower}%
\MapoToken{0.353,0.718,0.412}{-frequency}%
\MapoToken{0.271,0.678,0.373}{ prop}%
\MapoToken{0.263,0.675,0.369}{eller}%
\MapoToken{0.216,0.627,0.333}{ sound}%
\MapoToken{0.173,0.580,0.298}{ which}%
\MapoToken{0.188,0.600,0.314}{ is}%
\MapoToken{0.165,0.573,0.290}{ not}%
\MapoToken{0.125,0.533,0.263}{ present}%
\MapoToken{0.122,0.529,0.259}{ here}%
\MapoToken{0.216,0.627,0.333}{,}%
\MapoToken{0.243,0.655,0.353}{ and}%
\MapoToken{0.271,0.678,0.373}{ the}%
\MapoToken{0.173,0.580,0.298}{ sound}%
\MapoToken{0.180,0.588,0.302}{ in}%
\MapoToken{0.208,0.620,0.325}{ the}%
\MapoToken{0.145,0.553,0.278}{ audio}%
\MapoToken{0.216,0.627,0.333}{ lacks}%
\MapoToken{0.376,0.729,0.424}{ the}%
\MapoToken{0.357,0.722,0.416}{ specific}%
\MapoToken{0.294,0.690,0.384}{ high}%
\MapoToken{0.294,0.690,0.384}{-frequency}%
\MapoToken{0.208,0.620,0.325}{ components}%
\MapoToken{0.294,0.690,0.384}{ of}%
\MapoToken{0.408,0.745,0.439}{ a}%
\MapoToken{0.263,0.675,0.369}{ jet}%
\MapoToken{0.188,0.600,0.314}{ engine}%
\MapoToken{0.188,0.600,0.314}{.}%
\MapoToken{0.125,0.533,0.263}{ Finally}%
\MapoToken{0.216,0.627,0.333}{,}%
\MapoToken{0.231,0.643,0.345}{ the}%
\MapoToken{0.118,0.529,0.255}{ sound}%
\MapoToken{0.153,0.561,0.282}{ most}%
\MapoToken{0.180,0.588,0.302}{ closely}%
\MapoToken{0.184,0.592,0.306}{ matches}%
\MapoToken{0.200,0.612,0.322}{ the}%
\MapoToken{0.173,0.580,0.298}{ description}%
\MapoToken{0.208,0.620,0.325}{ of}%
\MapoToken{0.161,0.569,0.290}{ factory}%
\MapoToken{0.122,0.529,0.259}{ machinery}%
\MapoToken{0.125,0.533,0.263}{ due}%
\MapoToken{0.188,0.600,0.314}{ to}%
\MapoToken{0.294,0.690,0.384}{ its}%
\MapoToken{0.173,0.580,0.298}{ loud}%
\MapoToken{0.263,0.675,0.369}{,}%
\MapoToken{0.212,0.624,0.329}{ continuous}%
\MapoToken{0.337,0.710,0.404}{,}%
\MapoToken{0.227,0.639,0.341}{ mechanical}%
\MapoToken{0.263,0.675,0.369}{,}%
\MapoToken{0.345,0.714,0.408}{ and}%
\MapoToken{0.227,0.639,0.341}{ industrial}%
\MapoToken{0.125,0.533,0.263}{ character}%
\MapoToken{0.180,0.588,0.302}{,}%
\MapoToken{0.184,0.592,0.306}{ which}%
\MapoToken{0.208,0.620,0.325}{ is}%
\MapoToken{0.176,0.584,0.302}{ characteristic}%
\MapoToken{0.282,0.682,0.376}{ of}%
\MapoToken{0.337,0.710,0.404}{ large}%
\MapoToken{0.322,0.702,0.396}{-scale}%
\MapoToken{0.259,0.671,0.365}{ industrial}%
\MapoToken{0.122,0.529,0.259}{ equipment}%
\MapoToken{0.169,0.576,0.294}{,}%
\MapoToken{0.275,0.682,0.376}{ so}%
\MapoToken{0.271,0.678,0.373}{ the}%
\MapoToken{0.251,0.667,0.365}{ correct}%
\MapoToken{0.133,0.541,0.267}{ answer}%
\MapoToken{0.220,0.631,0.337}{ is}%
\MapoToken{0.137,0.545,0.271}{ Factory}%
\MapoToken{0.086,0.502,0.235}{ machinery}%
\MapoToken{0.004,0.431,0.176}{. }%
\MapoToken{0.004,0.431,0.176}{\textless{}/think\textgreater{}}%
\MapoToken{0.102,0.518,0.247}{\phantom{x}}%
\MapoToken{0.090,0.506,0.239}{Factory}%
\MapoToken{0.000,0.376,0.153}{ machinery}%
\end{MapoTokenStream}
\MapoMetricRule

\endgroup

%% file: appendix/exp_2.tex
\begingroup
\definecolor{MapoInk}{RGB}{28,36,48}
\definecolor{MapoMuted}{RGB}{92,102,116}
\definecolor{MapoFrame}{RGB}{202,211,222}
\definecolor{MapoSoft}{RGB}{247,249,252}
\definecolor{MapoMetricTitle}{RGB}{239,243,248}
\definecolor{MapoCorrect}{RGB}{34,119,80}
\definecolor{MapoCorrectBg}{RGB}{238,248,243}
\definecolor{MapoIncorrect}{RGB}{174,60,60}
\definecolor{MapoIncorrectBg}{RGB}{252,241,241}

\newcommand{\MapoToken}[2]{%
  \begingroup\colorbox[rgb]{#1}{\strut #2}\endgroup%
  \hspace{0.05em}\allowbreak%
}

\newcommand{\MapoBox}[2]{%
  \par\smallskip\noindent%
  {\setlength{\fboxsep}{4pt}\setlength{\fboxrule}{0.35pt}%
   \fcolorbox{MapoFrame}{#1}{%
     \parbox{\dimexpr\linewidth-2\fboxsep-2\fboxrule\relax}{#2}%
   }}%
  \par\smallskip%
}

\newcommand{\MapoModelHeader}[5]{%
  \MapoBox{#4}{%
    {\bfseries #1}\hfill{\bfseries\textcolor{#3}{#2}}\par
    {\normalsize\textcolor{MapoMuted}{Prediction:} #5}%
  }%
}

\newcommand{\MapoMetricHeader}[2]{%
  \MapoBox{MapoMetricTitle}{%
    {\normalsize\bfseries #1}\par
    {\normalsize\itshape #2}%
  }%
}

\newenvironment{MapoTokenStream}{%
  \par\noindent\begingroup%
  \normalsize\ttfamily\raggedright\sloppy%
  \setlength{\fboxsep}{0.75pt}%
  \setlength{\parindent}{0pt}%
  \setlength{\emergencystretch}{3em}%
  \hyphenpenalty=10000\exhyphenpenalty=10000%
}{\par\endgroup\smallskip}

\newcommand{\MapoMetricRule}{%
  \par\noindent{\color{MapoFrame}\rule{\linewidth}{0.25pt}}\par\smallskip%
}

% \subsection*{Sample 0013: Horse-drawn wagon}
\MapoBox{MapoSoft}{%
\textbf{Question:} What is the transportation mode referred to in the audio?\\
\textbf{Ground truth:} Horse-drawn wagon%
}

\MapoModelHeader{Baseline}{Incorrect}{MapoIncorrect}{MapoIncorrectBg}{Train}

\begin{MapoTokenStream}
\MapoToken{0.000,0.349,0.141}{\textless{}think\textgreater{}}%
\MapoToken{0.039,0.463,0.200}{\phantom{x}}%
\MapoToken{0.306,0.698,0.392}{Got}%
\MapoToken{0.573,0.824,0.561}{ it}%
\MapoToken{0.498,0.788,0.498}{,}%
\MapoToken{0.235,0.647,0.349}{ let}%
\MapoToken{0.514,0.796,0.510}{'s}%
\MapoToken{0.659,0.863,0.635}{ tackle}%
\MapoToken{0.000,0.267,0.106}{ the}%
\MapoToken{0.675,0.871,0.651}{ problem}%
\MapoToken{0.659,0.863,0.635}{.}%
\MapoToken{0.792,0.918,0.765}{ First}%
\MapoToken{0.082,0.498,0.231}{,}%
\MapoToken{0.247,0.663,0.361}{ the}%
\MapoToken{0.655,0.859,0.627}{ user}%
\MapoToken{0.710,0.882,0.682}{ has}%
\MapoToken{0.459,0.769,0.467}{ provided}%
\MapoToken{0.000,0.267,0.106}{ an}%
\MapoToken{0.157,0.565,0.286}{ audio}%
\MapoToken{0.388,0.737,0.431}{ file}%
\MapoToken{0.816,0.929,0.792}{ and}%
\MapoToken{0.776,0.910,0.749}{ a}%
\MapoToken{0.741,0.898,0.714}{ multiple}%
\MapoToken{0.776,0.910,0.749}{-choice}%
\MapoToken{0.824,0.929,0.800}{ question}%
\MapoToken{0.859,0.945,0.839}{ asking}%
\MapoToken{0.843,0.937,0.820}{ to}%
\MapoToken{0.788,0.918,0.761}{ identify}%
\MapoToken{0.533,0.808,0.529}{ the}%
\MapoToken{0.353,0.718,0.412}{ transportation}%
\MapoToken{0.600,0.835,0.584}{ mode}%
\MapoToken{0.808,0.925,0.784}{ from}%
\MapoToken{0.000,0.267,0.106}{ the}%
\MapoToken{0.827,0.933,0.804}{ given}%
\MapoToken{0.816,0.929,0.792}{ choices}%
\MapoToken{0.796,0.918,0.769}{:}%
\MapoToken{0.396,0.741,0.435}{ Automobile}%
\MapoToken{0.757,0.902,0.729}{,}%
\MapoToken{0.169,0.576,0.294}{ Train}%
\MapoToken{0.243,0.655,0.353}{,}%
\MapoToken{0.282,0.682,0.376}{ Aer}%
\MapoToken{0.000,0.267,0.106}{op}%
\MapoToken{0.592,0.831,0.573}{lane}%
\MapoToken{0.796,0.922,0.773}{,}%
\MapoToken{0.827,0.933,0.804}{ or}%
\MapoToken{0.067,0.482,0.220}{ Horse}%
\MapoToken{0.000,0.384,0.153}{-d}%
\MapoToken{0.557,0.816,0.545}{rawn}%
\MapoToken{0.671,0.867,0.647}{ wagon}%
\MapoToken{0.839,0.937,0.816}{.}%
\MapoToken{0.804,0.925,0.780}{ Then}%
\MapoToken{0.592,0.831,0.573}{,}%
\MapoToken{0.812,0.925,0.788}{ analyzing}%
\MapoToken{0.000,0.267,0.106}{ the}%
\MapoToken{0.192,0.604,0.314}{ audio}%
\MapoToken{0.694,0.878,0.671}{ content}%
\MapoToken{0.114,0.525,0.251}{ reveals}%
\MapoToken{0.000,0.267,0.106}{ a}%
\MapoToken{0.000,0.267,0.106}{ dominant}%
\MapoToken{0.000,0.267,0.106}{ rhyth}%
\MapoToken{0.000,0.267,0.106}{mic}%
\MapoToken{0.000,0.388,0.157}{,}%
\MapoToken{0.000,0.267,0.106}{ metallic}%
\MapoToken{0.000,0.267,0.106}{ cl}%
\MapoToken{0.000,0.267,0.106}{attering}%
\MapoToken{0.192,0.604,0.314}{ sound}%
\MapoToken{0.122,0.529,0.259}{ with}%
\MapoToken{0.090,0.506,0.239}{ a}%
\MapoToken{0.212,0.624,0.329}{ consistent}%
\MapoToken{0.000,0.412,0.165}{ '}%
\MapoToken{0.000,0.408,0.165}{cl}%
\MapoToken{0.012,0.439,0.180}{ack}%
\MapoToken{0.024,0.447,0.188}{-cl}%
\MapoToken{0.145,0.553,0.278}{ack}%
\MapoToken{0.000,0.408,0.165}{...}%
\MapoToken{0.090,0.506,0.239}{ cl}%
\MapoToken{0.302,0.694,0.388}{ack}%
\MapoToken{0.388,0.737,0.431}{-cl}%
\MapoToken{0.227,0.639,0.341}{ack}%
\MapoToken{0.396,0.741,0.435}{...'}%
\MapoToken{0.427,0.753,0.447}{ pattern}%
\MapoToken{0.584,0.827,0.569}{,}%
\MapoToken{0.306,0.698,0.392}{ accompanied}%
\MapoToken{0.102,0.518,0.247}{ by}%
\MapoToken{0.220,0.631,0.337}{ a}%
\MapoToken{0.169,0.576,0.294}{ low}%
\MapoToken{0.369,0.725,0.420}{-frequency}%
\MapoToken{0.325,0.706,0.400}{ rum}%
\MapoToken{0.490,0.784,0.494}{ble}%
\MapoToken{0.533,0.808,0.529}{ and}%
\MapoToken{0.451,0.769,0.463}{ higher}%
\MapoToken{0.776,0.910,0.749}{-p}%
\MapoToken{0.416,0.749,0.443}{itched}%
\MapoToken{0.451,0.769,0.463}{ sque}%
\MapoToken{0.506,0.792,0.506}{aling}%
\MapoToken{0.647,0.859,0.624}{ or}%
\MapoToken{0.592,0.831,0.573}{ grinding}%
\MapoToken{0.498,0.788,0.498}{ noise}%
\MapoToken{0.694,0.878,0.671}{,}%
\MapoToken{0.686,0.875,0.659}{ suggesting}%
\MapoToken{0.592,0.831,0.573}{ wheels}%
\MapoToken{0.690,0.875,0.667}{ moving}%
\MapoToken{0.682,0.871,0.655}{ over}%
\MapoToken{0.682,0.871,0.655}{ a}%
\MapoToken{0.722,0.890,0.698}{ joint}%
\MapoToken{0.737,0.894,0.710}{ed}%
\MapoToken{0.761,0.906,0.733}{ metal}%
\MapoToken{0.733,0.894,0.706}{ track}%
\MapoToken{0.733,0.894,0.706}{.}%
\MapoToken{0.890,0.957,0.871}{ Next}%
\MapoToken{0.843,0.937,0.820}{,}%
\MapoToken{0.894,0.961,0.875}{ evaluating}%
\MapoToken{0.851,0.941,0.827}{ the}%
\MapoToken{0.867,0.949,0.847}{ options}%
\MapoToken{0.835,0.937,0.812}{:}%
\MapoToken{0.706,0.882,0.678}{ Automobile}%
\MapoToken{0.824,0.929,0.800}{ is}%
\MapoToken{0.855,0.941,0.831}{ incorrect}%
\MapoToken{0.765,0.906,0.737}{ because}%
\MapoToken{0.690,0.875,0.667}{ cars}%
\MapoToken{0.733,0.894,0.706}{ produce}%
\MapoToken{0.694,0.878,0.671}{ continuous}%
\MapoToken{0.722,0.890,0.698}{ tire}%
\MapoToken{0.769,0.910,0.741}{ noise}%
\MapoToken{0.816,0.929,0.792}{ without}%
\MapoToken{0.800,0.922,0.776}{ metallic}%
\MapoToken{0.710,0.882,0.682}{ cl}%
\MapoToken{0.796,0.922,0.773}{attering}%
\MapoToken{0.835,0.937,0.812}{;}%
\MapoToken{0.796,0.922,0.773}{ Train}%
\MapoToken{0.855,0.941,0.831}{ matches}%
\MapoToken{0.867,0.949,0.847}{ perfectly}%
\MapoToken{0.800,0.922,0.776}{ with}%
\MapoToken{0.816,0.929,0.792}{ the}%
\MapoToken{0.729,0.890,0.702}{ '}%
\MapoToken{0.682,0.871,0.655}{cl}%
\MapoToken{0.729,0.890,0.702}{ack}%
\MapoToken{0.600,0.835,0.584}{-cl}%
\MapoToken{0.745,0.898,0.722}{ack}%
\MapoToken{0.784,0.914,0.757}{'}%
\MapoToken{0.796,0.922,0.773}{ of}%
\MapoToken{0.804,0.925,0.780}{ steel}%
\MapoToken{0.796,0.922,0.773}{ wheels}%
\MapoToken{0.835,0.937,0.812}{ on}%
\MapoToken{0.804,0.925,0.780}{ rail}%
\MapoToken{0.812,0.925,0.788}{ joints}%
\MapoToken{0.827,0.933,0.804}{ and}%
\MapoToken{0.812,0.925,0.788}{ characteristic}%
\MapoToken{0.816,0.929,0.792}{ rum}%
\MapoToken{0.843,0.937,0.820}{ble}%
\MapoToken{0.835,0.937,0.812}{/s}%
\MapoToken{0.820,0.929,0.796}{que}%
\MapoToken{0.812,0.925,0.788}{al}%
\MapoToken{0.851,0.941,0.827}{;}%
\MapoToken{0.824,0.929,0.800}{ Aer}%
\MapoToken{0.765,0.906,0.737}{op}%
\MapoToken{0.827,0.933,0.804}{lane}%
\MapoToken{0.890,0.957,0.871}{ is}%
\MapoToken{0.902,0.961,0.882}{ ruled}%
\MapoToken{0.894,0.961,0.875}{ out}%
\MapoToken{0.855,0.941,0.831}{ by}%
\MapoToken{0.847,0.941,0.824}{ its}%
\MapoToken{0.808,0.925,0.784}{ loud}%
\MapoToken{0.820,0.929,0.796}{ engine}%
\MapoToken{0.831,0.933,0.808}{ roar}%
\MapoToken{0.859,0.945,0.835}{ and}%
\MapoToken{0.839,0.937,0.816}{ air}%
\MapoToken{0.824,0.929,0.800}{ who}%
\MapoToken{0.824,0.929,0.800}{osh}%
\MapoToken{0.847,0.941,0.824}{ing}%
\MapoToken{0.863,0.949,0.843}{ sounds}%
\MapoToken{0.882,0.953,0.863}{;}%
\MapoToken{0.675,0.871,0.651}{ Horse}%
\MapoToken{0.831,0.933,0.808}{-d}%
\MapoToken{0.780,0.914,0.753}{rawn}%
\MapoToken{0.800,0.922,0.776}{ wagon}%
\MapoToken{0.871,0.949,0.851}{ is}%
\MapoToken{0.898,0.961,0.882}{ inconsistent}%
\MapoToken{0.871,0.949,0.851}{ due}%
\MapoToken{0.769,0.910,0.741}{ to}%
\MapoToken{0.753,0.902,0.725}{ wooden}%
\MapoToken{0.808,0.925,0.784}{/}%
\MapoToken{0.757,0.902,0.729}{iron}%
\MapoToken{0.800,0.922,0.776}{ wheel}%
\MapoToken{0.737,0.894,0.710}{ cl}%
\MapoToken{0.800,0.922,0.776}{atter}%
\MapoToken{0.820,0.929,0.796}{ on}%
\MapoToken{0.851,0.941,0.827}{ roads}%
\MapoToken{0.859,0.945,0.839}{ lacking}%
\MapoToken{0.871,0.949,0.851}{ metallic}%
\MapoToken{0.859,0.945,0.839}{ rail}%
\MapoToken{0.867,0.949,0.847}{ characteristics}%
\MapoToken{0.820,0.929,0.796}{ and}%
\MapoToken{0.855,0.941,0.831}{ absence}%
\MapoToken{0.788,0.918,0.761}{ of}%
\MapoToken{0.694,0.878,0.671}{ horse}%
\MapoToken{0.659,0.863,0.635}{ hoof}%
\MapoToken{0.733,0.894,0.706}{ sounds}%
\MapoToken{0.859,0.945,0.839}{.}%
\MapoToken{0.890,0.957,0.871}{ Finally}%
\MapoToken{0.859,0.945,0.839}{,}%
\MapoToken{0.808,0.925,0.784}{ the}%
\MapoToken{0.765,0.906,0.737}{ audio}%
\MapoToken{0.882,0.953,0.863}{ evidence}%
\MapoToken{0.898,0.961,0.882}{ strongly}%
\MapoToken{0.906,0.965,0.890}{ points}%
\MapoToken{0.835,0.937,0.812}{ to}%
\MapoToken{0.745,0.898,0.722}{ a}%
\MapoToken{0.776,0.910,0.749}{ train}%
\MapoToken{0.878,0.953,0.859}{ as}%
\MapoToken{0.882,0.953,0.863}{ the}%
\MapoToken{0.855,0.941,0.831}{ transportation}%
\MapoToken{0.890,0.957,0.871}{ mode}%
\MapoToken{0.890,0.957,0.871}{,}%
\MapoToken{0.831,0.933,0.808}{ with}%
\MapoToken{0.808,0.925,0.784}{ the}%
\MapoToken{0.753,0.902,0.725}{ rhyth}%
\MapoToken{0.784,0.914,0.757}{mic}%
\MapoToken{0.765,0.906,0.737}{ cl}%
\MapoToken{0.816,0.929,0.792}{atter}%
\MapoToken{0.886,0.957,0.867}{ being}%
\MapoToken{0.882,0.953,0.863}{ the}%
\MapoToken{0.886,0.957,0.867}{ key}%
\MapoToken{0.898,0.961,0.882}{ identifier}%
\MapoToken{0.690,0.875,0.667}{. }%
\MapoToken{0.663,0.863,0.639}{\textless{}/think\textgreater{}}%
\MapoToken{0.600,0.835,0.584}{\phantom{x}}%
\MapoToken{0.722,0.890,0.698}{Train}%
\end{MapoTokenStream}
\MapoMetricRule

\MapoModelHeader{MAPO-No-Attn-Loss}{Incorrect}{MapoIncorrect}{MapoIncorrectBg}{Train}

\begin{MapoTokenStream}
\MapoToken{0.000,0.267,0.106}{\textless{}think\textgreater{}}%
\MapoToken{0.000,0.349,0.141}{\phantom{x}}%
\MapoToken{0.200,0.612,0.322}{Got}%
\MapoToken{0.498,0.788,0.498}{ it}%
\MapoToken{0.365,0.725,0.420}{,}%
\MapoToken{0.114,0.525,0.251}{ let}%
\MapoToken{0.416,0.749,0.443}{'s}%
\MapoToken{0.565,0.820,0.553}{ tackle}%
\MapoToken{0.000,0.267,0.106}{ the}%
\MapoToken{0.592,0.831,0.573}{ problem}%
\MapoToken{0.545,0.812,0.537}{.}%
\MapoToken{0.745,0.898,0.722}{ First}%
\MapoToken{0.000,0.325,0.129}{,}%
\MapoToken{0.090,0.506,0.239}{ the}%
\MapoToken{0.541,0.808,0.533}{ user}%
\MapoToken{0.757,0.902,0.729}{ is}%
\MapoToken{0.765,0.906,0.737}{ asking}%
\MapoToken{0.722,0.890,0.698}{ to}%
\MapoToken{0.694,0.878,0.671}{ identify}%
\MapoToken{0.251,0.667,0.365}{ the}%
\MapoToken{0.145,0.553,0.278}{ transportation}%
\MapoToken{0.333,0.710,0.404}{ mode}%
\MapoToken{0.639,0.855,0.616}{ from}%
\MapoToken{0.000,0.267,0.106}{ the}%
\MapoToken{0.600,0.835,0.584}{ given}%
\MapoToken{0.718,0.886,0.694}{ choices}%
\MapoToken{0.706,0.882,0.678}{:}%
\MapoToken{0.216,0.627,0.333}{ Automobile}%
\MapoToken{0.675,0.871,0.651}{,}%
\MapoToken{0.000,0.408,0.165}{ Train}%
\MapoToken{0.071,0.486,0.224}{,}%
\MapoToken{0.145,0.553,0.278}{ Aer}%
\MapoToken{0.000,0.267,0.106}{op}%
\MapoToken{0.475,0.776,0.478}{lane}%
\MapoToken{0.737,0.894,0.710}{,}%
\MapoToken{0.000,0.267,0.106}{ Horse}%
\MapoToken{0.000,0.267,0.106}{-d}%
\MapoToken{0.400,0.741,0.435}{rawn}%
\MapoToken{0.400,0.741,0.435}{ wagon}%
\MapoToken{0.706,0.882,0.678}{.}%
\MapoToken{0.729,0.890,0.702}{ Then}%
\MapoToken{0.396,0.741,0.435}{,}%
\MapoToken{0.000,0.267,0.106}{ the}%
\MapoToken{0.000,0.318,0.125}{ audio}%
\MapoToken{0.000,0.267,0.106}{ features}%
\MapoToken{0.000,0.267,0.106}{ a}%
\MapoToken{0.000,0.267,0.106}{ prominent}%
\MapoToken{0.000,0.267,0.106}{ rhyth}%
\MapoToken{0.000,0.267,0.106}{mic}%
\MapoToken{0.000,0.267,0.106}{,}%
\MapoToken{0.000,0.267,0.106}{ cl}%
\MapoToken{0.000,0.267,0.106}{attering}%
\MapoToken{0.000,0.361,0.145}{,}%
\MapoToken{0.039,0.463,0.200}{ and}%
\MapoToken{0.000,0.396,0.161}{ metallic}%
\MapoToken{0.082,0.498,0.231}{ sound}%
\MapoToken{0.000,0.412,0.165}{ with}%
\MapoToken{0.000,0.361,0.145}{ a}%
\MapoToken{0.000,0.322,0.129}{ repeating}%
\MapoToken{0.000,0.267,0.106}{ '}%
\MapoToken{0.000,0.333,0.133}{clip}%
\MapoToken{0.169,0.576,0.294}{-c}%
\MapoToken{0.000,0.318,0.125}{lop}%
\MapoToken{0.141,0.549,0.275}{'}%
\MapoToken{0.475,0.776,0.478}{ or}%
\MapoToken{0.227,0.639,0.341}{ '}%
\MapoToken{0.220,0.631,0.337}{cl}%
\MapoToken{0.204,0.616,0.325}{ack}%
\MapoToken{0.231,0.643,0.345}{-cl}%
\MapoToken{0.216,0.627,0.333}{ack}%
\MapoToken{0.475,0.776,0.478}{'}%
\MapoToken{0.235,0.647,0.349}{ pattern}%
\MapoToken{0.502,0.792,0.502}{,}%
\MapoToken{0.718,0.886,0.694}{ which}%
\MapoToken{0.612,0.843,0.592}{ is}%
\MapoToken{0.741,0.898,0.714}{ characteristic}%
\MapoToken{0.459,0.769,0.467}{ of}%
\MapoToken{0.431,0.757,0.451}{ metal}%
\MapoToken{0.388,0.737,0.431}{ wheels}%
\MapoToken{0.475,0.776,0.478}{ moving}%
\MapoToken{0.525,0.800,0.522}{ over}%
\MapoToken{0.541,0.808,0.533}{ joints}%
\MapoToken{0.639,0.855,0.616}{ in}%
\MapoToken{0.686,0.875,0.659}{ a}%
\MapoToken{0.671,0.867,0.647}{ metal}%
\MapoToken{0.592,0.831,0.573}{ track}%
\MapoToken{0.584,0.827,0.569}{,}%
\MapoToken{0.420,0.753,0.447}{ along}%
\MapoToken{0.071,0.486,0.224}{ with}%
\MapoToken{0.000,0.424,0.173}{ a}%
\MapoToken{0.212,0.624,0.329}{ general}%
\MapoToken{0.247,0.659,0.357}{ low}%
\MapoToken{0.263,0.675,0.369}{-frequency}%
\MapoToken{0.235,0.647,0.349}{ rum}%
\MapoToken{0.486,0.784,0.486}{ble}%
\MapoToken{0.290,0.686,0.380}{ and}%
\MapoToken{0.420,0.753,0.447}{ higher}%
\MapoToken{0.647,0.859,0.624}{-p}%
\MapoToken{0.400,0.741,0.435}{itched}%
\MapoToken{0.420,0.753,0.447}{ sc}%
\MapoToken{0.502,0.792,0.502}{ree}%
\MapoToken{0.514,0.796,0.510}{ching}%
\MapoToken{0.592,0.831,0.573}{ sounds}%
\MapoToken{0.820,0.929,0.796}{ typical}%
\MapoToken{0.737,0.894,0.710}{ of}%
\MapoToken{0.686,0.875,0.659}{ trains}%
\MapoToken{0.620,0.843,0.596}{,}%
\MapoToken{0.553,0.816,0.541}{ especially}%
\MapoToken{0.447,0.765,0.459}{ when}%
\MapoToken{0.463,0.773,0.471}{ going}%
\MapoToken{0.627,0.851,0.608}{ around}%
\MapoToken{0.627,0.851,0.608}{ curves}%
\MapoToken{0.757,0.902,0.729}{ or}%
\MapoToken{0.729,0.890,0.702}{ over}%
\MapoToken{0.592,0.831,0.573}{ switches}%
\MapoToken{0.714,0.886,0.686}{.}%
\MapoToken{0.855,0.941,0.831}{ Next}%
\MapoToken{0.796,0.918,0.769}{,}%
\MapoToken{0.851,0.941,0.827}{ evaluating}%
\MapoToken{0.796,0.918,0.769}{ the}%
\MapoToken{0.831,0.933,0.808}{ choices}%
\MapoToken{0.796,0.922,0.773}{:}%
\MapoToken{0.663,0.863,0.639}{ Automobile}%
\MapoToken{0.780,0.914,0.753}{ is}%
\MapoToken{0.820,0.929,0.796}{ incorrect}%
\MapoToken{0.729,0.890,0.702}{ because}%
\MapoToken{0.655,0.859,0.627}{ cars}%
\MapoToken{0.722,0.890,0.698}{ have}%
\MapoToken{0.741,0.898,0.714}{ rubber}%
\MapoToken{0.741,0.898,0.714}{ tires}%
\MapoToken{0.788,0.918,0.761}{ on}%
\MapoToken{0.745,0.898,0.722}{ asphalt}%
\MapoToken{0.769,0.910,0.741}{,}%
\MapoToken{0.757,0.902,0.729}{ producing}%
\MapoToken{0.733,0.894,0.706}{ a}%
\MapoToken{0.788,0.918,0.761}{ much}%
\MapoToken{0.757,0.902,0.729}{ smoother}%
\MapoToken{0.804,0.925,0.780}{,}%
\MapoToken{0.780,0.914,0.753}{ quieter}%
\MapoToken{0.780,0.914,0.753}{,}%
\MapoToken{0.867,0.949,0.847}{ and}%
\MapoToken{0.784,0.914,0.757}{ wh}%
\MapoToken{0.784,0.914,0.757}{ir}%
\MapoToken{0.776,0.910,0.749}{ring}%
\MapoToken{0.776,0.910,0.749}{ sound}%
\MapoToken{0.796,0.922,0.773}{ without}%
\MapoToken{0.788,0.918,0.761}{ the}%
\MapoToken{0.784,0.914,0.757}{ distinct}%
\MapoToken{0.761,0.906,0.733}{ metallic}%
\MapoToken{0.675,0.871,0.651}{ cl}%
\MapoToken{0.733,0.894,0.706}{atter}%
\MapoToken{0.784,0.914,0.757}{;}%
\MapoToken{0.776,0.910,0.749}{ Train}%
\MapoToken{0.839,0.937,0.816}{ is}%
\MapoToken{0.835,0.937,0.812}{ a}%
\MapoToken{0.820,0.929,0.796}{ very}%
\MapoToken{0.859,0.945,0.839}{ strong}%
\MapoToken{0.851,0.941,0.827}{ match}%
\MapoToken{0.788,0.918,0.761}{ because}%
\MapoToken{0.706,0.882,0.678}{ the}%
\MapoToken{0.671,0.867,0.647}{ sound}%
\MapoToken{0.694,0.878,0.671}{ of}%
\MapoToken{0.706,0.882,0.678}{ wheels}%
\MapoToken{0.765,0.906,0.737}{ on}%
\MapoToken{0.769,0.910,0.741}{ rails}%
\MapoToken{0.808,0.925,0.784}{,}%
\MapoToken{0.745,0.898,0.722}{ the}%
\MapoToken{0.698,0.878,0.675}{ track}%
\MapoToken{0.753,0.902,0.725}{ joints}%
\MapoToken{0.757,0.902,0.729}{,}%
\MapoToken{0.827,0.933,0.804}{ and}%
\MapoToken{0.733,0.894,0.706}{ the}%
\MapoToken{0.757,0.902,0.729}{ overall}%
\MapoToken{0.769,0.910,0.741}{ rum}%
\MapoToken{0.780,0.914,0.753}{bling}%
\MapoToken{0.843,0.937,0.820}{ is}%
\MapoToken{0.784,0.914,0.757}{ classic}%
\MapoToken{0.624,0.847,0.600}{ train}%
\MapoToken{0.627,0.851,0.608}{ noise}%
\MapoToken{0.812,0.925,0.788}{;}%
\MapoToken{0.780,0.914,0.753}{ Aer}%
\MapoToken{0.714,0.886,0.686}{op}%
\MapoToken{0.796,0.922,0.773}{lane}%
\MapoToken{0.859,0.945,0.839}{ is}%
\MapoToken{0.878,0.953,0.859}{ incorrect}%
\MapoToken{0.855,0.941,0.831}{ as}%
\MapoToken{0.839,0.937,0.816}{ it}%
\MapoToken{0.824,0.929,0.800}{ would}%
\MapoToken{0.816,0.929,0.792}{ involve}%
\MapoToken{0.824,0.929,0.800}{ the}%
\MapoToken{0.800,0.922,0.776}{ loud}%
\MapoToken{0.831,0.933,0.808}{,}%
\MapoToken{0.812,0.925,0.788}{ continuous}%
\MapoToken{0.824,0.929,0.800}{ roar}%
\MapoToken{0.863,0.949,0.843}{ of}%
\MapoToken{0.843,0.937,0.820}{ jet}%
\MapoToken{0.824,0.929,0.800}{ engines}%
\MapoToken{0.882,0.953,0.863}{ or}%
\MapoToken{0.875,0.953,0.855}{ the}%
\MapoToken{0.863,0.949,0.843}{ buzzing}%
\MapoToken{0.855,0.941,0.831}{ prop}%
\MapoToken{0.851,0.941,0.827}{eller}%
\MapoToken{0.859,0.945,0.839}{ sound}%
\MapoToken{0.859,0.945,0.839}{,}%
\MapoToken{0.871,0.949,0.851}{ neither}%
\MapoToken{0.529,0.804,0.525}{ of}%
\MapoToken{0.827,0.933,0.804}{ which}%
\MapoToken{0.792,0.918,0.765}{ is}%
\MapoToken{0.800,0.922,0.776}{ present}%
\MapoToken{0.855,0.941,0.831}{;}%
\MapoToken{0.867,0.949,0.847}{ and}%
\MapoToken{0.624,0.847,0.600}{ Horse}%
\MapoToken{0.714,0.886,0.686}{-d}%
\MapoToken{0.722,0.890,0.698}{rawn}%
\MapoToken{0.757,0.902,0.729}{ wagon}%
\MapoToken{0.804,0.925,0.780}{ is}%
\MapoToken{0.867,0.949,0.847}{ incorrect}%
\MapoToken{0.800,0.922,0.776}{ because}%
\MapoToken{0.800,0.922,0.776}{ it}%
\MapoToken{0.796,0.918,0.769}{ would}%
\MapoToken{0.722,0.890,0.698}{ produce}%
\MapoToken{0.714,0.886,0.686}{ the}%
\MapoToken{0.780,0.914,0.753}{ sound}%
\MapoToken{0.722,0.890,0.698}{ of}%
\MapoToken{0.596,0.835,0.580}{ horse}%
\MapoToken{0.753,0.902,0.725}{ ho}%
\MapoToken{0.690,0.875,0.667}{oves}%
\MapoToken{0.780,0.914,0.753}{ on}%
\MapoToken{0.812,0.925,0.788}{ a}%
\MapoToken{0.820,0.929,0.796}{ road}%
\MapoToken{0.757,0.902,0.729}{ (}%
\MapoToken{0.753,0.902,0.725}{a}%
\MapoToken{0.847,0.941,0.824}{ different}%
\MapoToken{0.859,0.945,0.835}{ kind}%
\MapoToken{0.796,0.922,0.773}{ of}%
\MapoToken{0.659,0.863,0.635}{ '}%
\MapoToken{0.706,0.882,0.678}{clip}%
\MapoToken{0.600,0.835,0.584}{-c}%
\MapoToken{0.792,0.918,0.765}{lop}%
\MapoToken{0.831,0.933,0.808}{'}%
\MapoToken{0.863,0.949,0.843}{ that}%
\MapoToken{0.855,0.941,0.831}{ is}%
\MapoToken{0.851,0.941,0.827}{ less}%
\MapoToken{0.847,0.941,0.824}{ metallic}%
\MapoToken{0.843,0.937,0.820}{ and}%
\MapoToken{0.851,0.941,0.827}{ more}%
\MapoToken{0.839,0.937,0.816}{ organic}%
\MapoToken{0.851,0.941,0.827}{)}%
\MapoToken{0.843,0.937,0.820}{ and}%
\MapoToken{0.808,0.925,0.784}{ the}%
\MapoToken{0.690,0.875,0.667}{ c}%
\MapoToken{0.769,0.910,0.741}{reak}%
\MapoToken{0.824,0.929,0.800}{ing}%
\MapoToken{0.851,0.941,0.827}{ of}%
\MapoToken{0.831,0.933,0.808}{ wooden}%
\MapoToken{0.851,0.941,0.827}{ wheels}%
\MapoToken{0.859,0.945,0.839}{ and}%
\MapoToken{0.843,0.937,0.820}{ wagon}%
\MapoToken{0.804,0.925,0.780}{ body}%
\MapoToken{0.839,0.937,0.816}{,}%
\MapoToken{0.816,0.929,0.792}{ but}%
\MapoToken{0.686,0.875,0.659}{ the}%
\MapoToken{0.729,0.890,0.702}{ audio}%
\MapoToken{0.808,0.925,0.784}{ is}%
\MapoToken{0.808,0.925,0.784}{ clearly}%
\MapoToken{0.792,0.918,0.765}{ metallic}%
\MapoToken{0.800,0.922,0.776}{ and}%
\MapoToken{0.784,0.914,0.757}{ mechanical}%
\MapoToken{0.816,0.929,0.792}{.}%
\MapoToken{0.859,0.945,0.835}{ Finally}%
\MapoToken{0.816,0.929,0.792}{,}%
\MapoToken{0.745,0.898,0.722}{ the}%
\MapoToken{0.682,0.871,0.655}{ audio}%
\MapoToken{0.827,0.933,0.804}{ evidence}%
\MapoToken{0.871,0.949,0.851}{ strongly}%
\MapoToken{0.890,0.957,0.871}{ points}%
\MapoToken{0.796,0.922,0.773}{ to}%
\MapoToken{0.694,0.878,0.671}{ a}%
\MapoToken{0.710,0.882,0.682}{ train}%
\MapoToken{0.843,0.937,0.820}{,}%
\MapoToken{0.847,0.941,0.824}{ as}%
\MapoToken{0.761,0.906,0.733}{ the}%
\MapoToken{0.796,0.918,0.769}{ specific}%
\MapoToken{0.769,0.910,0.741}{ sound}%
\MapoToken{0.776,0.910,0.749}{ of}%
\MapoToken{0.776,0.910,0.749}{ wheels}%
\MapoToken{0.827,0.933,0.804}{ on}%
\MapoToken{0.816,0.929,0.792}{ a}%
\MapoToken{0.835,0.937,0.812}{ track}%
\MapoToken{0.867,0.949,0.847}{ is}%
\MapoToken{0.878,0.953,0.859}{ unmistak}%
\MapoToken{0.859,0.945,0.839}{able}%
\MapoToken{0.871,0.949,0.851}{,}%
\MapoToken{0.875,0.953,0.855}{ so}%
\MapoToken{0.863,0.949,0.843}{ the}%
\MapoToken{0.859,0.945,0.835}{ answer}%
\MapoToken{0.843,0.937,0.820}{ is}%
\MapoToken{0.808,0.925,0.784}{ Train}%
\MapoToken{0.596,0.835,0.580}{. }%
\MapoToken{0.608,0.839,0.588}{\textless{}/think\textgreater{}}%
\MapoToken{0.506,0.792,0.506}{\phantom{x}}%
\MapoToken{0.671,0.867,0.647}{Train}%
\end{MapoTokenStream}
\MapoMetricRule

\MapoModelHeader{Target-MAPO+Attn-Loss}{Correct}{MapoCorrect}{MapoCorrectBg}{Horse-drawn wagon}

\begin{MapoTokenStream}
\MapoToken{0.051,0.471,0.212}{\textless{}think\textgreater{}}%
\MapoToken{0.090,0.506,0.239}{\phantom{x}}%
\MapoToken{0.118,0.529,0.255}{Got}%
\MapoToken{0.196,0.608,0.318}{ it}%
\MapoToken{0.169,0.576,0.294}{,}%
\MapoToken{0.000,0.267,0.106}{ let}%
\MapoToken{0.212,0.624,0.329}{'s}%
\MapoToken{0.153,0.561,0.282}{ tackle}%
\MapoToken{0.000,0.267,0.106}{ the}%
\MapoToken{0.133,0.541,0.267}{ problem}%
\MapoToken{0.169,0.576,0.294}{.}%
\MapoToken{0.161,0.569,0.290}{ First}%
\MapoToken{0.000,0.314,0.125}{,}%
\MapoToken{0.067,0.482,0.220}{ the}%
\MapoToken{0.145,0.553,0.278}{ user}%
\MapoToken{0.235,0.647,0.349}{ is}%
\MapoToken{0.243,0.655,0.353}{ asking}%
\MapoToken{0.271,0.678,0.373}{ to}%
\MapoToken{0.204,0.616,0.325}{ identify}%
\MapoToken{0.180,0.588,0.302}{ the}%
\MapoToken{0.075,0.490,0.224}{ transportation}%
\MapoToken{0.114,0.525,0.251}{ mode}%
\MapoToken{0.184,0.592,0.306}{ from}%
\MapoToken{0.000,0.267,0.106}{ the}%
\MapoToken{0.227,0.639,0.341}{ given}%
\MapoToken{0.184,0.592,0.306}{ choices}%
\MapoToken{0.247,0.659,0.357}{:}%
\MapoToken{0.000,0.384,0.153}{ Automobile}%
\MapoToken{0.271,0.678,0.373}{,}%
\MapoToken{0.000,0.376,0.153}{ Train}%
\MapoToken{0.039,0.463,0.200}{,}%
\MapoToken{0.082,0.498,0.231}{ Aer}%
\MapoToken{0.000,0.267,0.106}{op}%
\MapoToken{0.039,0.463,0.200}{lane}%
\MapoToken{0.247,0.659,0.357}{,}%
\MapoToken{0.294,0.690,0.384}{ or}%
\MapoToken{0.000,0.361,0.145}{ Horse}%
\MapoToken{0.000,0.267,0.106}{-d}%
\MapoToken{0.161,0.569,0.290}{rawn}%
\MapoToken{0.094,0.510,0.239}{ wagon}%
\MapoToken{0.184,0.592,0.306}{,}%
\MapoToken{0.216,0.627,0.333}{ based}%
\MapoToken{0.145,0.553,0.278}{ on}%
\MapoToken{0.000,0.282,0.114}{ the}%
\MapoToken{0.094,0.510,0.239}{ provided}%
\MapoToken{0.000,0.408,0.165}{ audio}%
\MapoToken{0.000,0.369,0.149}{ clip}%
\MapoToken{0.137,0.545,0.271}{.}%
\MapoToken{0.145,0.553,0.278}{ Then}%
\MapoToken{0.024,0.447,0.188}{,}%
\MapoToken{0.000,0.314,0.125}{ the}%
\MapoToken{0.000,0.267,0.106}{ audio}%
\MapoToken{0.000,0.267,0.106}{ contains}%
\MapoToken{0.000,0.267,0.106}{ a}%
\MapoToken{0.000,0.267,0.106}{ distinct}%
\MapoToken{0.000,0.267,0.106}{,}%
\MapoToken{0.000,0.267,0.106}{ rhyth}%
\MapoToken{0.000,0.267,0.106}{mic}%
\MapoToken{0.000,0.267,0.106}{,}%
\MapoToken{0.000,0.267,0.106}{ cl}%
\MapoToken{0.000,0.267,0.106}{attering}%
\MapoToken{0.000,0.267,0.106}{ sound}%
\MapoToken{0.000,0.286,0.114}{ that}%
\MapoToken{0.000,0.318,0.125}{ is}%
\MapoToken{0.000,0.337,0.133}{ a}%
\MapoToken{0.000,0.267,0.106}{ combination}%
\MapoToken{0.000,0.267,0.106}{ of}%
\MapoToken{0.000,0.333,0.133}{ mechanical}%
\MapoToken{0.000,0.318,0.125}{ noise}%
\MapoToken{0.000,0.361,0.145}{ and}%
\MapoToken{0.000,0.322,0.129}{ a}%
\MapoToken{0.051,0.471,0.212}{ specific}%
\MapoToken{0.000,0.325,0.129}{ pattern}%
\MapoToken{0.000,0.267,0.106}{:}%
\MapoToken{0.000,0.267,0.106}{ a}%
\MapoToken{0.000,0.388,0.157}{ '}%
\MapoToken{0.000,0.404,0.161}{clip}%
\MapoToken{0.200,0.612,0.322}{-c}%
\MapoToken{0.000,0.271,0.110}{lop}%
\MapoToken{0.000,0.302,0.122}{'}%
\MapoToken{0.071,0.486,0.224}{ or}%
\MapoToken{0.173,0.580,0.298}{ '}%
\MapoToken{0.145,0.553,0.278}{clo}%
\MapoToken{0.000,0.325,0.129}{pping}%
\MapoToken{0.000,0.396,0.161}{'}%
\MapoToken{0.000,0.373,0.149}{ sound}%
\MapoToken{0.055,0.475,0.212}{ characteristic}%
\MapoToken{0.051,0.471,0.212}{ of}%
\MapoToken{0.078,0.494,0.227}{ ho}%
\MapoToken{0.012,0.439,0.180}{oves}%
\MapoToken{0.047,0.467,0.208}{ hitting}%
\MapoToken{0.082,0.498,0.231}{ a}%
\MapoToken{0.122,0.529,0.259}{ hard}%
\MapoToken{0.039,0.463,0.200}{ surface}%
\MapoToken{0.153,0.561,0.282}{ like}%
\MapoToken{0.039,0.463,0.200}{ pavement}%
\MapoToken{0.192,0.604,0.314}{ or}%
\MapoToken{0.173,0.580,0.298}{ wooden}%
\MapoToken{0.004,0.431,0.176}{ tracks}%
\MapoToken{0.000,0.369,0.149}{,}%
\MapoToken{0.012,0.439,0.180}{ overl}%
\MapoToken{0.000,0.424,0.173}{aid}%
\MapoToken{0.000,0.412,0.165}{ with}%
\MapoToken{0.067,0.482,0.220}{ the}%
\MapoToken{0.000,0.427,0.173}{ sound}%
\MapoToken{0.012,0.439,0.180}{ of}%
\MapoToken{0.043,0.467,0.204}{ wheels}%
\MapoToken{0.027,0.451,0.192}{ rolling}%
\MapoToken{0.102,0.518,0.247}{ and}%
\MapoToken{0.145,0.553,0.278}{ possibly}%
\MapoToken{0.173,0.580,0.298}{ the}%
\MapoToken{0.149,0.557,0.278}{ c}%
\MapoToken{0.086,0.502,0.235}{reak}%
\MapoToken{0.114,0.525,0.251}{ing}%
\MapoToken{0.212,0.624,0.329}{ of}%
\MapoToken{0.247,0.663,0.361}{ a}%
\MapoToken{0.129,0.537,0.267}{ harness}%
\MapoToken{0.271,0.678,0.373}{ or}%
\MapoToken{0.165,0.573,0.290}{ carriage}%
\MapoToken{0.000,0.412,0.165}{,}%
\MapoToken{0.000,0.267,0.106}{ with}%
\MapoToken{0.000,0.361,0.145}{ the}%
\MapoToken{0.000,0.267,0.106}{ sound}%
\MapoToken{0.000,0.412,0.165}{ being}%
\MapoToken{0.000,0.376,0.153}{ consistent}%
\MapoToken{0.000,0.427,0.173}{ and}%
\MapoToken{0.000,0.290,0.118}{ continuous}%
\MapoToken{0.082,0.498,0.231}{ indicating}%
\MapoToken{0.067,0.482,0.220}{ continuous}%
\MapoToken{0.000,0.408,0.165}{ motion}%
\MapoToken{0.055,0.475,0.212}{,}%
\MapoToken{0.000,0.408,0.165}{ and}%
\MapoToken{0.114,0.525,0.251}{ the}%
\MapoToken{0.043,0.467,0.204}{ rhythm}%
\MapoToken{0.078,0.494,0.227}{ being}%
\MapoToken{0.102,0.518,0.247}{ very}%
\MapoToken{0.031,0.455,0.196}{ regular}%
\MapoToken{0.141,0.549,0.275}{ suggesting}%
\MapoToken{0.196,0.608,0.318}{ a}%
\MapoToken{0.129,0.537,0.267}{ steady}%
\MapoToken{0.075,0.490,0.224}{ pace}%
\MapoToken{0.137,0.545,0.271}{.}%
\MapoToken{0.173,0.580,0.298}{ Next}%
\MapoToken{0.231,0.643,0.345}{,}%
\MapoToken{0.204,0.616,0.325}{ evaluating}%
\MapoToken{0.282,0.682,0.376}{ the}%
\MapoToken{0.173,0.580,0.298}{ choices}%
\MapoToken{0.220,0.631,0.337}{:}%
\MapoToken{0.247,0.663,0.361}{ for}%
\MapoToken{0.090,0.506,0.239}{ Automobile}%
\MapoToken{0.161,0.569,0.290}{,}%
\MapoToken{0.157,0.565,0.286}{ it}%
\MapoToken{0.176,0.584,0.302}{ would}%
\MapoToken{0.196,0.608,0.318}{ produce}%
\MapoToken{0.165,0.573,0.290}{ engine}%
\MapoToken{0.141,0.549,0.275}{ noise}%
\MapoToken{0.247,0.663,0.361}{,}%
\MapoToken{0.220,0.631,0.337}{ tire}%
\MapoToken{0.176,0.584,0.302}{ noise}%
\MapoToken{0.243,0.655,0.353}{ on}%
\MapoToken{0.184,0.592,0.306}{ asphalt}%
\MapoToken{0.224,0.635,0.341}{ (}%
\MapoToken{0.247,0.663,0.361}{a}%
\MapoToken{0.231,0.643,0.345}{ more}%
\MapoToken{0.224,0.635,0.341}{ continuous}%
\MapoToken{0.322,0.702,0.396}{,}%
\MapoToken{0.235,0.647,0.349}{ lower}%
\MapoToken{0.169,0.576,0.294}{-p}%
\MapoToken{0.271,0.678,0.373}{itched}%
\MapoToken{0.204,0.616,0.325}{ hum}%
\MapoToken{0.216,0.627,0.333}{),}%
\MapoToken{0.282,0.682,0.376}{ and}%
\MapoToken{0.337,0.710,0.404}{ possibly}%
\MapoToken{0.314,0.698,0.392}{ the}%
\MapoToken{0.157,0.565,0.286}{ sound}%
\MapoToken{0.337,0.710,0.404}{ of}%
\MapoToken{0.439,0.761,0.455}{ a}%
\MapoToken{0.506,0.792,0.506}{ carb}%
\MapoToken{0.463,0.773,0.471}{ure}%
\MapoToken{0.271,0.678,0.373}{tor}%
\MapoToken{0.525,0.800,0.522}{ or}%
\MapoToken{0.282,0.682,0.376}{ exhaust}%
\MapoToken{0.173,0.580,0.298}{,}%
\MapoToken{0.161,0.569,0.290}{ but}%
\MapoToken{0.196,0.608,0.318}{ the}%
\MapoToken{0.149,0.557,0.278}{ audio}%
\MapoToken{0.243,0.655,0.353}{ lacks}%
\MapoToken{0.188,0.600,0.314}{ engine}%
\MapoToken{0.176,0.584,0.302}{ noise}%
\MapoToken{0.235,0.647,0.349}{ and}%
\MapoToken{0.263,0.675,0.369}{ the}%
\MapoToken{0.145,0.553,0.278}{ sound}%
\MapoToken{0.184,0.592,0.306}{ is}%
\MapoToken{0.208,0.620,0.325}{ a}%
\MapoToken{0.184,0.592,0.306}{ sharp}%
\MapoToken{0.231,0.643,0.345}{,}%
\MapoToken{0.129,0.537,0.267}{ perc}%
\MapoToken{0.078,0.494,0.227}{uss}%
\MapoToken{0.235,0.647,0.349}{ive}%
\MapoToken{0.133,0.541,0.267}{ clip}%
\MapoToken{0.157,0.565,0.286}{-c}%
\MapoToken{0.157,0.565,0.286}{lop}%
\MapoToken{0.176,0.584,0.302}{ rather}%
\MapoToken{0.369,0.725,0.420}{ than}%
\MapoToken{0.431,0.757,0.451}{ a}%
\MapoToken{0.427,0.753,0.447}{ rubber}%
\MapoToken{0.345,0.714,0.408}{ tire}%
\MapoToken{0.165,0.573,0.290}{ sound}%
\MapoToken{0.184,0.592,0.306}{,}%
\MapoToken{0.243,0.655,0.353}{ so}%
\MapoToken{0.196,0.608,0.318}{ it}%
\MapoToken{0.153,0.561,0.282}{ is}%
\MapoToken{0.165,0.573,0.290}{ incorrect}%
\MapoToken{0.212,0.624,0.329}{;}%
\MapoToken{0.369,0.725,0.420}{ for}%
\MapoToken{0.149,0.557,0.278}{ Train}%
\MapoToken{0.200,0.612,0.322}{,}%
\MapoToken{0.212,0.624,0.329}{ it}%
\MapoToken{0.220,0.631,0.337}{ has}%
\MapoToken{0.247,0.663,0.361}{ a}%
\MapoToken{0.247,0.663,0.361}{ characteristic}%
\MapoToken{0.357,0.722,0.416}{ '}%
\MapoToken{0.216,0.627,0.333}{click}%
\MapoToken{0.306,0.698,0.392}{ety}%
\MapoToken{0.388,0.737,0.431}{-cl}%
\MapoToken{0.153,0.561,0.282}{ack}%
\MapoToken{0.196,0.608,0.318}{'}%
\MapoToken{0.192,0.604,0.314}{ sound}%
\MapoToken{0.259,0.671,0.365}{ from}%
\MapoToken{0.322,0.702,0.396}{ steel}%
\MapoToken{0.243,0.655,0.353}{ wheels}%
\MapoToken{0.333,0.710,0.404}{ on}%
\MapoToken{0.353,0.718,0.412}{ steel}%
\MapoToken{0.184,0.596,0.310}{ tracks}%
\MapoToken{0.176,0.584,0.302}{ which}%
\MapoToken{0.231,0.643,0.345}{ is}%
\MapoToken{0.275,0.682,0.376}{ a}%
\MapoToken{0.239,0.651,0.353}{ very}%
\MapoToken{0.224,0.635,0.341}{ regular}%
\MapoToken{0.290,0.686,0.380}{,}%
\MapoToken{0.259,0.671,0.365}{ metallic}%
\MapoToken{0.192,0.604,0.314}{ sound}%
\MapoToken{0.169,0.576,0.294}{,}%
\MapoToken{0.169,0.576,0.294}{ but}%
\MapoToken{0.200,0.612,0.322}{ the}%
\MapoToken{0.157,0.565,0.286}{ audio}%
\MapoToken{0.169,0.576,0.294}{ has}%
\MapoToken{0.231,0.643,0.345}{ a}%
\MapoToken{0.239,0.651,0.353}{ more}%
\MapoToken{0.227,0.639,0.341}{ organic}%
\MapoToken{0.282,0.682,0.376}{,}%
\MapoToken{0.353,0.718,0.412}{ '}%
\MapoToken{0.271,0.678,0.373}{th}%
\MapoToken{0.271,0.678,0.373}{udd}%
\MapoToken{0.204,0.616,0.325}{ing}%
\MapoToken{0.271,0.678,0.373}{'}%
\MapoToken{0.220,0.631,0.337}{ quality}%
\MapoToken{0.294,0.690,0.384}{ like}%
\MapoToken{0.251,0.667,0.365}{ ho}%
\MapoToken{0.216,0.627,0.333}{oves}%
\MapoToken{0.188,0.600,0.314}{ and}%
\MapoToken{0.243,0.655,0.353}{ the}%
\MapoToken{0.129,0.537,0.267}{ sound}%
\MapoToken{0.125,0.533,0.263}{ is}%
\MapoToken{0.259,0.671,0.365}{ not}%
\MapoToken{0.306,0.698,0.392}{ pure}%
\MapoToken{0.290,0.686,0.380}{ metal}%
\MapoToken{0.486,0.784,0.486}{-on}%
\MapoToken{0.239,0.651,0.353}{-metal}%
\MapoToken{0.204,0.616,0.325}{ but}%
\MapoToken{0.212,0.624,0.329}{ has}%
\MapoToken{0.259,0.671,0.365}{ a}%
\MapoToken{0.227,0.639,0.341}{ biological}%
\MapoToken{0.161,0.569,0.290}{ component}%
\MapoToken{0.184,0.592,0.306}{,}%
\MapoToken{0.220,0.631,0.337}{ so}%
\MapoToken{0.141,0.549,0.275}{ it}%
\MapoToken{0.180,0.588,0.302}{ is}%
\MapoToken{0.192,0.604,0.314}{ unlikely}%
\MapoToken{0.247,0.663,0.361}{;}%
\MapoToken{0.475,0.776,0.478}{ for}%
\MapoToken{0.384,0.733,0.427}{ Aer}%
\MapoToken{0.447,0.765,0.459}{op}%
\MapoToken{0.231,0.643,0.345}{lane}%
\MapoToken{0.314,0.698,0.392}{,}%
\MapoToken{0.325,0.706,0.400}{ it}%
\MapoToken{0.275,0.682,0.376}{ would}%
\MapoToken{0.259,0.671,0.365}{ be}%
\MapoToken{0.243,0.655,0.353}{ dominated}%
\MapoToken{0.427,0.753,0.447}{ by}%
\MapoToken{0.463,0.773,0.471}{ the}%
\MapoToken{0.243,0.655,0.353}{ sound}%
\MapoToken{0.525,0.800,0.522}{ of}%
\MapoToken{0.580,0.827,0.565}{ a}%
\MapoToken{0.518,0.800,0.514}{ powerful}%
\MapoToken{0.345,0.714,0.408}{ engine}%
\MapoToken{0.490,0.784,0.494}{ (}%
\MapoToken{0.502,0.792,0.502}{a}%
\MapoToken{0.306,0.698,0.392}{ roar}%
\MapoToken{0.498,0.788,0.498}{ or}%
\MapoToken{0.533,0.808,0.529}{ a}%
\MapoToken{0.263,0.675,0.369}{ drone}%
\MapoToken{0.239,0.651,0.353}{)}%
\MapoToken{0.357,0.722,0.416}{ and}%
\MapoToken{0.439,0.761,0.455}{ the}%
\MapoToken{0.239,0.651,0.353}{ sound}%
\MapoToken{0.463,0.773,0.471}{ of}%
\MapoToken{0.384,0.733,0.427}{ air}%
\MapoToken{0.322,0.702,0.396}{ rushing}%
\MapoToken{0.243,0.655,0.353}{,}%
\MapoToken{0.212,0.624,0.329}{ which}%
\MapoToken{0.204,0.616,0.325}{ is}%
\MapoToken{0.227,0.639,0.341}{ completely}%
\MapoToken{0.188,0.600,0.314}{ different}%
\MapoToken{0.161,0.569,0.290}{ from}%
\MapoToken{0.216,0.627,0.333}{ the}%
\MapoToken{0.180,0.588,0.302}{ audio}%
\MapoToken{0.263,0.675,0.369}{,}%
\MapoToken{0.294,0.690,0.384}{ so}%
\MapoToken{0.227,0.639,0.341}{ it}%
\MapoToken{0.263,0.675,0.369}{ is}%
\MapoToken{0.180,0.588,0.302}{ incorrect}%
\MapoToken{0.259,0.671,0.365}{;}%
\MapoToken{0.314,0.698,0.392}{ and}%
\MapoToken{0.376,0.729,0.424}{ for}%
\MapoToken{0.169,0.576,0.294}{ Horse}%
\MapoToken{0.239,0.651,0.353}{-d}%
\MapoToken{0.247,0.663,0.361}{rawn}%
\MapoToken{0.173,0.580,0.298}{ wagon}%
\MapoToken{0.204,0.616,0.325}{,}%
\MapoToken{0.247,0.659,0.357}{ it}%
\MapoToken{0.184,0.596,0.310}{ fits}%
\MapoToken{0.192,0.604,0.314}{ perfectly}%
\MapoToken{0.239,0.651,0.353}{ because}%
\MapoToken{0.247,0.663,0.361}{ the}%
\MapoToken{0.169,0.576,0.294}{ sound}%
\MapoToken{0.212,0.624,0.329}{ is}%
\MapoToken{0.239,0.651,0.353}{ the}%
\MapoToken{0.259,0.671,0.365}{ classic}%
\MapoToken{0.200,0.612,0.322}{ sound}%
\MapoToken{0.212,0.624,0.329}{ of}%
\MapoToken{0.275,0.682,0.376}{ a}%
\MapoToken{0.161,0.569,0.290}{ horse}%
\MapoToken{0.212,0.624,0.329}{'s}%
\MapoToken{0.247,0.663,0.361}{ ho}%
\MapoToken{0.184,0.592,0.306}{oves}%
\MapoToken{0.251,0.667,0.365}{ (}%
\MapoToken{0.271,0.678,0.373}{the}%
\MapoToken{0.165,0.573,0.290}{ clip}%
\MapoToken{0.227,0.639,0.341}{-c}%
\MapoToken{0.192,0.604,0.314}{lop}%
\MapoToken{0.247,0.659,0.357}{),}%
\MapoToken{0.400,0.741,0.435}{ the}%
\MapoToken{0.200,0.612,0.322}{ sound}%
\MapoToken{0.337,0.710,0.404}{ of}%
\MapoToken{0.431,0.757,0.451}{ the}%
\MapoToken{0.263,0.675,0.369}{ wagon}%
\MapoToken{0.302,0.694,0.388}{'s}%
\MapoToken{0.216,0.627,0.333}{ wheels}%
\MapoToken{0.275,0.682,0.376}{ (}%
\MapoToken{0.239,0.651,0.353}{which}%
\MapoToken{0.247,0.659,0.357}{ could}%
\MapoToken{0.306,0.698,0.392}{ be}%
\MapoToken{0.302,0.694,0.388}{ wooden}%
\MapoToken{0.498,0.788,0.498}{ or}%
\MapoToken{0.365,0.725,0.420}{ metal}%
\MapoToken{0.259,0.671,0.365}{)}%
\MapoToken{0.200,0.612,0.322}{ rolling}%
\MapoToken{0.204,0.616,0.325}{ on}%
\MapoToken{0.259,0.671,0.365}{ a}%
\MapoToken{0.220,0.631,0.337}{ road}%
\MapoToken{0.184,0.596,0.310}{,}%
\MapoToken{0.243,0.655,0.353}{ and}%
\MapoToken{0.302,0.694,0.388}{ the}%
\MapoToken{0.204,0.616,0.325}{ harness}%
\MapoToken{0.251,0.667,0.365}{ c}%
\MapoToken{0.184,0.592,0.306}{reak}%
\MapoToken{0.161,0.569,0.290}{ing}%
\MapoToken{0.161,0.569,0.290}{,}%
\MapoToken{0.165,0.573,0.290}{ with}%
\MapoToken{0.176,0.584,0.302}{ the}%
\MapoToken{0.086,0.502,0.235}{ rhyth}%
\MapoToken{0.169,0.576,0.294}{mic}%
\MapoToken{0.196,0.608,0.318}{ nature}%
\MapoToken{0.196,0.608,0.318}{ matching}%
\MapoToken{0.275,0.682,0.376}{ a}%
\MapoToken{0.216,0.627,0.333}{ horse}%
\MapoToken{0.047,0.467,0.208}{ trot}%
\MapoToken{0.133,0.541,0.267}{ting}%
\MapoToken{0.357,0.722,0.416}{ or}%
\MapoToken{0.306,0.698,0.392}{ can}%
\MapoToken{0.133,0.541,0.267}{tering}%
\MapoToken{0.184,0.596,0.310}{.}%
\MapoToken{0.165,0.573,0.290}{ Therefore}%
\MapoToken{0.196,0.608,0.318}{,}%
\MapoToken{0.271,0.678,0.373}{ the}%
\MapoToken{0.247,0.659,0.357}{ transportation}%
\MapoToken{0.306,0.698,0.392}{ mode}%
\MapoToken{0.212,0.624,0.329}{ referred}%
\MapoToken{0.173,0.580,0.298}{ to}%
\MapoToken{0.192,0.604,0.314}{ in}%
\MapoToken{0.000,0.369,0.149}{ the}%
\MapoToken{0.192,0.604,0.314}{ audio}%
\MapoToken{0.243,0.655,0.353}{ is}%
\MapoToken{0.247,0.663,0.361}{ Horse}%
\MapoToken{0.325,0.706,0.400}{-d}%
\MapoToken{0.306,0.698,0.392}{rawn}%
\MapoToken{0.169,0.576,0.294}{ wagon}%
\MapoToken{0.086,0.502,0.235}{. }%
\MapoToken{0.110,0.522,0.251}{\textless{}/think\textgreater{}}%
\MapoToken{0.165,0.573,0.290}{\phantom{x}}%
\MapoToken{0.259,0.671,0.365}{H}%
\MapoToken{0.208,0.620,0.325}{orse}%
\MapoToken{0.322,0.702,0.396}{-d}%
\MapoToken{0.224,0.635,0.341}{rawn}%
\MapoToken{0.075,0.490,0.224}{ wagon}%
\end{MapoTokenStream}
\MapoMetricRule

\endgroup

%% file: tables/pos_gate.tex
% Generated by scripts/make_table_appendix.py; do not edit by hand.
\begin{table}[!htb]
\centering
\caption{POS-gate ablation under full fine-tuning. Entries are mean $\pm$ sample SD across three seeds. Avg. is the unweighted four-benchmark mean computed within each seed.}
\label{tab:pos_gate}
\setlength{\tabcolsep}{3pt}
\begin{tabular}{@{}lrrrrr@{}}
\toprule
Configuration & MMAU & MMAR & MMSU & MMAU-Pro & Avg. \\
\midrule
Without POS & $76.73\pm0.64$ & $69.87\pm0.65$ & $78.77\pm0.25$ & $64.20\pm0.36$ & $72.39\pm0.22$ \\
Full MAPO & $77.83\pm0.25$ & $70.83\pm0.21$ & $79.35\pm0.15$ & $64.98\pm0.28$ & $73.25\pm0.09$ \\
\bottomrule
\end{tabular}
\end{table}

%% file: tables/training_cost.tex
% Generated by scripts/make_table_appendix.py; do not edit by hand.
\begin{table}[!htb]
\centering
\caption{Training time on the 29k-example audio QA dataset using 24 GPUs. The first two rows give reported per-step times for each step type; the final row gives elapsed time over the full 499-step window. Relative increase is measured against GRPO.}
\label{tab:training_cost}
\setlength{\tabcolsep}{9pt}
\begin{tabular}{@{}lrrr@{}}
\toprule
Window & MAPO & GRPO & Increase \\
\midrule
Rollout + update (124 steps) & 285 s & 262 s & 8.8\% \\
Update only (375 steps) & 59 s & 51 s & 15.7\% \\
Total (499 steps) & 15.96 h & 14.34 h & 11.3\% \\
\bottomrule
\end{tabular}
\end{table}